\documentclass[acmtog]{acmart}
\usepackage{booktabs} % For formal tables
\usepackage{multirow} % for table by sam
\usepackage{xcolor} % for table by sam
\usepackage{colortbl} % for table by sam
\usepackage{threeparttable} % for table by sam
\usepackage{subfigure} % for figure by +D

\let\oldtextcolor\textcolor
\renewcommand{\textcolor}[2]{%
  \ifstrequal{#1}{red}{%
    \oldtextcolor{black}{#2}%
  }{%
    \ifstrequal{#1}{orange}{%
      \oldtextcolor{black}{#2}%
    }{%
      \oldtextcolor{#1}{#2}%
    }%
  }%
}

\usepackage{hyperref} % for section reference

\usepackage{amssymb, amsfonts} % for math
\usepackage{verbatim}
\newcommand{\BestCellColor}{\cellcolor{red!25}}
\newcommand{\SecondBestCellColor}{\cellcolor{orange!25}}

\makeatletter
\DeclareRobustCommand\onedot{\futurelet\@let@token\@onedot}
\def\@onedot{\ifx\@let@token.\else.\null\fi\xspace}
\def\eg{\emph{e.g}\onedot} 
\def\ie{\emph{i.e}\onedot} 
 
\def\etc{\emph{etc}\onedot} 
 
\def\etal{\emph{et al}\onedot}
\makeatother

\usepackage[ruled]{algorithm2e} % For algorithms

\SetAlFnt{\small}
\SetAlCapFnt{\small}
\SetAlCapNameFnt{\small}
\SetAlCapHSkip{0pt}

\setcopyright{rightsretained}
\acmJournal{TOG}
\acmYear{2024} \acmVolume{43} \acmNumber{6} \acmArticle{} \acmMonth{12}\acmDOI{10.1145/3687762}
\begin{document}
% Title portion
\title{LetsGo: Large-Scale Garage Modeling and Rendering via LiDAR-Assisted
Gaussian Primitives}

\author{Jiadi Cui}\authornote{Equal contributions.}
\affiliation{%
	\institution{ShanghaiTech University}
	\city{Shanghai}
	\country{China}}
\affiliation{%
	\institution{Stereye Intelligent Technology Co.,Ltd.}
	%\city{Shanghai}
	\country{China}}
\email{cuijd@shanghaitech.edu.cn}

\author{Junming Cao}\authornotemark[1]
\affiliation{%
	\institution{Shanghai Advanced Research Institute, Chinese Academy of Sciences}
	\city{Shanghai}
	\country{China}}
\affiliation{%
        \institution{University of Chinese Academy of Sciences}
        \city{Beijing}
        \country{China}}
\email{caojm@sari.ac.cn}

\author{Fuqiang Zhao}\authornotemark[1]
\affiliation{%
	\institution{ShanghaiTech University}
	\city{Shanghai}
	\country{China}}
\affiliation{%
	\institution{NeuDim Digital Technology (Shanghai) Co.,Ltd.}
	%\city{Shanghai}
	\country{China}}
\email{zhaofq@shanghaitech.edu.cn}

\author{Zhipeng He}
\affiliation{
	\institution{ShanghaiTech University}
	\city{Shanghai}
	\country{China}
}
\email{hezhp2023@shanghaitech.edu.cn}

\author{Yifan Chen}
\affiliation{
	\institution{ShanghaiTech University}
	\city{Shanghai}
	\country{China}
}
\email{chenyf2022@shanghaitech.edu.cn}

\author{Yuhui Zhong}
\affiliation{
	\institution{DGene Digital Technology Co., Ltd.}
	%\city{Shanghai}
	\country{China}
}
\email{yuhui.zhong@dgene.com}

\author{Lan Xu}
\affiliation{
	\institution{ShanghaiTech University}
	\city{Shanghai}
	\country{China}
}
\email{xulan1@shanghaitech.edu.cn}

\author{Yujiao Shi}\authornote{Corresponding author.}
\affiliation{
	\institution{ShanghaiTech University}
	\city{Shanghai}
	\country{China}
}
\email{s.yujiao93@gmail.com}

\author{Yingliang Zhang}\authornotemark[2]
\affiliation{
	\institution{DGene Digital Technology Co., Ltd.}
	%\city{Shanghai}
	\country{China}
}
\email{yingliang.dgene@gmail.com}

\author{Jingyi Yu}\authornotemark[2]
\affiliation{
	\institution{ShanghaiTech University}
	\city{Shanghai}
	\country{China}
}
\email{yujingyi@shanghaitech.edu.cn}

\begin{abstract}

Large garages are ubiquitous yet intricate scenes that present unique challenges due to their monotonous colors, repetitive patterns, reflective surfaces, and transparent vehicle glass. Conventional Structure from Motion (SfM) methods for camera pose estimation and 3D reconstruction often fail in these environments due to poor correspondence construction. To address these challenges, we introduce LetsGo, a LiDAR-assisted Gaussian splatting framework for large-scale garage modeling and rendering.
We develop a handheld scanner, Polar, equipped with IMU, LiDAR, and a fisheye camera, to facilitate accurate data acquisition. Using this Polar device, we present the GarageWorld dataset, consisting of eight expansive garage scenes with diverse geometric structures, which will be made publicly available for further research.
Our approach demonstrates that LiDAR point clouds collected by the Polar device significantly enhance a suite of 3D Gaussian splatting algorithms for garage scene modeling and rendering. We introduce a novel depth regularizer that effectively eliminates floating artifacts in rendered images.
Additionally, we propose a multi-resolution 3D Gaussian representation designed for Level-of-Detail (LOD) rendering. This includes adapted scaling factors for individual levels and a random-resolution-level training scheme to optimize the Gaussians across different resolutions. This representation enables efficient rendering of large-scale garage scenes on lightweight devices via a web-based renderer.
Experimental results on our GarageWorld dataset, as well as on ScanNet++ and KITTI-360, demonstrate the superiority of our method in terms of rendering quality and resource efficiency.
\end{abstract}

\begin{CCSXML}
<ccs2012>
   <concept>
       <concept_id>10010147.10010371.10010382.10010236</concept_id>
       <concept_desc>Computing methodologies~Computational photography</concept_desc>
       <concept_significance>500</concept_significance>
       </concept>
   <concept>
       <concept_id>10010147.10010371.10010382.10010385</concept_id>
       <concept_desc>Computing methodologies~Image-based rendering</concept_desc>
       <concept_significance>500</concept_significance>
       </concept>
 </ccs2012>
\end{CCSXML}

\ccsdesc[500]{Computing methodologies~Computational photography}
\ccsdesc[500]{Computing methodologies~Image-based rendering}


\keywords{Neural rendering, large-scale garage modeling, LiDAR scanning, 3D Gaussian splatting, garage dataset, level-of-detail rendering}

\begin{teaserfigure}
	\centering
	\includegraphics[width=1.0\linewidth]{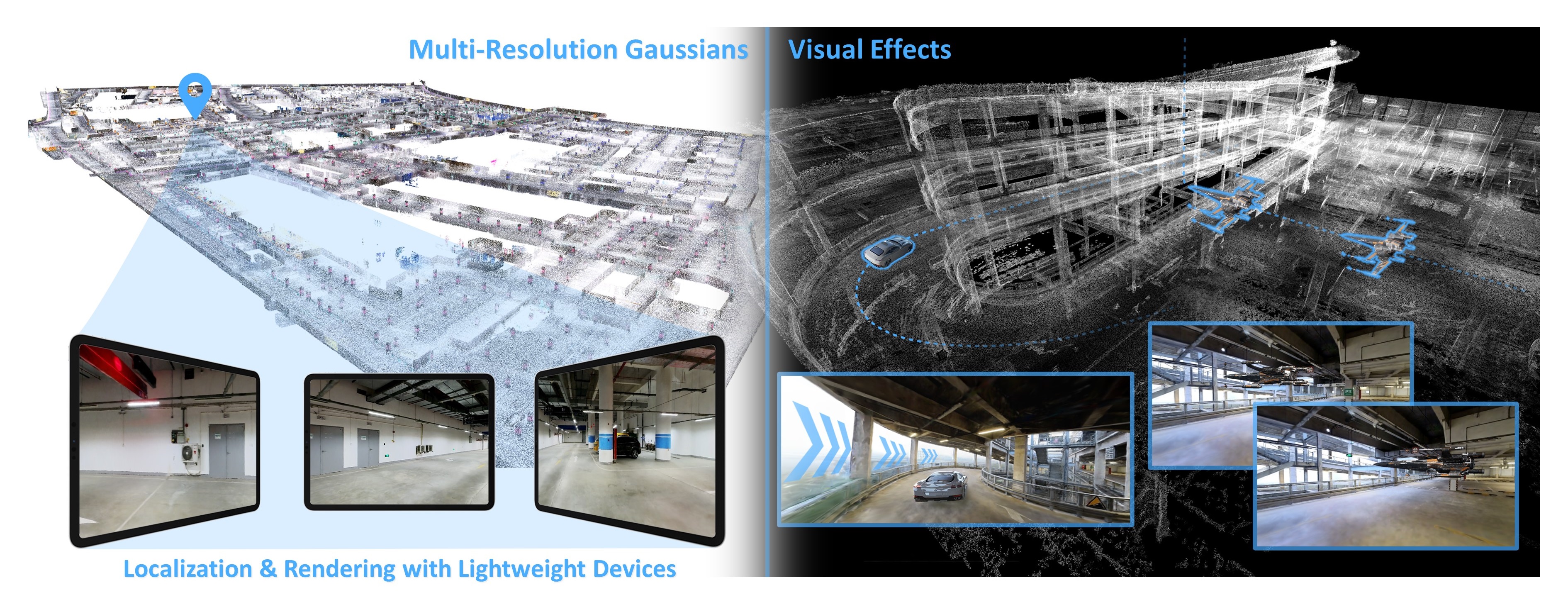}
	\vspace{-21pt}
	\caption{We present LetsGo - an explicit and efficient end-to-end framework for high-fidelity rendering of large-scale garages. We design a handheld Polar scanner to capture RGBD data of expansive parking environments and have scanned a garage dataset, named GarageWorld, comprising eight garages with different structures. Our LiDAR-assisted Gaussian primitives approach along with GarageWorld dataset enables various applications, such as autonomous vehicle localization, navigation and parking, as well as VFX production.}
	\label{fig:teaser}
\end{teaserfigure}
\maketitle

\section{Introduction}

Modeling garage environments accurately is crucial for various applications such as autonomous vehicle testing, architectural planning, and game design. Garages present a unique set of challenges due to their complex geometries, varying lighting conditions, and frequent presence of obstacles. The ability to create detailed and realistic 3D models of garages can significantly enhance the effectiveness of these applications, providing more accurate simulations and analyses. For visual artists, garages represent a frontier in visual simulation that merges the aesthetic with the technical, offering a canvas where the intricacies of light, shadow, texture, and space coalesce. The complex interplay of artificial and natural lighting within the confines of a garage, with its reflective surfaces, varying materials, and intricate geometries, provides a rigorous testbed for modeling and rendering, pushing the boundaries of what is achievable in virtual environments.

However, capturing and rendering garage environments pose\textcolor{orange}{s} significant difficulties. 
% generally poorly lit and are mainly composed of texture-less walls, sloped surfaces with distorted lines, cluttered and often reflective environments, and complex occlusions between walls, cars, and pillars . 
These spaces often have low-light conditions, textureless surfaces, and a high degree of clutter (Fig.~\ref{fig:dataset_gallery}). 
% making traditional modeling techniques less effective. 
The intricacies of garage layouts, including narrow spaces and reflective surfaces, further complicate the modeling process. 
% requiring advanced techniques to achieve high-quality results. 
Garages with internal circular or spiral paths often lead to incomplete data and ambiguities in spatial relationships. Additionally, their extensive spatial area underscores the necessity for efficient 3D representations and lightweight rendering techniques to facilitate real-time interaction and visualization, particularly for scenarios demanding rapid situational assessment, such as navigation and path adjustment.

Existing methods for garage modeling typically rely on either manual measurements or conventional photogrammetry and LiDAR scanning. Traditional computer vision techniques, such as Structure from Motion (SfM) ~\cite{snavely2008skeletal,schonberger2016structure} and Multi-view Stereo (MVS)~\cite{yao2018mvsnet,5226635}, often struggle in these environments due to the prevalence of texture-less regions and repetitive structural designs. These methods frequently fail to extract sufficient feature points and establish accurate feature correspondences necessary for estimating camera poses. Active sensing technologies based on LiDAR can calculate camera poses and scene geometry using SLAM algorithms, but the reflective materials and transparent car windows common in garages lead to geometric inaccuracies. Moreover, LiDAR data tend to be sparse, containing many holes that corrupt high-frequency textures essential for rendering the color appearance of the scene. 

While recent advances in neural representation, particularly Neural Radiance Fields (NeRF)~\cite{mildenhall2021nerf}, have shown promise in producing high-quality renderings, they come with high computational costs and lengthy training times. Although enhancements~\cite{SunSC22,zhang2020nerf++, xu2022point, kangle2021dsnerf} to NeRF aim to optimize training duration and visual rendering quality, integrating such implicit representations into conventional graphics rendering pipelines and tools for rapid 3D content applications remains challenging. The emerging 3D Gaussian Splatting (3DGS)~\cite{kerbl3Dgaussians} method revisits explicit representations, using 3D Gaussians to articulate the geometry and appearance of scenes, achieving high-quality scene modeling and rendering. 
Recent works~\cite{hierarchicalgaussians24, liu2024citygaussian, ren2024octree, scaffoldgs, LoG} extend the 3DGS approach to model large-scale outdoor scenes, achieving impressive results. 
% However, few methods pay attention to modeling large-scale indoor scenes, such as underground garages, which present unique challenges due to low lighting conditions, large texture-less regions, and repetitive patterns. These factors complicate the establishment of sufficient feature correspondences between different images.
However, few methods address the unique challenges of modeling large-scale indoor scenes like underground garages, where low lighting, large texture-less regions, and repetitive patterns complicate the establishment of sufficient feature correspondences between different images.

This paper introduces LetsGo, an explicit and efficient end-to-end modeling scheme for high-fidelity rendering of large-scale garages. Our key innovation is the integration of calibrated LiDAR points into 3D Gaussian splatting algorithms. We design a handheld Polar scanner, which combines IMU, LiDAR, and a fisheye camera for robust relative pose estimation, specifically tailored for expansive garage data collection. We scan eight large-scale garages, collectively named GarageWorld, using this Polar scanner. To our knowledge, this dataset is the first of its kind aimed at large-scale garages and will be made available to the community. Our experiments demonstrate that LiDAR points collected by the Polar device effectively support various Gaussian splatting algorithms for detailed garage scene representation. To enhance the quality of 3D Gaussian rendering, we introduce a depth regularizer that uses depth priors as supervisory signals, significantly reducing floating artifacts and enabling high rendering quality.

As the scene size increases, the memory demands for rendering large amounts of 3D Gaussians can exceed the capabilities of even high-end GPUs. To address this challenge, we propose a multi-resolution 3D Gaussian representation tailored for Level-of-Detail (LOD) rendering. This approach dynamically adjusts based on the camera's position, orientation, and viewing frustum, allowing for real-time, high-quality rendering of expansive scenes. Specifically, we construct different levels of Gaussians at varying resolutions, where lower-resolution levels capture coarse scene characteristics and higher-resolution levels reconstruct fine, high-frequency details. We employ tailored scaling factors for each level and a random-resolution-level training scheme to optimize the Gaussians across different levels.

During rendering, we introduce a novel level selection strategy that optimizes the trade-off between visual fidelity and device performance by considering the distance between 3D Gaussians and the rendering viewpoint. With our multi-resolution 3D Gaussian framework, we develop an LOD PC viewer that achieves rendering speeds up to four times faster than traditional 3DGS viewers on high-performance GPUs (\eg, RTX 3090). Additionally, we offer a lightweight web renderer designed to support LOD rendering across various consumer-level devices, including laptops and tablets. 
\textcolor{orange}{We have released our source codes, including our training code, high-performance PC viewer, and lightweight web viewer, to facilitate reproducible research. Please refer to our \href{https://zhaofuq.github.io/LetsGo/}{\textit{\textcolor{blue}{project page}}}.} 

Our results, gathered from the GarageWorld dataset as well as ScanNet++~\cite{dai2017scannet} and KITTI-360~\cite{liao2022kitti} datasets, indicate that our approach not only surpasses other methods in rendering quality but also maintains high rendering efficiency. The GarageWorld dataset and the LetsGo framework for large-scale garage modeling and rendering enable various applications, including autonomous driving, localization, navigation, visual effects, \etc.

\section{Related Work}

% \subsection{Conventional Explicit Visual Reconstruction}

% Conventional algorithms for reconstructing large-scale scenes encompass a trio of pivotal methodologies. These include Structure from Motion (SFM)\cite{snavely2008skeletal, crandall2011discrete}, a technique dedicated to discerning the three-dimensional structure of a scene through the sequential analysis of two-dimensional image frames. Simultaneous Localization and Mapping (SLAM)\cite{leonard1991simultaneous,huang2019survey} is another cornerstone, aiming to concurrently determine the position of a sensor and create a map of the surrounding environment. Completing this triumvirate is Multi-View Stereo (MVS)~\cite{seitz2006comparison, goesele2007multi}, a method that leverages multiple images to construct a comprehensive three-dimensional model of the scene.

% \hfill

\paragraph{Conventional Explicit Visual Reconstruction.}
Conventional algorithms for reconstructing large-scale scenes include Structure from Motion (SfM)~\cite{wu2013towards, sweeney2016large, moulon2013adaptive}, Simultaneous Localization and Mapping (SLAM)~\cite{leonard1991simultaneous, bavle2023slam, bujanca2021robust, ceriani2015pose} and Multi-View Stereo (MVS)~\cite{seitz2006comparison, goesele2007multi, 5539802}. 
They are dedicated to discerning the three-dimensional structure of a scene through the sequential or multi-view analysis of two-dimensional image frames.
All these methods leverage feature tracking and multi-view consistency to recover the 3D scene structures. 
SfM- and SLAM-based methods~\cite{crandall2011discrete, agarwal2011building, frahm2010building, heinly2015reconstructing, teed2021droid, schmuck2021covins, li2020attention} estimate poses of input images and recover the scene structure jointly. However, their main purpose is pose estimation, and the recovered scene point clouds are always sparse, making it difficult for high-quality free-view synthesis.  
While MVS-based methods~\cite{li2022ds, dai2019mvs2, huang2021m3vsnet, yang2021self, xu2021digging, xu2021self}, especially \textcolor{red}{depth-based} approaches, compute a dense depth map for each input image, the constructed scenes often lack accuracy and robustness in texture-less and complex scenes.

% \subsection{Implicit Neural Scene Representations}
% \hfill

\begin{figure*}[ht]
    \centering
    \setlength{\abovecaptionskip}{0pt}
\setlength{\belowcaptionskip}{0pt}
    \subfigure[Scene]{
\label{fig:device:a}
\includegraphics[height=1.58in]{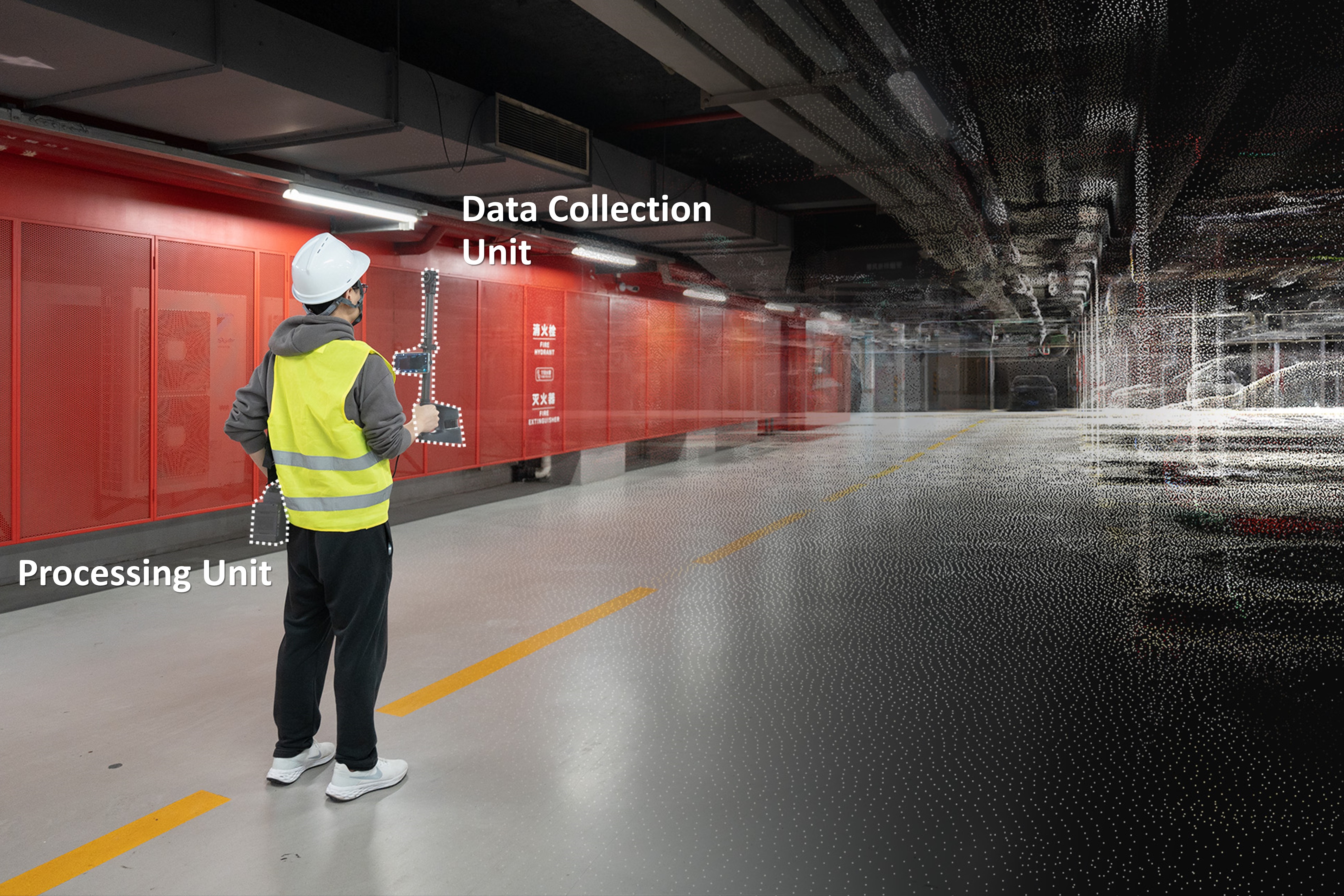}}
    \subfigure[Device]{
\label{fig:device:b}
\includegraphics[height=1.58in]{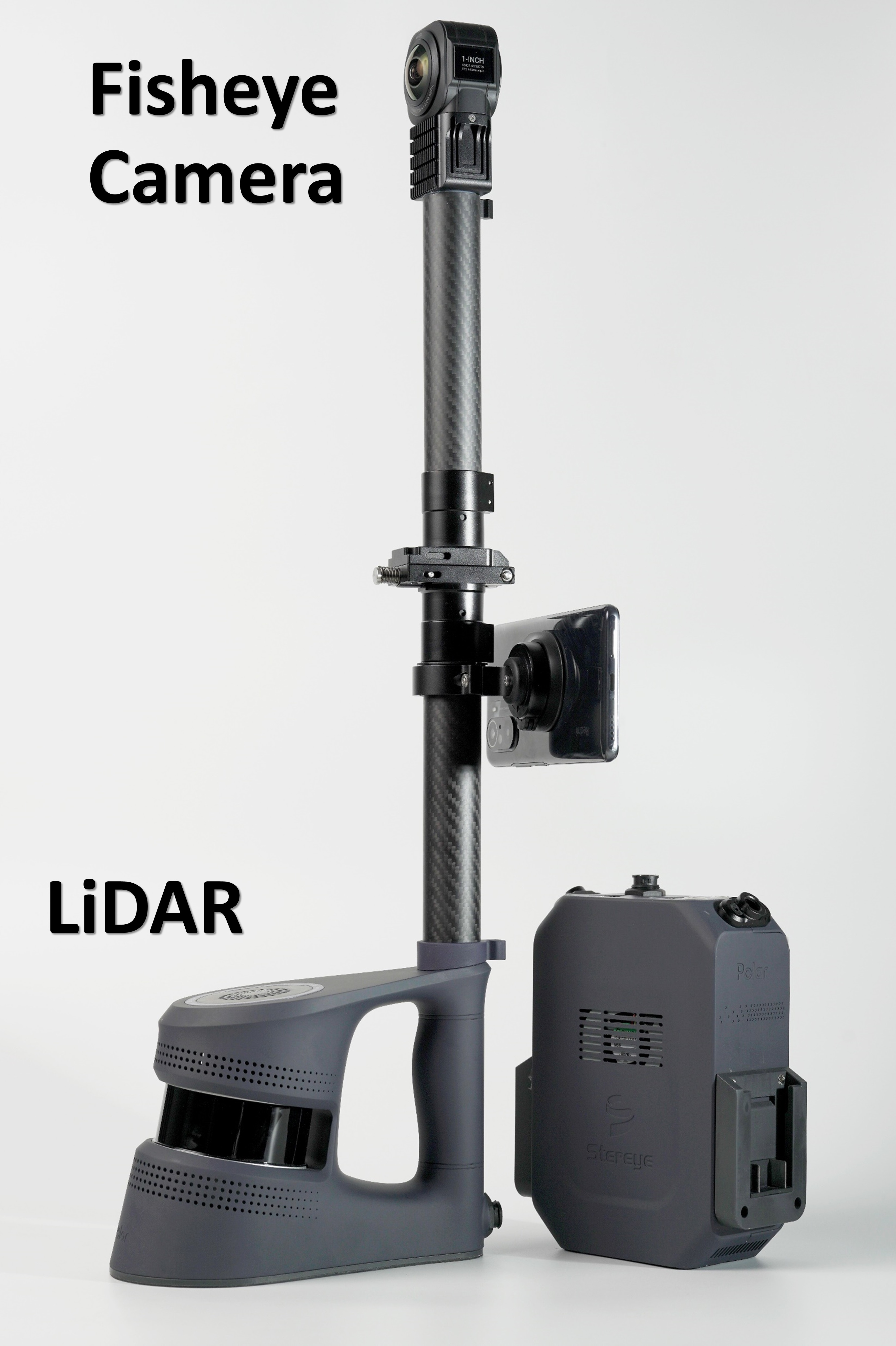}}
    \subfigure[Point Cloud]{
\label{fig:pcloud}
\includegraphics[height=1.58in]{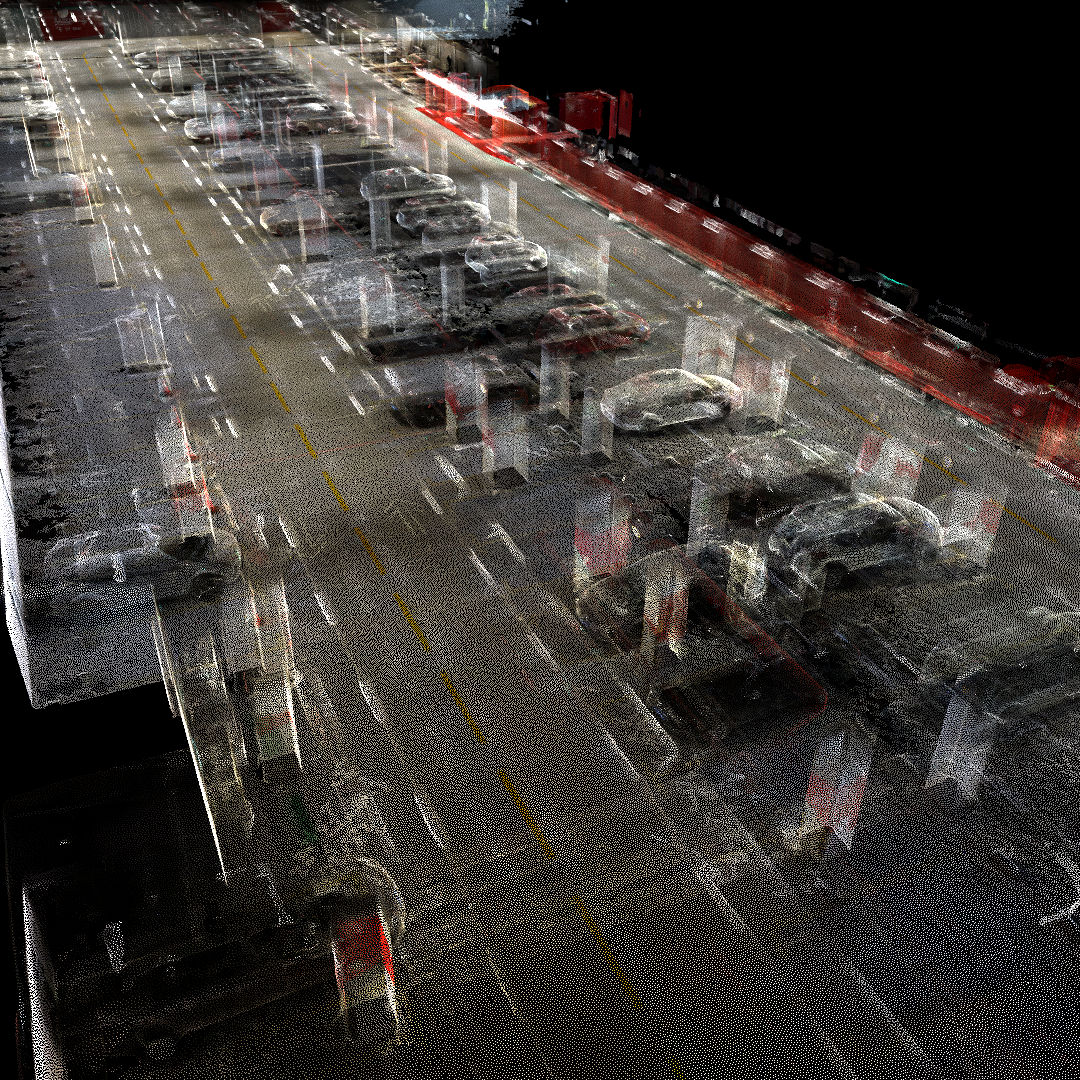}}
    \subfigure[Fisheye Image]{
\label{fig:fisheye}
\includegraphics[height=1.58in]{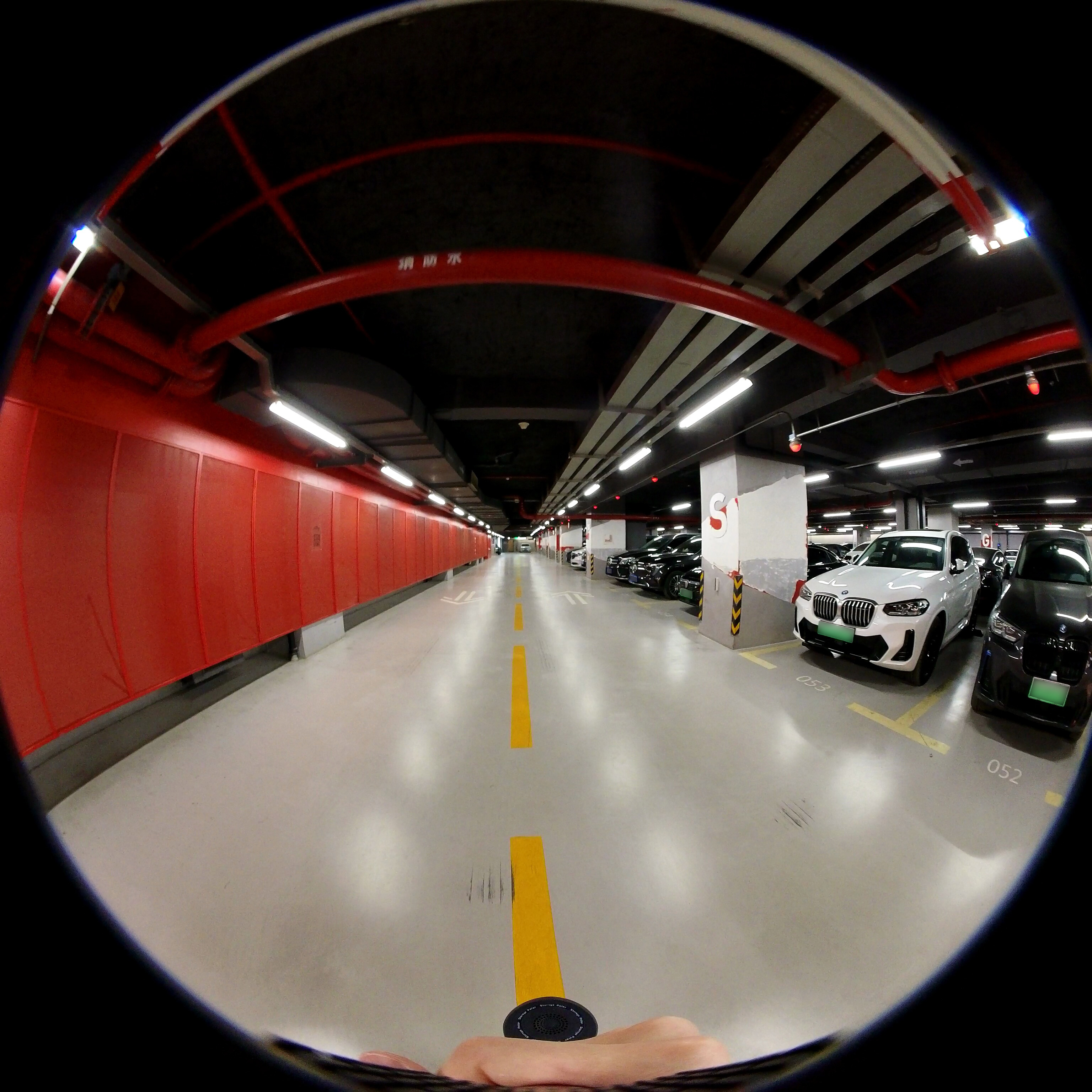}}
    \caption{
    Our compact Polar scanner (b) is engineered for capturing expansive garage environments (a). It is optimized for handheld operation or vehicular mounting, enabling versatile data capture in extensive spaces.
    At the core of Polar's data acquisition unit lies a high-fidelity LiDAR sensor, capturing precise 3D point clouds (c), complemented by a fisheye camera that procures wide-angle 2D RGB images (d) for a complete scene modeling.
    % Our handheld Polar scanner (b) for scanning large-scale garage scenes (a) consists of a Data Collection Unit and a Data Processing Unit, which can also be attached to a vehicle for data collection to reduce the effort due to the vastness of the environment. 
    % The data collection unit includes a LiDAR sensor to collect 3D point cloud (c) and a fisheye camera for 2D RGB image (d) capturing of the environment. 
    % Our data collection device consists of LiDAR and fisheye camera (Fig. \ref{fig:device:b}). We typically use a handheld device to scan parking garage scenes (Fig. \ref{fig:device:a}), but sometimes, we attach the device to a go-kart and drive around to collect data due to the vastness of the environment. Fig. \ref{fig:pcloud} represents the colored point cloud after our scanning and modeling, while Fig. \ref{fig:fisheye}  shows the images captured by the fisheye camera during scanning. In the original 3D GS training, we crop the fisheye image into five pinhole images based on five directions (up, down, left, right, front) to meet the training requirements.
    }
    \label{fig:device}
\end{figure*}

\begin{table*}[ht]
    \centering
    \setlength{\abovecaptionskip}{0pt}
\setlength{\belowcaptionskip}{0pt}
\caption{Detailed illustrations of our GarageWorld dataset, including various datasets categorized by their environment type, geometric features, area size, number of images, point count, face count, and lighting conditions
% GarageWorld dataset overview: The dataset encompasses detailed characteristics of five large-scale garage environments. It delineates the categorical classification of each garage, the geometric intricacies, the total scanned area, and the aggregate of fisheye images captured. Furthermore, it quantifies the points within the point clouds, the faces constituting the meshes, and depicts the lighting conditions in each dataset instance.
    }
    \resizebox{\textwidth}{!}{
    \begin{threeparttable}
    \begin{tabular}{lllcccccc}
    \toprule
    Dataset & Category  & Geometry & Aera($m^2$) & Image Num & Point Num &  Face Num & Lighting Condition \\ %& Time Cost \\ 
    \midrule
    Campus  1 & Underground  & Flat \& Sloped Paths & 38447.86 & 8479 & 1.9B  & 121.9M & Uniform Lighting \\%& $\sim$90min  \\ 
    Campus  2 & Underground  & Flat \& Sloped Paths  & 28046.37 & 7772 & 1.4B  & 95.8M  & Uniform Lighting \\%& $\sim$90min  \\
    Shopping Mall 1 & Indoor (Multi-floors)  & Spiral \& Circular Paths & 32646.68 & 5792 & 1.14B & 40.8M & Uneven Lighting  \\%& $\sim$200min \\
    Shopping Mall 2 & Outdoor      & Spiral \& Sloped Paths & 13495.92 & 2280 & 0.6B  & 77.1M  & Natural Lighting  \\%& $\sim$40min  \\%\\
    Shopping Mall 3 & Underground  & Flat   \& Sloped Paths & 30246.07 & 13296 & 1.6B  & 106.1M  & Uniform Lighting  \\%& $\sim$40min  \\%\\
    Office Building & Underground         & With Mechanical Parking System  & 22159.25 & 9308 & 1.15B & 72.3M  & Motion Sensor Lighting  \\%& $\sim$1ss40min \\
    Arts Center & Underground & Flat \& Sloped Paths  & 10392.32& 5779 & 0.52B & 31.4M & Uniform Lighting \\
    Subway Garage & Underground & Flat \& Sloped Paths & 7109.12 & 3607 & 0.70B & 43.5M & Uniform Lighting \\
    % Campus              & UPG\tnote{1}  & 133590.21 & 32502 & 6.6B & 435.4M & Perfect & $\sim$360min\\
    \bottomrule
    \end{tabular}
    % \begin{tablenotes}
    %     \footnotesize
        % \item[1] Underground Parking Garage 
        % \item[2] Surface Parking Lot
        % \item[3] Multi-Story Parking Garage
        % \item[1] with a Mechanical Parking System
    % \end{tablenotes}
    \end{threeparttable}
    }
    \label{tab:Dataset1}
    \vspace{-2mm}
\end{table*}

\paragraph{NeRF and 3D Gaussian Splatting Variants.}
Recent advances~\cite{noguchi2021neural, chen2021mvsnerf,yu2021pixelnerf,kerr2023lerf,rebain2021derf,zhao2022humannerf, zhao2022human} in neural scene representation have significantly impacted novel view synthesis. 
% Leveraging deep learning, these methods derive implicit representations directly from data, offering enhanced capabilities in capturing complex 3D scenes and rendering realistic views. 
Neural Radiance Fields (NeRF), introduced by Mildenhall et al.\textcolor{orange}{~\shortcite{mildenhall2021nerf}}, have revolutionized 3D reconstruction with a novel framework for detailed scene capture. Subsequent improvements~\cite{sun2023pointnerf++, kulhanek2023tetra, roessle2022depthpriorsnerf, rematas2022urban} have aimed at enhancing the visual fidelity and computational efficiency of NeRF. MipNeRF~\cite{barron2021mip} and MipNeRF360~\cite{barron2022mip} address aliasing via a novel conical frustum rendering technique and a nonlinear scene representation, respectively.
% include NeRF++~\cite{zhang2020nerf++}, which adapts NeRF for unbounded scenes with an inverted sphere parameterization, and MipNeRF~\cite{barron2021mip} along with MipNeRF360~\cite{barron2022mip}, which reduce aliasing through a novel conical frustum rendering scheme and a nonlinear scene representation respectively.
% To accelerate rendering speed, Plenoxels~\cite{fridovich2022plenoxels} optimizes a 3D grid of spherical harmonics. Instant-NGP~\cite{muller2022instant} employs a hash encoding method to maintain rendering quality with smaller networks and boost rendering speed. TensoRF~\cite{Chen2022ECCV} uses 4D tensors and tensor decompositions to boost rendering quality and reduce memory usage. 
Techniques such as Plexnoxels~\cite{fridovich2022plenoxels}, Instant-NGP~\cite{muller2022instant}, and TensoRF~\cite{Chen2022ECCV} have expedited rendering by integrating explicit encoding methods with compact MLP networks. Furthermore, Block-NeRF~\cite{tancik2022block} and Mega-NeRF~\cite{turki2022mega} facilitate the application of NeRF to extensive scenes.
% To address large-scale scenes, Block-NeRF~\cite{tancik2022block} partitions the environment into separate NeRFs to streamline updates and rendering times. Mega-NeRF~\cite{turki2022mega} adjusts to varied lighting across extensive urban areas. 
F2-NeRF~\cite{wang2023f2} introduces a space-warping method for handling arbitrary camera trajectories. ScaNeRF~\cite{wu2023scanerf} optimizes camera pose and scene representation jointly to address pose drift in large-scale scene reconstruction. 
Unlike these NeRF-based approaches, our method employs an explicit 3D Gaussian representation with depth priors from our RGBD scanner, enhancing realism and efficiency.
% Unlike these NeRF-based approaches, our method employs an explicit 3D Gaussian representation coupled with precise depth priors derived from our RGBD scanner, significantly enhancing the realism and efficiency of rendered images.

% Moreover, 3D Gaussian Splatting (3DGS)~\cite{kerbl3Dgaussians} has emerged as a transformative approach in novel view synthesis, characterized by its efficiency and lifelike visual quality.
3D Gaussian Splatting (3DGS)~\cite{kerbl3Dgaussians} is a transformative approach characterized by its efficiency and lifelike visual quality. 
% This method derives explicit 3D Gaussians from point clouds generated via SfM/SLAM/LiDAR, optimizing each Gaussian's position, anisotropic covariance, opacity, and spherical harmonic (SH) coefficients. Benefiting from a tile-based rasterizer, 3DGS achieves super-fast rendering with ultra-fine quality.
Extensive research~\cite{scaffoldgs, niedermayr2024compressed, keetha2024splatam, MatsukiCVPR2024, charatan2024pixelsplat, wu20244d, jiang2024hifi4g, tang2023dreamgaussian} has built on 3DGS.
Mip-Splatting~\cite{yu2024mip} mitigates artifacts from varying view sampling rates, and GaussianPro~\cite{cheng2024gaussianpro}, which enhances Gaussian distribution with 2D image constraints. DN-Splatter~\cite{turkulainen2024dnsplatter} and DNGaussian~\cite{li2024dngaussian} use depth and normal cues for refinement. SuGaR~\cite{guedon2024sugar} introduces Gaussian alignment regularization for explicit mesh extraction, and GaussianSurfels~\cite{Dai2024GaussianSurfels} flattens 3D Gaussian ellipsoids into 2D ellipses for accurate surface reconstructions. 
2DGS~\cite{Huang2DGS2024} transforms 3D volumes into 2D planar Gaussian disks, employing a perspective-accurate 2D splatting process aligned with geometric surfaces.

% TRIPS 结合3DGS和优秀的point based rendering两者的优点，使用了将点栅格化到屏幕空间图像金字塔的思路，实现对复杂细节高质量重建及渲染。
\textcolor{orange}{TRIPS~\cite{franke2024trips} combines the strengths of 3D Gaussian Splatting and advanced point-based rendering~\cite{kopanas2021point, aliev2020neural} by employing the concept of rasterizing points into a screen-space image pyramid, enabling high-quality reconstruction and rendering of complex details.} 
Despite these advances focusing on small-scale or object-level scenes, our approach innovates with a depth regularizer and a multi-resolution Gaussian representation designed for large-scale scenes, achieving realistic visual quality and enhanced rendering efficiency.

\paragraph{Gaussian Splatting for Large-Scale Scenes.} 
To effectively manage large-scale scenes, VastGaussian~\cite{lin2024vastgaussian} explores various partitioning strategies and introduces an appearance embedding module to enhance Gaussian training. Concurrent works such as DrivingGaussian~\cite{zhou2024drivinggaussian} and Street Gaussian~\cite{yan2024street} adapt Gaussian methodologies to the dynamic contexts of urban landscapes and autonomous driving scenarios, respectively. In addition, multiple studies are integrating the Level of Detail (LOD) rendering technique with 3D Gaussian Splatting to balance rendering quality and speed. 
Traditional mesh-based LOD strategies~\cite{Ponchio2016MultiresolutionAF, 10.1145/1015706.1015802} create varying levels of detail in object meshes, selecting the appropriate level based on the viewer’s distance to the target or the desired quality. Point-based LOD methods, like Potree~\cite{schutz2016potree} and FastLOD~\cite{schutz2020fast}, are designed to efficiently render large 3D point clouds, which is particularly beneficial for LiDAR data visualization.
% Mesh-based LOD techniques~\cite{XXX} typically involve creating levels of detail in meshes for objects. These meshes are selected based on the distance between viewpoint and rendering targets or the required quality of the view. Point-based LOD approaches~\cite{schutz2016potree, schutz2020fast} aim to render large 3D point clouds efficiently and are especially useful in applications like LiDAR data visualization. 

Recently, Octree-GS~\cite{ren2024octree} proposes representing Gaussians within an octree structure to render fine details at different viewing scales. CityGaussian~\cite{liu2024citygaussian} employs an LOD strategy for efficiently training and rendering of large-scale 3DGS. However, these methods lack on-demand rendering schemes and require high-computational devices to load complete scenes for real-time rendering, making them unsuitable for lightweight devices such as tablets or laptops. \textcolor{red}{
% Besides, Hierarchical 3DGS~\cite{hierarchicalgaussians24} presents a hierarchical 3DGS, achieving promising outcomes in rendering extensive walk-through datasets without the need for high-computational devices for real-time rendering. Nevertheless, they do not provide a lightweight viewer that supports rendering on lightweight devices, like Laptops and iPads. 
Hierarchical 3DGS~\cite{hierarchicalgaussians24} delivers promising results in rendering large-scale walk-through datasets without requiring high-performance computational devices. Nonetheless, it lacks a web-based viewer that supports rendering on lightweight devices, such as laptops and iPads.} 
In contrast, our LetsGo method employs a multi-resolution Gaussian representation coupled with an on-demand rendering scheme. This approach achieves rendering speeds four times faster than the original 3DGS and ensures compatibility with lightweight devices, facilitating real-time rendering in large-scale garage scenes.

\section{Garage Data Capture}

% subsection{Data Acquisition}
% 1. polar device
% 2. how to collect data with polar
%  3.2 Msh generation and pre-processing
% 1. calibration
% 2. Mesh generation
% 3.3 3D Gaussian
% 1. preliminary
% 2. drawback of fish-eye image gaussian (data, algorithm, scene scale)
% 3. so we need to process fish-eye img 1-5 (data)
%    hybrid (algorithm)
%    block participation (scene scale)

% Sec. 4 (Liao)
% why two-stage?
% block-wise training
% training details include black participation

\begin{figure*}
    \centering
    \setlength{\abovecaptionskip}{0pt}
\setlength{\belowcaptionskip}{0pt}
    \includegraphics[width=\linewidth]{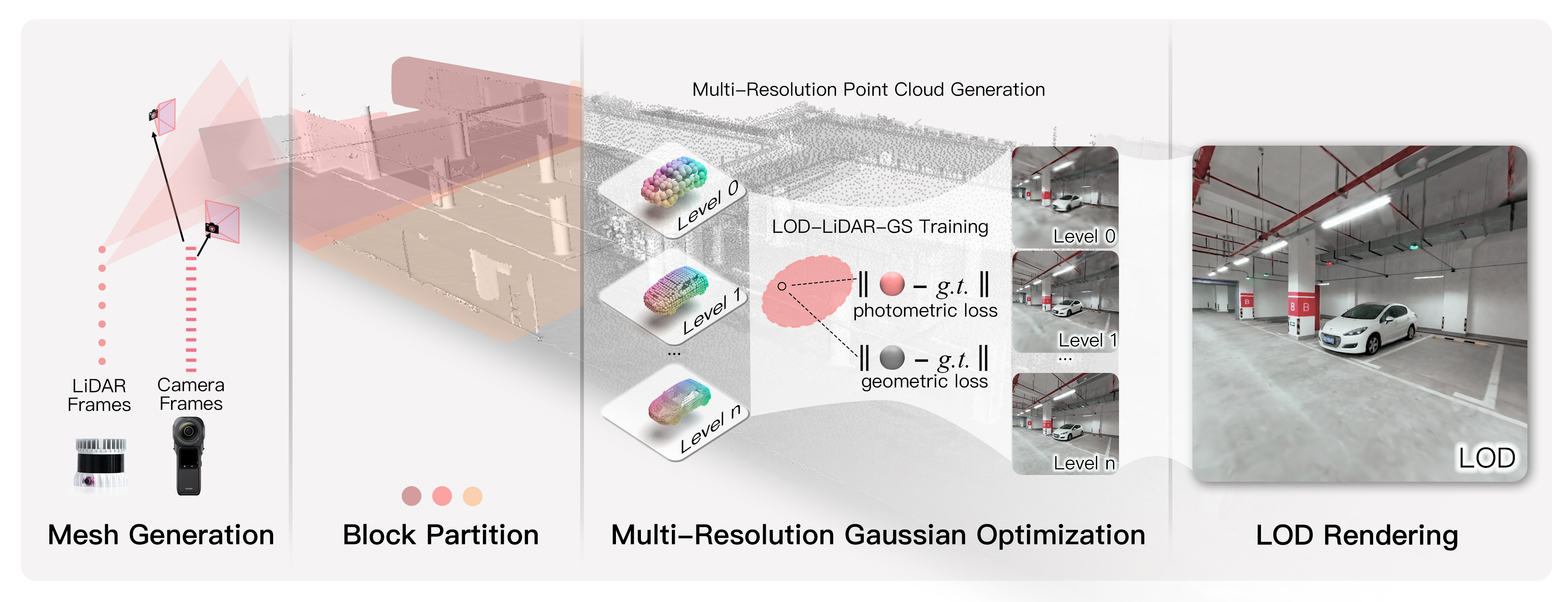}
    \caption{Overview of our LiDAR-assisted Gaussian splatting framework. Initially, we generate a base mesh using color and depth data collected by our self-designed Polar device. The data is then partitioned into blocks for parallel and rapid processing. Next, we downsample the high-quality scanned point clouds into multi-resolution point cloud for our LOD-LiDAR-RGS (Sec.~\ref{sec:lod-lidargs}) method initialization. In addition to photometric supervision, we apply our novel unbiased Gaussian depth regularizer (Sec.~\ref{sec:lidargs}) for geometric supervision. Finally, our system produces photorealistic LOD rendering results based on the optimized multi-resolution Gaussian representation.}
    %看introduction
    %我们首先用我们自己设计的Polar device采集的color和depth数据生成一个base mesh。Then , the captured data is partitioned into blocks for 并行地快速处理。Next， 我们使用我们扫描到地高质量点云用作我们LiDAR-GS和LiDAR-RGS的初始化。并且，除了photometric的监督， 我们还使用我们design的无偏高斯depth计算方法进行geometric监督。最后， 我们photorealistic的渲染结果可以支持一系列application包括Real-time Localization and Navigation， Autonomous Vehicle Parking和VFX Production。
    \label{fig:pipeline}
    \vspace{-2mm}
\end{figure*}

\subsection{Raw Data Acquisition} % 采集设备及方式
% -	为什么需要Polar采集设备，颜色单一，LIDAR，IMU
% Scanning and modeling a large garage is a complex task due to various technical challenges. 
% First, the color scheme in a garage is usually monotonous, with solid-colored walls and floors. The colors of parked vehicles lack texture. The paint surface on the vehicles might be reflective, while the glass is transparent. This makes it difficult to extract enough feature points and establish accurate feature matches from color images, complicating the estimation of the camera pose based on image data alone.
% Second, while LiDAR sensor can provide detailed geometric information, the resulting point cloud lacks color. This means we can't use LiDAR output for high-fidelity rendering. Additionally, large garages, especially underground garages, cannot receive accurate GPS signals, preventing us from using GPS as prior information for reconstruction.
% To tackle these challenges, we have designed a lightweight handheld scanning device called 'Polar.' Polar comprises color fisheye camera, LIDAR, and IMU sensor hardware modules, which collect both color and geometric information in the garage. Fig.\ref{fig:device} illustrates the design of Polar.

Scanning and modeling a large garage is a non-trivial task. Underground and indoor garages often face challenges in receiving GPS signals due to the physical barriers presented by the structures and materials surrounding them, making camera pose estimation for scanning and modeling difficult. 
Furthermore, there are always large-scale texture-less regions inside a garage, \eg, floors and walls. The parked vehicles often contain transparent glasses, and their surfaces are sometimes reflective. This complicates feature matching between images for camera pose estimation and 3D geometry reconstruction. Using a LiDAR sensor for scanning and modeling can provide detailed geometric information. However, the RGB color for each scanned point is not associated. 

% \hfill

\textit{Capturing Device.}
To address these problems, we design a lightweight handheld scanning device named ``Polar'' to jointly collect color and geometric information of the garage. 
The data collection unit of the Polar device comprises a color fisheye camera, a LiDAR sensor, and an IMU sensor, as visualized in Fig.~\ref{fig:device}.
The fisheye camera captures RGB color information in 30 FPS. It has a resolution of 6K and a field of view (FOV) of $180\times180$ degrees, allowing for quick and comprehensive recording of color data.
The LiDAR sensor collects 3D point clouds, recording geometric information at a rate of 2.6 million points per second. With a measurement accuracy of 1 to 1.5 cm and a maximum detection distance of 50 m, it is ideal for 3D scanning of large garage scenes. 
The IMU sensor provides acceleration information on the device's motion, enabling more accurate pose estimation. 
In addition to the data acquisition unit, we also equip the Polar device with a data processing unit consisting of a mini PC for real-time SLAM calculations. The data processing unit is powered by two removable batteries that provide more than 30 minutes of single scan endurance. 
With this unit, one can use a smartphone to connect with the Polar device and preview 3D point cloud reconstruction results in real-time. More information about Polar device along with the calibration process is provided in Appendix.~\ref{app_polar_specs}.
% Additionally, the Polar scanner can use a smartphone to preview 3D point cloud reconstruction results in real-time. 
% The entire Polar device weighs no more than 1kg, making it suitable for handheld or wearable scanning applications. 

% -	我们怎么使用Polar采集车库

% \footnote

\textit{Scanning Scheme.} We use the Polar device to scan various garages for modeling and free-viewpoint rendering. The scanning trajectory for each garage is meticulously designed to emulate drive-through and parking. We collect data along each trajectory four times: one for forward-facing, one for backward-facing, one for side-left, and one for side-right, ensuring comprehensive capturing of the garage. 
We set the camera to auto exposure and auto ISO to accommodate complex lighting changes in the garage. 
For garages that are partially open-air and partially covered, the data is collected when the sunlight is weak, \ie, during the early morning or evening. This ensures 
the images captured in the transition between open-air and covered areas share similar illumination. 
We also apply a short pause for data capturing at the transition area, allowing the sensor to adapt to different lighting conditions and thus ensuring the images are neither overexposed nor underexposed. 
For garages with motion-activated lights, we ensure the data is collected after the light turns on, maintaining consistent lighting conditions across all images. 
The travel speed for data collection is at around 1.0±0.2 m/s, with a turning speed of 15 °±3 °/s. 
To avoid motion blur in areas with insufficient light, these parameters are reduced to 0.5 m/s and 10 °/s, respectively. 
Some garages are very large, for example, over 30,000 square meters. 
Thus, we divide large garage\textcolor{orange}{s} into small subsections and collect data for each subsection. The data is fused later after collection. 
\subsection{GarageWorld Dataset}

% \begin{table*}[ht]
%     \centering
%     %\resizebox{\textwidth}{!}{
%     \begin{tabular}{llllll}
%     \toprule
%     Dataset & Campus1 & Campus2 & Super Regional Mall & Box Retailers&  Business Park \\ 
%     \midrule
%     Category & \multicolumn{2}{c}{Underground Parking Garage} & Surface Parking Lot & Multi-Story Parking Garage & Mechanical Parking System \\ 
%     Area($m^2$) & 38447.86 & 28046.37 & 13495.92 & 32646.68 & 22,159.25\\
%     Image Num & 8479 & 7772 & 2280 & 5792 & 9308\\
%     Point Num & 1.9B & 1.4B & 0.6B & 1.14B & 1.15B\\
%     Face Num & 121.9M & 95.8M & 77.1M & 40.78M & 72.3M \\
%     Lighting Source & \multicolumn{2}{c}{LED Daylight Ceiling Light}&  Sunlight & LED light \& Sunlight & Motion-activated lighting \\
%     Traffic & \multicolumn{2}{c}{Pedestrians and Few Cars }& None & Lots of Cars & Few Pedestrians and Cars\\
%     Time Cost & $\sim$90min & $\sim$90min & $\sim$40min & $\sim$200min & $\sim$140min\\
%     \bottomrule
%     \end{tabular}
%     \caption{Dataset Overview(Double Column)}
%     \label{tab:Dataset}
% \end{table*}

% Large garages can be categorized by how vehicles move between floors. These include garages with internal ramps, internal circular or spiral paths, external ramps, vehicle elevators, and robotic automatic parking systems, etc. 
% Todo

We collect data for eight garages\footnote{All garage data is captured with the necessary permits.}, including six underground garages with various types of inside geometries, one indoor garage with multi-floors at Shopping Mall One, and one outdoor surface parking at Shopping Mall Two. % and one indoor garage with a mechanical parking system in an Office Building. 
These garages comprise various challenging structures, such as sloped surfaces with distorted lines, internal circular or spiral paths, vehicle elevators, \etc, as shown in Fig.~\ref{fig:dataset_gallery}.
% Almost all our scanned garages contain electric vehicle charging stations, making them eco-friendly. 
Tab.~\ref{tab:Dataset1} provides an overall description of the garages. 

\textit{Underground Garages. } 
% We first collect data for two underground garages on campus. 
Most of our captured garages are located underground. This architectural choice is prevalent in regions where above-ground space is scarce, allowing the conservation of surface area for alternative applications. Our underground garages on campus feature a single parking level, characterized by flat and sloped surfaces, and are lit by regular fluorescent light tubes affixed to the ceilings. In contrast, the garages at Shopping Mall Three and the Arts Center have vibrant design elements and superior lighting conditions. Additionally, we conduct a scan of a compact indoor garage within an office building, equipped with a mechanical parking system and featuring colored surfaces.

% The coverage of the two garages is around 38.4k and 28k square meters, respectively, and they only have one floor for vehicle parking with flat and sloped surfaces.
% We collect 8,479 and 7,772 images for the two garages, respectively, with 1.9 and 1.4 billion point clouds. 

%  \textit{Indoor Garage with a Mechanical Parking System.} 
% We also scan several indoor garages with a mechanical parking system in an office building. 
% These garages are typically small in size and contain colored surfaces. 
% The total coverage of these garages is around 22k square meters, containing 9,308 images and 1.15 billion points in point clouds. 

\textit{Indoor Garage with Multi-floor. }
Staking garages to multi-floor is a common design in dense urban environments. 
We collect data for this type of parking garage in Shopping Mall One. 
The different floors of this garage are connected by spiral and circular paths, which have a semi-open structure and are partially illuminated by sunlight and partially by indoor lights. 
Given that the lights in the shopping mall are not turned on in the early morning, we collect data for this garage during the early evening to maintain consistent illumination between different images. 
% The total coverage for this garage data is over 32k square meters, with 5,792 images and a point cloud containing 1.14 billion points.

\textit{Outdoor Parking.}
The outdoor surface parking facility is often in areas with large spaces or on the top of a commercial building. 
We collect data for a garage with this type located on the top of Shopping Mall Two. 
The entrance to this parking space is from indoor to outdoor, containing spiral and sloped paths. 
During daylight time, the outdoor illumination is stronger than indoor.
At night, the limited dim streetlights are insufficient for photography requirements. 
Therefore, we conduct our data capture during the early morning and evening when the sunlight is soft, and the illumination between indoors and outdoors is similar. 
\subsection{Initial Mesh Reconstruction}\label{mesh_gen}

We employ the off-the-shell LiDAR-Inertial-Visual (LIV) SLAM~\cite{shan2021lvi} to estimate the relative poses of the sensor between different time steps. The LIV-SLAM system integrates a tightly coupled LiDAR-Inertial Odometry and Visual-Inertial Odometry, along with a joint optimization approach between LiDAR and camera data, for relative pose estimation. The pose estimation system leverages the unique capabilities of each sensor of our Polar device to enhance overall accuracy and robustness. The IMU delivers reliable short-term motion estimates, while the LiDAR contributes precise distance measurements. Visual sensors complement these by enriching pose estimation in environments abundant with visual features, ensuring a robust and precise outcome.
% Specifically, we employ LIO to fuse IMU data with point clouds collected at different time steps by the LiDAR sensor. This process yields accurate distance measurements and effectively mitigates drift in the IMU data. Subsequently, using the IMU data updated by LIO, we implement VIO among captured fisheye images to estimate their relative poses. 
% Following this, we utilize joint factor graph optimization~\cite{lin2022r} to refine the estimated poses of the LiDAR and camera. The initial pose graph is constructed by incorporating the updated LiDAR, IMU, and camera poses. Then, bundle adjustment is applied to further optimize the graph.

% \paragraph{Mesh Generation}
After merging the point cloud data collected at different time steps using the estimated relative pose, we apply Poisson Reconstruction~\cite{kazhdan2006poisson} to convert the point cloud data into a mesh. To ensure the accuracy and integrity of the resulting mesh, we also compare the reconstructed \textcolor{red}{mesh} to the original point cloud data to remove incorrect faces (More details in Appendix.~\ref{app_data_clean}). 
% we address any inconsistencies by selectively removing incorrect faces. This rectification process is guided by a comparison between the reconstructed mesh and the original point cloud data.
% Next, we merge the point cloud data collected at different time steps using the estimated relative pose. Subsequently, Poisson Reconstruction \cite{kazhdan2006poisson} is applied to convert the point cloud data into an uncolored mesh. To ensure the accuracy and integrity of the resulting mesh, we address any inconsistencies by selectively removing incorrect faces. This rectification process is guided by a comparison between the reconstructed mesh and the original point cloud data. 
This geometry-based mesh reconstruction method performs well for Lambertian surface reconstruction. However, it faces challenges for reflective and transparent surfaces, \eg, vehicle glass windows, which is especially common in garage environments. 
\section{LiDAR-Assisted LoD Gaussian}
% Vanilla 3DGS use sfm results as the input, which doesn't work on garage data

3D Gaussian representation excels at modeling transparent and reflective surfaces compared to \textcolor{orange}{a mesh representation}. However, it requires a sparse 3D point cloud of the scene and accurate camera parameters obtained from SfM. 
As discussed previously, the large-scale garage scenes are challenging for SfM algorithms, with low lighting conditions, large texture-less regions, repetitive patterns, \etc, making sufficient feature correspondences between different images hard to establish. 
As a result, the capacity of the original 3DGS for high-quality modeling and rendering is limited. 

Our Polar scanner, equipped with calibrated IMU, LiDAR, and fisheye camera sensors, effectively addresses this challenge. By integrating the unique strengths of these diverse sensors, our system achieves precise camera localization and detailed 3D point cloud generation.
In the following sections, we first delve into the technical details of our novel LiDAR-assisted Gaussian splatting technique (Sec.~\ref{sec:lidargs}), which incorporates a custom depth regularizer. By leveraging geometry priors derived from our RGBD scans, this method enhances the realism of rendered images and minimizes the occurrence of floater artifacts.
Subsequently, we introduce an innovative framework incorporating Level-of-Detail (LOD) technology into our Gaussian training and rendering process (Sec.~\ref{sec:lod-lidargs}). This framework utilizes a multi-resolution point cloud as its input, significantly boosting rendering speeds. Moreover, we show how it can facilitate rendering expansive garage environments on lightweight devices through a specialized web-based renderer (Sec.~\ref{sec:LOD_rendering}).
% , demonstrating the framework's capability to handle large-scale scenes efficiently.
Fig.~\ref{fig:pipeline} provides an overview of our LiDAR-assisted Gaussian splatting framework. 

\subsection{Gaussian Splatting with LiDAR Inputs}
\label{sec:lidargs}

% We briefly recap 3D Gaussian splatting first. 
3DGS~\cite{kerbl3Dgaussians} represents 3D points with 3D Gaussians, parameterized by position $\mu$, opacity $\alpha$, anisotropic covariance $\Sigma$, and spherical harmonic (SH) coefficients representing view-dependent color $c$. The projection from 3D Gaussians to 2D images~\cite{zwicker2001surface} is given by 
\begin{equation}
    \Sigma' = JW\Sigma W^T J^T,
\end{equation}
where J represents the Jacobian of the affine approximation of the projective transformation, and $W$ corresponds to the viewing transformation.

The \textcolor{orange}{goal} is to optimize the Gaussian parameters so that the rendered images from the 3D Gaussians are as close to their ground truth (GT) images as possible. 
% \begin{equation}
%     \mathcal{L}_{\mathrm{rgb}} = (1-\lambda) \mathcal{L}_1 + \lambda \mathcal{L}_\text{D-SSIM}, 
% \end{equation}
% where $\mathcal{L}_1$ denote the $L_1$ distance between the rendered and GT images, and $\mathcal{L}_\text{D-SSIM}$ denote the structured similarity between the images. 

% 通过应用旋转矩阵和缩放矩阵，我们可以将标准三维高斯球体优雅地转换为空间中任意位置和形状的椭球体。同时，为了保证优化过程中协方差矩阵的半正定性
\textcolor{orange}{A standard three-dimensional Gaussian sphere can be transformed into an ellipsoid of arbitrary shape within space by applying a covariance matrix $\Sigma$. Additionally, so as to ensure the semi-definiteness of the covariance matrix during the optimization process, }
Kerbl \etal proposes to optimize a scaling matrix $S$ and rotation matrix $R$, and compute the covariance matrix as:
\begin{equation}
    \Sigma = RSS^T R^T,
\end{equation}
where $S$ is parameterized as a 3D vector and $R$ is parameterized as quaternion, a 4D vector with unit norm. 

% In this paper,
Since conventional SfM approaches fail in large garage environments due to hard correspondence establishment between textureless and transparent regions, we use our Polar device with a LiDAR sensor to scan point clouds of the garages. 
Considering the originally scanned point clouds contain \textcolor{orange}{ noise}, we resample a set of new points from the reconstructed mesh with a uniform sampling strategy. 
These resampled points, in conjunction with the camera parameters, are used to train the 3DGS representations.

In addition to the image reconstruction loss in the original Gaussian splatting, we further introduce a depth-regularizer for our LiDAR-assisted 3DGS training, which incorporates depth priors derived from the high-quality LiDAR data during training. 
We denote this method as LiDAR-GS and the original 3DGS method with LiDAR-assisted point cloud for Gaussian initialization as 3DGS$^*$. 

\hfill

\textit{Depth Regularizer.} 
Inspired by the depth calculation from NeRF \cite{mildenhall2021nerf}, we utilize the rasterization pipeline of Gaussians to compute the depth of each Gaussian primitive:

\begin{equation}
% \begin{split}
D_\text{G}=\sum_{i \in N} d_i \alpha_i T_i, \quad
\quad T_i=\prod_{j=1}^{i-1}\left(1-\alpha_j\right),
% \end{split}
\end{equation}
% XXX to be fixed
where $D_\text{G}$ is the rendered depth 
% of the 3D Gaussian primitive 
and $d_i$ is the depth of each Gaussian splat in camera perspective. % seen by the camera

It should be noted that the center of the Gaussian is not directly employed for depth computation. Due to variations in the shape and orientation of the Gaussian, the depth at the precise point where the ray intersects the Gaussian deviates from the depth at its center. 
% As shown in Fig.\ref{fig:depth}, our approach computes the expected depth at the specific point of intersection with the Gaussian as follows:
Our approach computes the expected depth at the specific point of intersection with the Gaussian as follows:

% Note that, we don't simply use the center position of the Gaussian for depth calculation. Due to the differences in the shape and orientation of the Gaussian, the depth of the actual intersection of the ray and the Gaussian will be different from the depth of the center. Therefore, we are calculating the expected depth by the ray hitting the Gaussian at the location. It is calculated as:
\begin{equation}
d_i=\frac{1}{l}\left(p_2-\frac{\left(\boldsymbol{\mathbf{\Sigma}}^{-1}\right)_{0,2}}{\left(\boldsymbol{\mathbf{\Sigma}}^{-1}\right)_{2,2}}\left(x_0-p_0\right)-\frac{\left(\boldsymbol{\mathbf{\Sigma}}^{-1}\right)_{1,2}}{\left(\boldsymbol{\mathbf{\Sigma}}^{-1}\right)_{2,2}}\left(x_1-p_1\right)\right)
\label{eq6}
\end{equation}
where $\mathbf{p} = [p_0, p_1, p_2] $ represents the position of the Gaussian center in the ray space, and $\mathbf{\Sigma}$ is the $3 \times 3$ covariance matrix. $\mathbf{(\cdot)}_{m,n}$ represents the corresponding element in the matrix. For detailed derivation and understanding of this process, please refer to Appendix.~\ref{app_depth}.

Given the $K$ captured views, we compute the depth loss using the following equation:
\begin{equation}
\mathcal{L}_{\mathrm{depth}}=\sum_{k \in K}\left\|{D}^{k}_{G}-D^k\right\|_1,
\end{equation}
where $D^k$ represents the ${k^{\text{th}}}$ inherent depth prior from LiDAR data.

The total loss function is as follows:
\begin{equation}
\mathcal{L}_{\mathrm{total}}= \mathcal{L}_{\mathrm{rgb}}+\lambda_\mathrm{depth} \mathcal{L}_{\mathrm{depth}}, 
\end{equation}

where $\mathcal{L}_{\mathrm{rgb}}$ is the RGB image reconstruction loss, following the original 3DGS, and $\lambda_\mathrm{depth}$ is the weight for our depth term.
% We further introduce a novel method called LiDAR Gaussian splatting (LiDAR-GS). This approach harnesses the high-fidelity point cloud data acquired via LiDAR, capitalizing on the inherent depth priors. On the contrary, 3DGS* only employs the point cloud data during the initialization of the Gaussian kernel, without constraining the generation of the Gaussian distribution throughout the training process. Our in-depth analysis reveals that these depth priors can significantly guide the optimization of the Gaussian distribution during training.
By incorporating the depth constraint, our LiDAR-GS effectively minimizes the occurrence of floating artifacts and aligns the Gaussian kernel more closely with the depth information inherent in the LiDAR data.

% \subsection{Level-of-Detail for LiDAR-GS}  
% \subsection{Training and Rendering with LoD}
\subsection{Multi-Resolution Representation}
\label{sec:lod-lidargs}
The original 3D Gaussian representation requires a vast number of 3D Gaussians, making it resource-intensive and inefficient for lightweight devices. 
% The original 3D Gaussian representation relies on an extensive number of 3D Gaussians, which poses challenges for rendering on lightweight computing devices due to substantial resource requirements.
However, simultaneously loading all Gaussians of the entire scene is inefficient and unnecessary for rendering a specific view. 

Inspired by advanced rendering techniques~\cite{schutz2016potree} for massive LiDAR point clouds, we introduce a multi-resolution Gaussian framework called LOD-LiDAR-GS. This framework integrates the Level-of-Detail (LOD) rendering to the LiDAR-GS, making it suitable for various devices. 
We represent the 3D scene with Gaussians at different resolution levels. 
\textcolor{red}{When initializing the multi-resolution Gaussians, each Gaussian is assigned an LOD attribute, alongside its original attributes like 3D position, spherical harmonic coefficients, and opacity, to support our LOD rendering.} The LOD value dynamically adjusts the cloning and splitting threshold during training. This multi-resolution representation enables fast and lightweight rendering on web-based devices.

\begin{figure*}
    \centering
    \setlength{\abovecaptionskip}{0pt}
\setlength{\belowcaptionskip}{0pt}
    \includegraphics[width=0.92\linewidth]{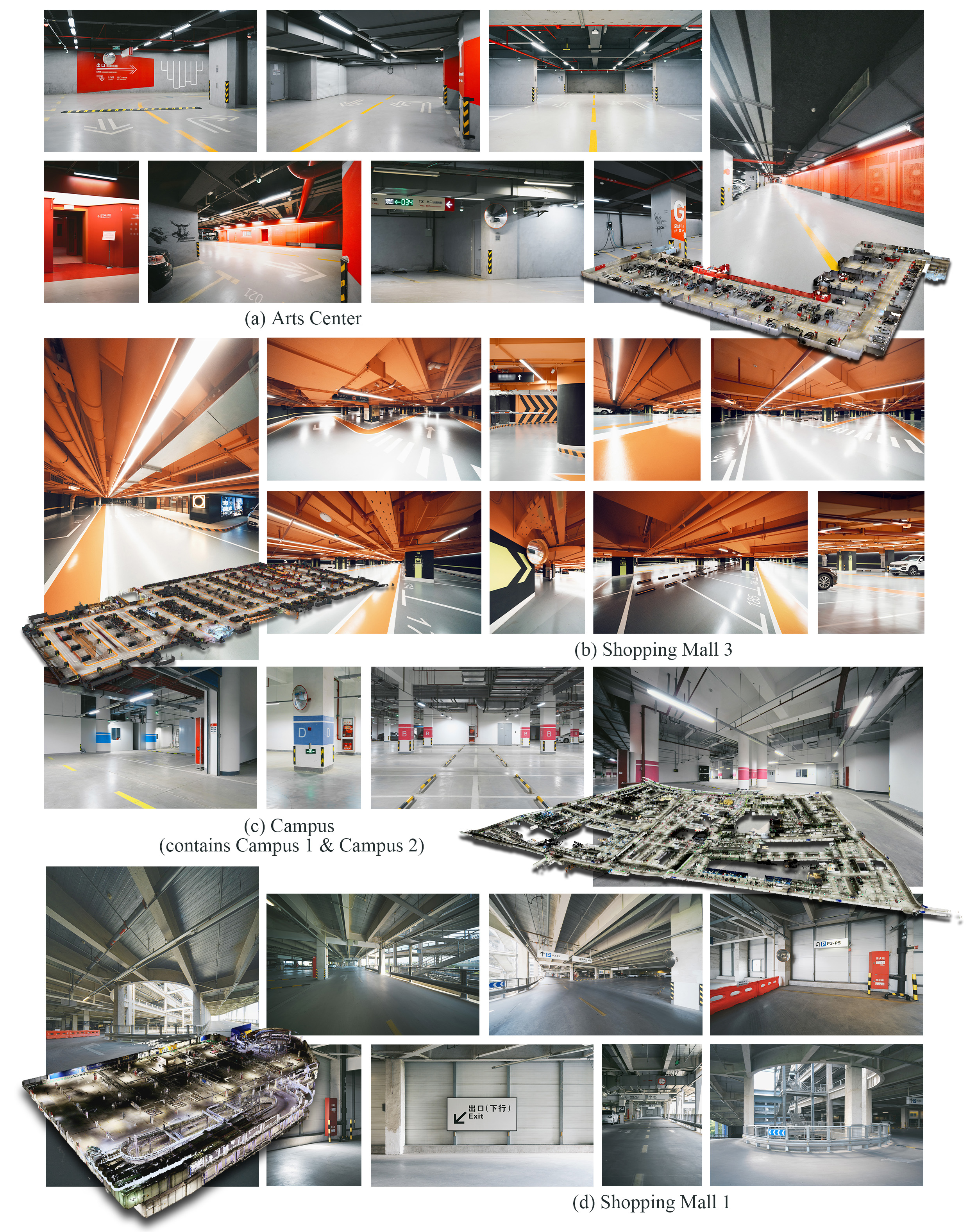}
    % \caption{Our GarageWorld Dataset consists of eight large-scale garages that exhibit a diverse array of types: underground garages, outdoor parking, and an indoor garage with multi-floor.}
    \caption{Visualization of examples from our GarageWorld dataset.}
    \label{fig:dataset_gallery}
\end{figure*}

\paragraph{Multi-Resolution Gaussian Initialization.} 
We first construct multi-resolution point cloud data for Gaussian initialization. 
Our Polar device generates dense point clouds with superior accuracy and consistency, even in challenging garage conditions. 
We use a spacing attribute, $\tau$, to represent the resolution of a point cloud, defined as the minimal distance between points. 
The finest resolution corresponds to the original point cloud sampled from our reconstructed mesh at the lowest $\tau$. We then increase $\tau$ to $2\tau$ and downsample the point cloud to derive a coarser point cloud, repeating until the point count drops below a threshold, $\epsilon_p$. Points in the finest level get an LOD value of $N = L-1$, and those in the coarsest level get an LOD value of $N = 0$. We set $\tau$ to 4.0 cm and $\epsilon_p$ to 10,000 points. The downsampling operation employs an approximate Poisson-disk sampling method from the PotreeConverter library~\cite{schutz2016potree}.

\begin{figure}
\centering
\setlength{\abovecaptionskip}{0pt}
\setlength{\belowcaptionskip}{0pt}
\includegraphics[width=\linewidth]{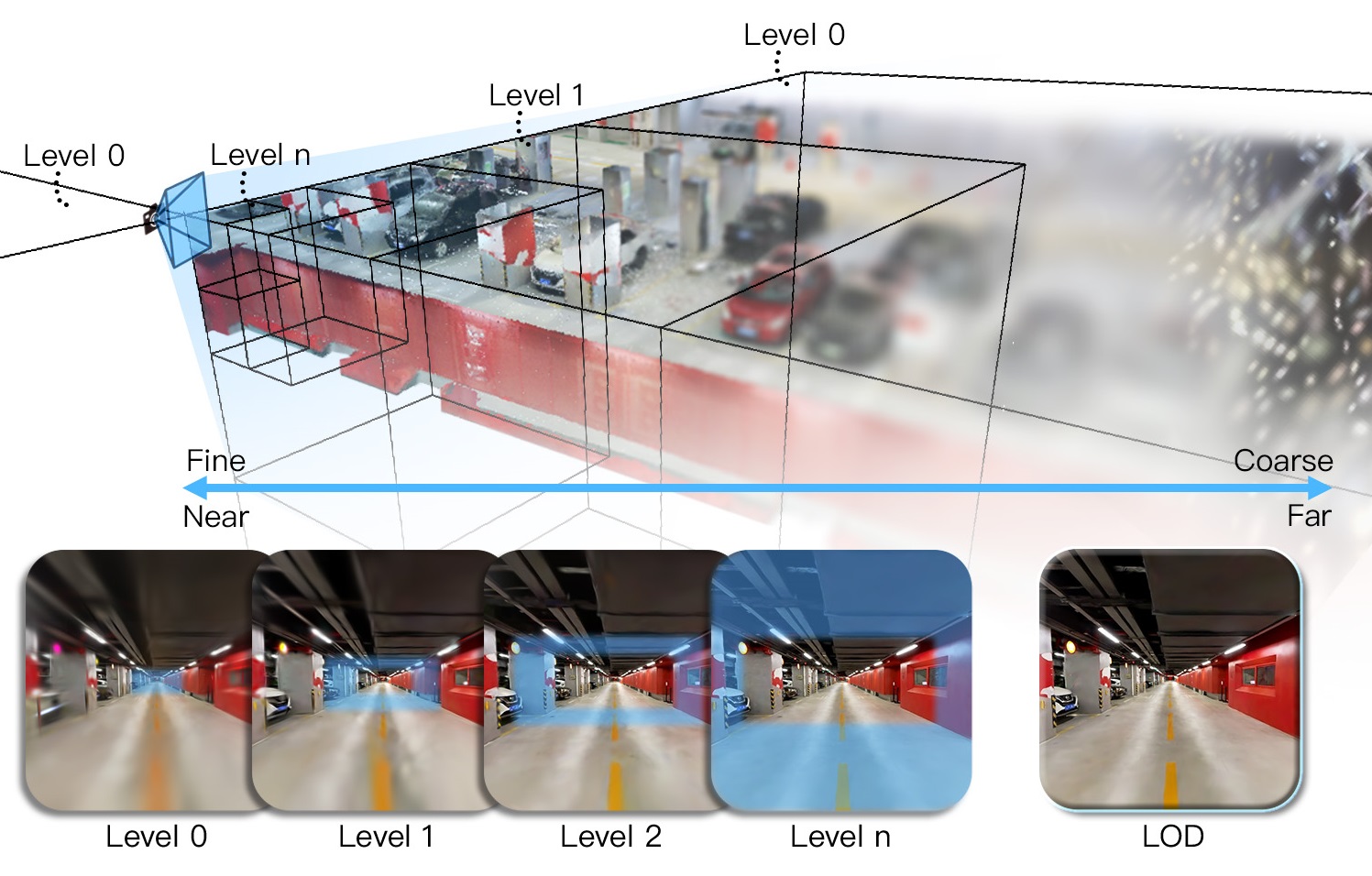} 
\caption{\textcolor{red}{Illustration} of our LOD-LiDAR-GS within an Octree Structure, accompanied by an LOD rendering strategy. Distance varies from near to far, with Gaussian level resolution transitioning from fine to coarse. The bottom row illustrates the level composition in the rendered view.}
\label{fig:LOD-rendering}
\vspace{-2mm}
\end{figure}

\begin{algorithm}[t]
	\caption{Octree Construction for LOD-LiDAR-GS}
	\DontPrintSemicolon
	\SetAlgoLined
	\SetKwInOut{Input}{Input}
	\SetKwInOut{Output}{Output}
	
	\Input{A set of multi-resolution Gaussians stored in .ply files}
	\Output{Gaussian in Octree structure stored in .bin files}
	\BlankLine
	
	\tcp{Initialization} 
	Create an octree with depth $L$\;
	Compute bounding box $B$ that covers all Gaussian levels\;
	Initialize $queue = \{ \text{root of octree}, B, \text{lvl} = 0 \}$\;
	
	\tcp{First loop: Distribute Gaussians into octree nodes} 
	\While{$\text{length}(queue) > 0$}{
		$currentNode \gets queue.pop()$\;
		\ForEach{childNode of currentNode}{
			Assign bounding box $B_{child}$ and $lvl = \{currentNode.lvl\} + 1$ to childNode\;
			queue.push(\{childNode, $B_{child}$, $lvl$\})\;
		}
		Read Gaussians from $level\_\{currentNode.lvl\}.ply$\;
		Distribute Gaussians within $currentNode.B$\;
	}
	
	Initialize $queue = \{ \text{root of octree}, B, \text{lvl} = 0 \}$\;
	$byteOffset, byteSize = 0, 0$\;
	
	\tcp{Second loop: Prune empty nodes and store data}
	\While{$\text{length}(queue) > 0$}{
		$currentNode \gets queue.pop()$\;
		\ForEach{childNode of currentNode}{
			\If{childNode.numPoints > 0}{
				queue.push(\{childNode, childNode.B, lvl\})\;
			}
		}
		$byteSize = \text{size}(currentNode.points)$\;
		Flush $currentNode.points$ into \textbf{octree.bin} file\;
		Flush $currentNode.B$, $currentNode.numPoints$, $byteOffset$,               $byteSize$ into \textbf{hierarchy.bin} file\;
		$byteOffset \gets byteOffset + byteSize$\;
	}
	\label{alg:octree_construction}
        
\end{algorithm}

\paragraph{LOD-LiDAR-GS Training.}
In contrast to LiDAR-GS, our LOD-LiDAR-GS framework processes a multi-resolution point cloud, wherein each resolution level has a distinct LOD value.
We devise a new training strategy to handle the complexity caused by multi-resolution input effectively. 
We initialize a Gaussian model for each resolution level, which operates independently during optimization. 
Following the original 3DGS~\cite{kerbl3Dgaussians}, the decision to clone or split a Gaussian is based on threshold values for position gradients $\sigma_{pos}$ and variance $\sigma_{var}$. We introduce a scaling factor $s$, which varies with the LOD value $l$ of each level, modifying the thresholds as $s \cdot \sigma_{pos}$ and $s \cdot \sigma_{var}$:
\begin{equation}
s_k = \min(\beta_s^{L-1 - l}, s_{max}), \quad l \in [0, L-1]
\end{equation}
where $\beta_s$ and $s_{max}$ are hyperparameters set to $\sqrt{2}$ and $4.0$ respectively.

In our approach, the Gaussian representation at the lower resolution levels is employed to encapsulate the low-frequency content of the scene, while higher resolution levels utilize finer Gaussians to capture the intricate high-frequency details. To use our multi-resolution Gaussian models for image rendering, we adopt a technique analogous to LOD rendering used in traditional computer graphics pipelines.
Gaussian subsets from each resolution level are selected based on the rendering camera's viewing frustum and the projected depth $d$ of each Gaussian. 
The appropriate Gaussian level $L^{(d)}$ is determined as:
\begin{equation}
L^{(d)} = \text{clamp}(\lfloor L ^{1 - d / d_{max}} \rfloor, 0, L-1)
\label{LOD-rendering}
\end{equation}
where $d_{max}$ is the maximum depth value projected onto the training viewpoint from the point cloud. 

During training, we select portions of Gaussians from different resolution levels based on depth ranges to accurately predict the image. We compute the loss by comparing model predictions with ground truth images. Our random-resolution-level (RRL) training strategy renders each training sample with a 50\% probability using either the LOD strategy from multiple levels or a single Gaussian level, preventing overfitting and biases in camera pose distribution.

\subsection{Web-based Lightweight Renderer}

With the proposed multi-resolution Gaussian representation, our LOD-LiDAR-GS significantly enhances rendering efficiency.
Instead of using a large number of 3D Gaussians for each image, LOD-LiDAR-GS dynamically selects Gaussian subsets from different resolution levels based on depth ranges. This reduces the complexity of sorting, projection, and accumulation operations, thereby accelerating rendering speed. On GPU systems such as the NVIDIA RTX 3090, we implement a 3D Gaussian viewer based on the original 3DGS SIBR viewer. By integrating our LOD rendering strategy, as detailed in Equation~\ref{LOD-rendering}, we achieve approximately a 4x increase in rendering speed.
Additionally, \textcolor{red}{we introduce a web-based, on-demand renderer developed with JavaScript and WebGL, which supports real-time rendering on lightweight devices.} Unlike mainstream lightweight Gaussian renderers~\cite{antimatter15_splat,huggingface_gsplat}, which load all Gaussians into VRAM simultaneously, our renderer dynamically loads only the necessary data into memory according to the current viewpoint and LOD strategy. This method avoids the need for full data loading, enabling the rendering of large-scale environments on web platforms.

\paragraph{Converting Multi-Resolution Gaussians into Octree Structure}

Our multi-resolution Gaussian representation is well-suited for storage within an Octree structure, with Gaussians at each level efficiently stored in the nodes at the corresponding depth of the Octree. 
We begin by constructing an Octree of depth $L$, where the root node host level 0 and the leaf nodes contain the finest level, $L-1$. Intermediate levels are stored at the inner nodes of the Octree. 
% During rendering, the necessary data chunks located at these nodes are dynamically loaded from disk based on the viewpoint and the LOD rendering strategy. 
We outline the conversion of our Gaussian representation into the Octree format in Algorithm~\ref{alg:octree_construction}.

\paragraph{Coarse-to-Fine Rendering} \label{sec:LOD_rendering}

Rendering large-scale garage scenes requires managing a substantial number of Gaussians, often exceeding in-memory storage capacity. 
To address this, we implement a coarse-to-fine loading and rendering scheme that facilitates real-time rendering on resource-constrained devices. Initially, low-resolution Gaussian levels are preloaded into memory at viewer initialization. These levels occupy minimal storage and are accessed frequently during subsequent rendering cycles.
As shown in Fig.~\ref{fig:LOD-rendering}, to render high-frequency details, our system dynamically selects high-resolution octree chunks from disk based on the viewer's frustum and the conditions set by our LOD rendering strategy, as outlined in Equation~\ref{LOD-rendering}. 
The rendering process begins with a traversal of the entire octree, determining each node's visibility based on the center position of the node's bounding box. A second traversal then loads visible Gaussian data into a buffer, continuing until either the traversal is complete or the number of loaded Gaussians reaches the maximum capacity. 
The buffer is updated by replacing preloaded Gaussians with newly loaded high-resolution Gaussians, which are then dispatched to a WebGL worker for rendering. 
Recent works, such as OCT-GS~\cite{ren2024octree} and LoG~\cite{LoG}, can also be structured into an octree format. However, OCT-GS fails to represent the entire scene at certain levels, making it unsuitable for our lightweight rendering strategy. Additionally, LoG does not provide a solution for lightweight rendering.

\section{Experimental Results}

\subsection{Training Details}
\label{sec4.1_training}

\begin{figure*}[htbp]
    \centering
    \setlength{\abovecaptionskip}{0pt}
\setlength{\belowcaptionskip}{0pt}
    \includegraphics[width=\linewidth]{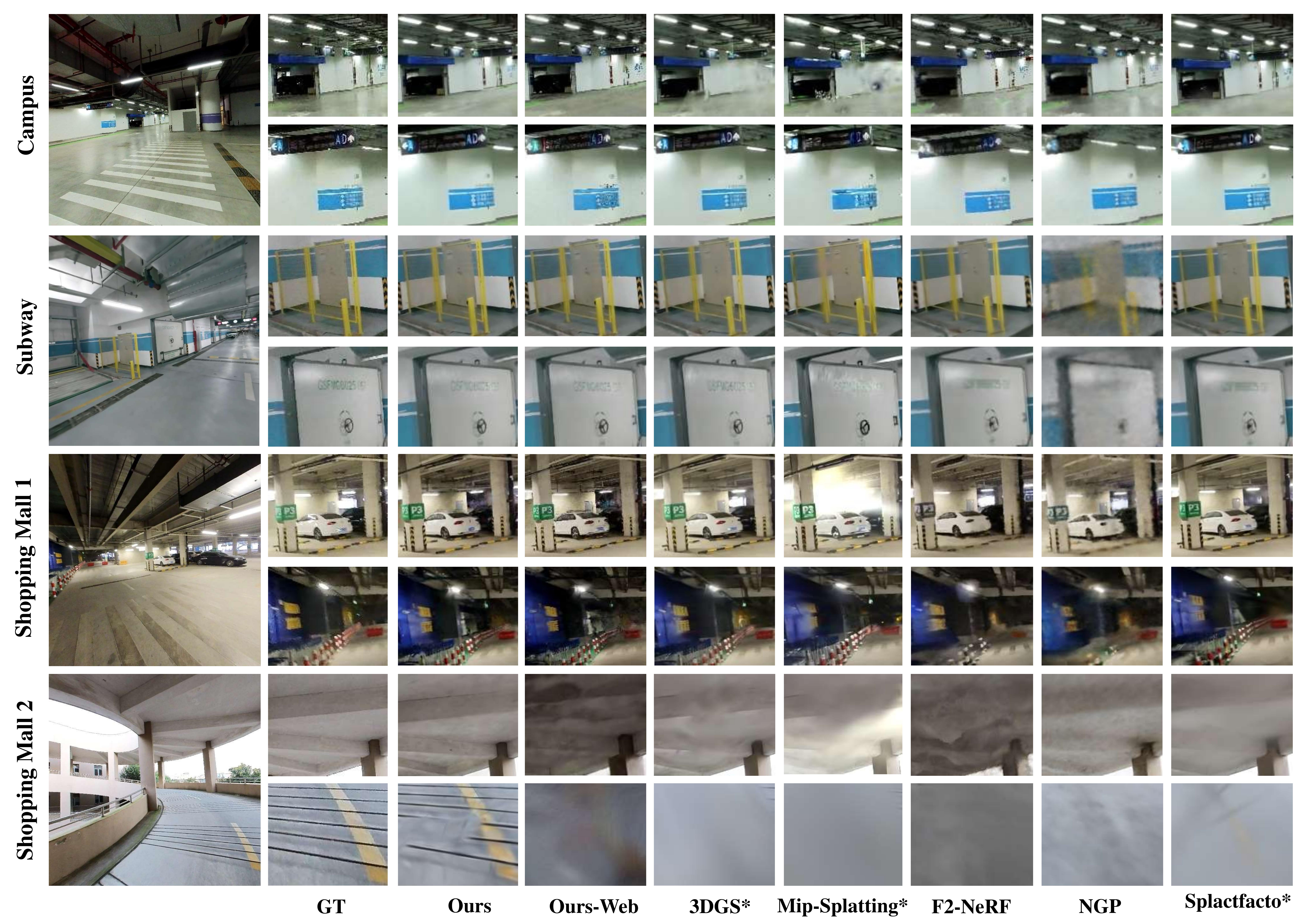}
    \caption{
    Qualitative comparison between our LOD-LiDAR-GS, 3DGS$^*$,  Mip-Splatting$^*$, F2-NeRF, NGP and Splatfacto$^*$ on the various datasets. Here, we use a \textcolor{orange}{superscript} $^*$ to denote their Gaussian primitives are initialized according to our LiDAR-assisted 3D point cloud. 
    % We present a visual comparison of rendering results among our LiDAR-GS method, 3DGS*, SuGaR*, Mip-Splatting*, NGP and F2-NeRF. 
    % The showcased scenes, arranged from top to bottom, include the Shopping Mall 2 from our proprietary datasets, scenes from ScanNet++, and scenes from KITTI-360. Each column corresponds to a different method, and the images are juxtaposed with their corresponding ground truth from held-out test views.
    }
    \label{fig:comparison}
\end{figure*}

\begin{table*}[htbp]
\small
\setlength{\abovecaptionskip}{0pt}
\setlength{\belowcaptionskip}{0pt}
    \caption{
    Quantitative comparison among our LOD-LiDAR-GS, 3DGS$^*$, Mip-Splatting$^*$, F2-NeRF, NGP, Splatfacto$^*$, OCT-GS$^*$, and LoG-GS$^*$ on the various datasets.}
    \label{tab:commands}
    %\resizebox{\textwidth}{!}
    \setlength{\tabcolsep}{4.1mm}{
    \begin{tabular}{lccccccccc}
    
        \toprule
%        Color Annotations: & \multicolumn{3}{c}{ \colorbox{red!25}{\textbf{Best}} \colorbox{orange!25}{SecondBest} } \\
%        \hline
        \multirow{2}{*}{Method} & \multicolumn{3}{c}{GarageWorld}  & \multicolumn{3}{c}{ScanNet++} & \multicolumn{3}{c}{KITTI-360} \\
        \cline{2-10}
          & PSNR$\uparrow$ & SSIM$\uparrow$ & LPIPS↓ & PSNR$\uparrow$ & SSIM$\uparrow$ & LPIPS$\downarrow$ & PSNR$\uparrow$ & SSIM$\uparrow$ & LPIPS$\downarrow$\\
        \midrule
        3DGS*            & {23.52} & \BestCellColor{\textbf{0.822}} & {0.412} & {27.41} & {0.902} & {0.149} & 20.39 & \SecondBestCellColor{0.698} & \SecondBestCellColor{0.289} \\
        Mip-Splatting*    & 22.08 & 0.791 & 0.448 & 25.74 & {0.898} & 0.179 & 20.19 & 0.682 & 0.318 \\
        F2-NeRF          & 18.88 & 0.739 & 0.552 & 23.59 & 0.888 & 0.237 & 18.60 & 0.653 & 0.414 \\
        NGP              & 20.68 & 0.734 & 0.507 & {28.72} & 0.896 & 0.230 & {21.21} & 0.655 & 0.406 \\
        Splatfacto*            & \SecondBestCellColor{25.45} & {0.793} & {0.261} & 28.35 & 0.906 & {0.065} & {14.76} &{0.481} & {0.381} \\
        OCT-GS*            & {25.19} & {0.782} & {0.315} & \BestCellColor{\textbf{31.55}} & \BestCellColor{\textbf{0.941}} & \BestCellColor{\textbf{0.063}} & \SecondBestCellColor{21.52} & 0.692 & 0.322 \\
        LoG-GS*            & {21.53} & {0.737} & {0.278}  & {27.80} & {0.907} & {0.086}  & 18.62 & 0.671 & {0.300}\\
        \midrule
        \textbf{Ours (LOD-LiDAR-GS)} & \BestCellColor{\textbf{25.77}} & \SecondBestCellColor{0.812} & \BestCellColor{\textbf{0.210}}  & \SecondBestCellColor{29.19} & \SecondBestCellColor{0.927} & \SecondBestCellColor{0.064} & \BestCellColor{\textbf{24.53}} & \BestCellColor{\textbf{0.811}} & \BestCellColor{\textbf{0.167}} \\
        \textbf{Ours (Web)} & {22.56} & {0.730} & \SecondBestCellColor{{0.211}}  & {22.79} & {0.824} & {0.140} & {19.20} & {0.586} & {0.247} \\
        \bottomrule
    \end{tabular}}
    \vspace{-2mm}
\end{table*}

% 要有点discussion在details
%train
% 我们利用PyTorch Framework 在 an NVIDIA RTX A6000 GPU进行训练。由于我们场景非常大并且利用到了LiDAR点云作为初始化，我们将缩放学习率降低到0.0015，并将高斯的初始位置学习率设置为0.000016。此外，我们将透明度重置间隔提高到2,000,000，并将高斯密集化步骤的开始延迟到75,000。整个过程的迭代次数是根据捕获的图像数量的20倍经验性设定的。我们设置SH degree为2，正则项的权重 lambda_depth = 0.8。
% We train our model using the PyTorch Framework on NVIDIA RTX A6000 GPUs.
All experiments are conducted on an NVIDIA RTX A6000 GPU using the PyTorch Framework.
Given the large scale of our scenes and the initialization with LiDAR point clouds, we set the scaling learning rate to 0.0015 and the initial position learning rate for Gaussian primitives to 0.000016. We disable the opacity reset option and delay the start of the Gaussian densification step to 75,000 iterations. The total number of iterations is empirically set to twenty times the number of captured images. We use a spherical harmonics (SH) degree of 2 and set $\lambda_\mathrm{depth}$ to 0.8.
\textcolor{red}{We employ a partitioning strategy similar to~\cite{tancik2022block}, expanding each block outward by 30\% to address the poor reconstruction quality on the boundaries.}
%其中，每个场景会分为3~10个block不等，每个block视其点云大小和图像数量，会训练1.5~4小时不等，少数因为地库形状不规则导致特别大的block的训练时间会超过8小时。
\textcolor{orange}{The overall end-to-end duration, including data capture, preprocessing, and training, typically ranges from 12 to 16 hours. Each scene is divided into 3 to 10 blocks, depending on its complexity. The training duration for each block varies from 1.5 to 4 hours, based on the size of the point cloud and the number of images. In a few cases, irregular shapes in underground garages result in particularly large blocks, extending the training time to over 8 hours. }
%实际没用alpha 就不写了
% While it's possible to train on the more memory-rich A100 GPU, please note that, as previously stated in \ref{res-gpu}, that even the A100's memory capacity is insufficient for training such large-scale scenes using 3DGS.
% Considering the large scale of our scene and the use of LiDAR point clouds for initialization, we adjust the scaling learning rate to 0.0015 and set the initial position learning rate for Gaussians primitives to 0.000016. Additionally, we disable the opacity reset option and delay the Gaussian densification step start to 75,000. The total iteration count is empirically set to twenty times the number of captured images. We set the spherical harmonics (SH) degree to 2 and the weights for regularization terms $\lambda_\mathrm{depth}$ and $\lambda^*_\mathrm{depth}$ to 0.8, respectively. 

\paragraph{Training Image Selection and Pre-processing.}
Regarding training view selection for each block, camera views originating within the partition bounds are retained. 
Furthermore, for camera views positioned outside the partition, we project the mesh in the partition block and the original mesh onto each camera view and compute the projected pixel overlap ratio between the quadrilateral-partitioned mesh and the whole mesh. Views with a ratio exceeding a threshold of $0.8$ are retained. 
The original Gaussian framework does not support direct input of fisheye images. To retain more information from the images, we avoided traditional fisheye distortion correction. Instead, we split a single fisheye image into five pinhole images. Specifically, the fisheye image is projected onto a hemispherical surface. Then, five virtual cameras, each oriented at 45 degrees to the top, bottom, left, right, and straight ahead, reproject the hemispherical image into their respective views. This process generates new images and extrinsic parameters.
Additionally, since the downward-facing view includes part of the capturing device, we use a mask to obscure this portion of the image.

\subsection{Comparison} \label{sec:exp}

%引一下 我们用了哪些dataset，标黑
%In this section, 我们与其他方法在一系列有挑战性的large scale dataset进行比较，包括我们的Garage World Dataset， \textbf{KITTI-360}\cite{liao2022kitti}，一个 large outdoor street scene，  \textbf{ScanNet++} \cite{yeshwanth2023scannet++} 一个indoor scene with complex geometry and variable lighting。

% In this section, we demonstrate the capabilities of our method in various challenging large-scale scenarios compared with other methods. We adopt 2 scenes called SIST and SEM from our \textbf{Garage} datasets illustrated above,  from the \textbf{KITTI-360}\cite{liao2022kitti} large outdoor street scene. and \textbf{ScanNet++} \cite{yeshwanth2023scannet++} indoor scene with complex geometry and variable lighting.

\begin{figure}[t]
    \centering
    \includegraphics[width=\linewidth]{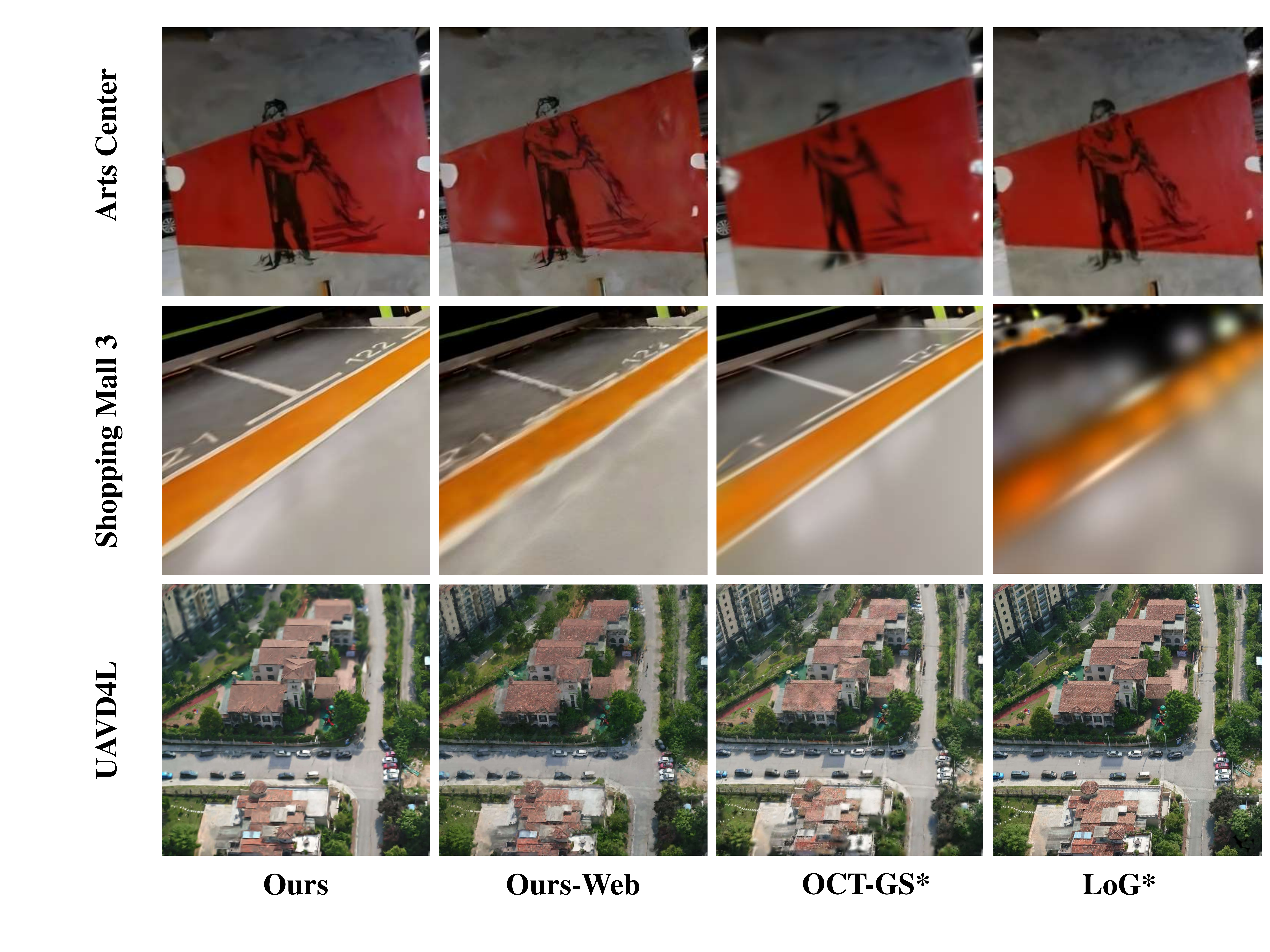}
    \caption{ Qualitative comparison of our method and concurrent works, including recent OCT-GS* and LoG* methods, on the GarageWorld and UAVD4L datasets.}
    \label{fig:Concurent work}
    \vspace{-2mm}
\end{figure}

\begin{figure}[t]
    \centering
    \includegraphics[width=\linewidth]{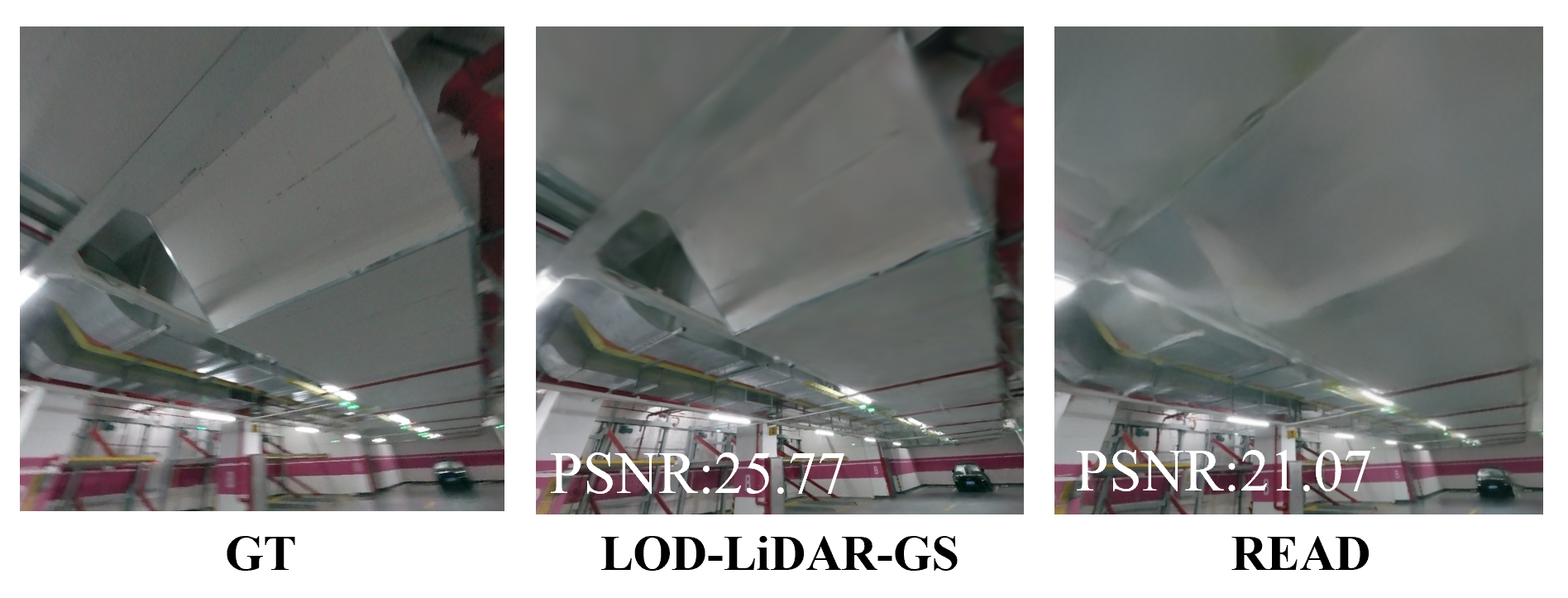}
    \caption{ \textcolor{red} {Qualitative and Quantitative comparison of our method and READ approach on the GarageWorld datasets.}}
    \label{fig:READ}
    \vspace{-2mm}
\end{figure}

\paragraph{Datasets. }
We compare our approach with state-of-the-art methods and concurrent works on various challenging datasets, including our \textbf{GarageWorld}, \textbf{KITTI-360}~\cite{liao2022kitti} (large outdoor street scene), and \textbf{ScanNet++} ~\cite{yeshwanth2023scannet++} (indoor scene characterized by complex geometry and variable lighting). For each dataset, we use 10\% of the images as test sets and the remaining 90\% as training sets.

% In this section, we compare our approach with the state-of-the-art methods and concurrent works on various challenging large-scale datasets, including our \textbf{GarageWorld}, \textbf{KITTI-360} \cite{liao2022kitti}, a large outdoor street scene, and \textbf{ScanNet++} \cite{yeshwanth2023scannet++}, an indoor scene characterized by complex geometry and variable lighting. 
% We adopt 10\% images of each scene as test sets, others are training sets. 
 % This comparative analysis showcases the robustness and versatility of our method across diverse and demanding real-world environments.

\begin{figure}[t]
    \centering
    \includegraphics[width=\linewidth]{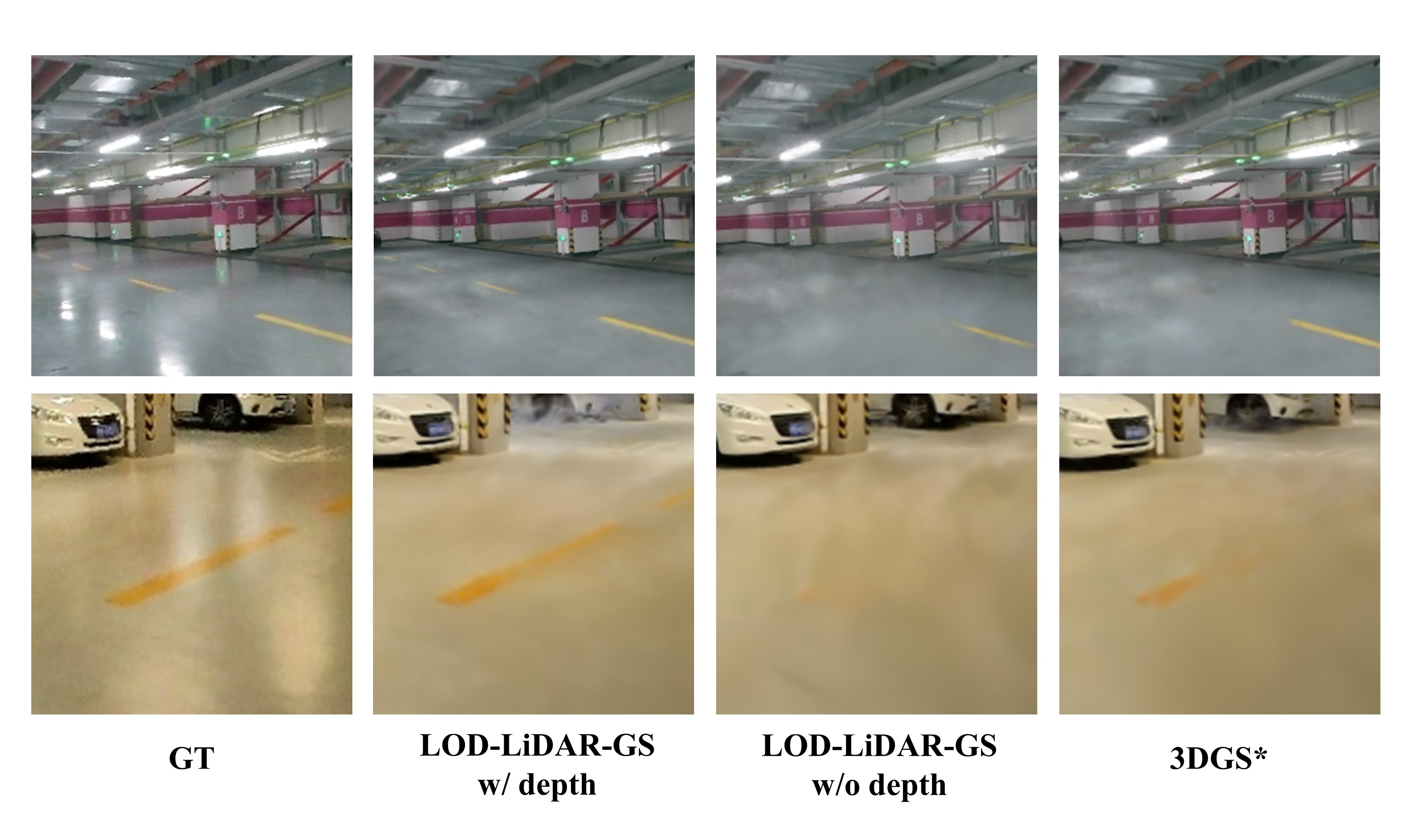}
    \caption{Qualitative ablation study on depth regularization within our large-scale GarageWorld dataset clearly demonstrates improvements in resolving the floater problem on the ground.}
    \label{fig:Ablation_Depth}
    \vspace{-2mm}
\end{figure}

\begin{figure}[t]
    \centering
    \includegraphics[width=\linewidth]{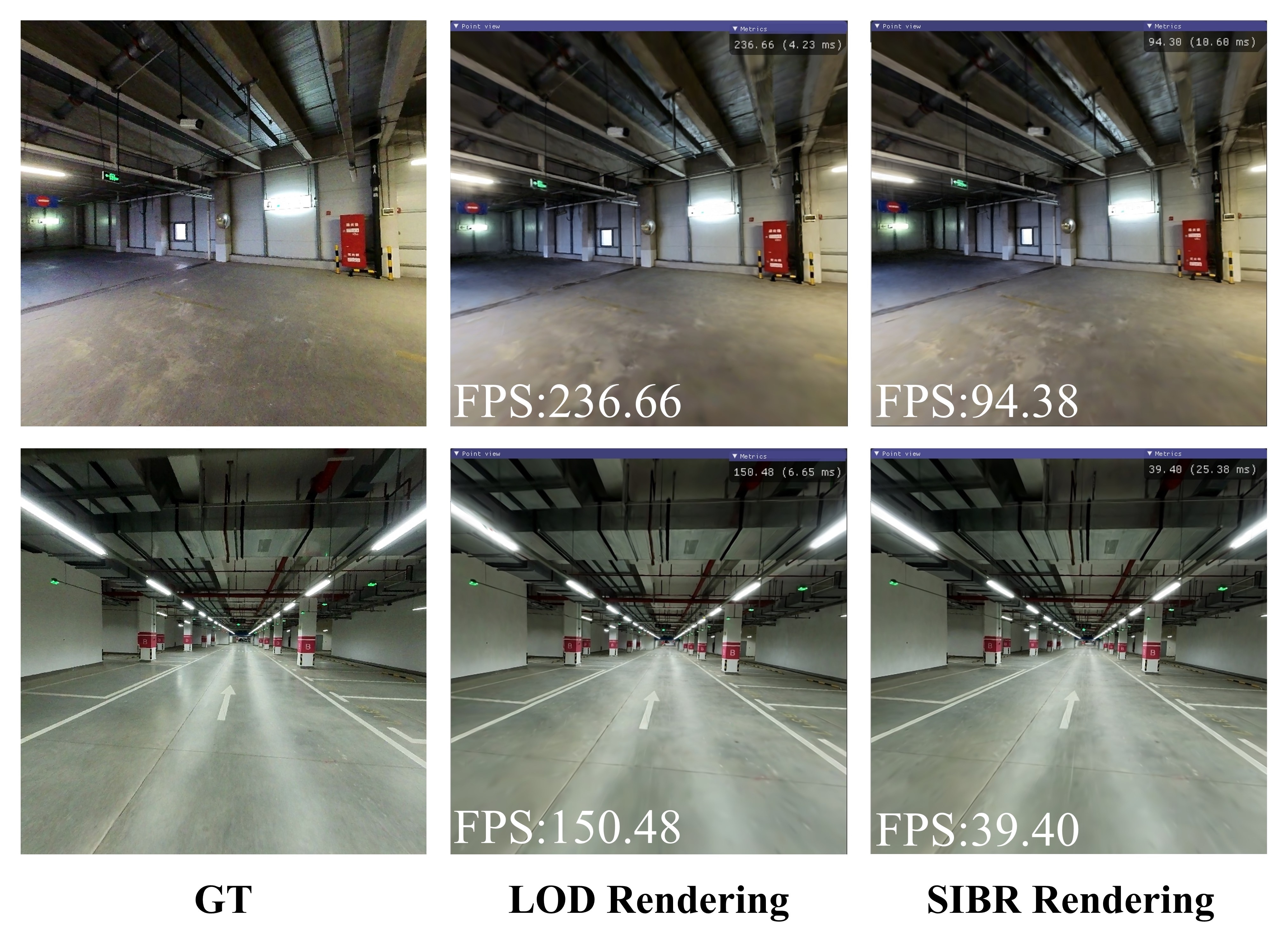}
    \caption{Ablation study on our LOD rendering strategy demonstrates that our real-time LOD renderer operates approximately 3-4 times faster than the original 3DGS SIBR viewer when tested on an NVIDIA RTX 3090 GPU.}
    \label{fig:Ablation_FPS}
    \vspace{-2mm}
\end{figure}

\paragraph{Competing Methods}
We compare our method against a suite of 3DGS approaches, including \textbf{3DGS*}~\cite{kerbl3Dgaussians}, \textbf{Mip-Splatting*}~\cite{yu2024mip}, and \textbf{Splactfacto*}~\cite{nerfstudio}, as well as recent concurrent work, including \textbf{OCT-GS*}~\cite{ren2024octree} and \textbf{LoG*}~\cite{LoG}. Additionally, we benchmark our approach against recent implicit representation methods, namely \textbf{F2-NeRF}~\cite{wang2023f2} and \textbf{Instant-NGP}~\cite{muller2022instant}.
To extend recent 3DGS-based methods to large-scale garage scenes, we initialize them with our LiDAR-derived point clouds instead of SFM point clouds, enabling their application to challenging garage environments. Methods utilizing our LiDAR-derived point clouds are denoted with an asterisk (*).
%我们将我们的方法和a suite of 3DGS方法比，包括3DGS， Mip-Splatting*, SuGaR*。他们都使用我们 LiDAR-derived point clouds作为初始化。我也和最近implicit 表达的方法比，包括f2nerf和 ingp。
%
% To extend recent 3DGS-based methods to large-scale garage scenes, we initialize these methods with our LiDAR-derived point clouds instead of SFM point clouds. 
% This adaptation enables their application to the challenging garage environments. In the following sections, methods utilizing our LiDAR-derived point clouds are denoted with an asterisk ($*$).
% We compare our method against a suite of 3DGS approaches, including \textbf{3DGS*}\cite{kerbl3Dgaussians}, \textbf{Mip-Splatting*}\cite{Yu2023MipSplatting}, and \textbf{Splactfacto*}\cite{nerfstudio}. 
% We also compare our method against recent concurrent work, including \textbf{OCT-GS*}\cite{ren2024octree} and \textbf{LoG*} \cite{LoG}.
% Additionally, our approach is benchmarked against recent implicit representation methods, namely \textbf{F2-NeRF}\cite{wang2023f2} and \textbf{Instant-NGP}\cite{muller2022instant}, 

% %
% They all utilized our LiDAR-derived point clouds as initialization, enabling their application to the challenging garage scenes. 
%

% 我们选取了六组使用上述设备构建的大型地下车库的数据集，以及kitti360室外场景和scannet++室内场景的数据集进行对比，我们平均选取了其中的10%用作测试集。公平起见，我们对每个方法都迭代了60000次，并渲染了输入图片的原始分辨率。图？展示了各种方法的qualitative对比，可以看到，3D-GS和sugar的方法通常在镜头前方以及场景边缘产生很多漂浮高斯点，导致了渲染更加模糊以及渲染质量的下降，而对于NGP这样使用hash encoding的方法，在这样的大型场景上的渲染结果则十分模糊。而我们的方法则成功恢复出了包括地面的几何以及纹理，并且由于depth term的作用，空中漂浮的点云被优化掉，得到了更好的渲染质量。
% 对于定量比较，我们选取了psnr，ssim，lpips作为我们评判渲染质量的指标，我们在表？中提供了我们在8组大型场景数据上测试集的指标。我们的方法在这些数据集的指标超过了绝大多数的方法，进一步证明了我们的方法在各种大场景数据集上有优秀的渲染质量。

\paragraph{\textcolor{red}{Quantitative} Comparison Results. }
For quantitative comparisons, we use Peak Signal-to-Noise Ratio (PSNR), Structural Similarity (SSIM), and Learned Perceptual  Image Patch Similarity (LPIPS)~\cite{zhang2018unreasonable} as metrics to evaluate rendering quality. The results of these comparisons are presented in Tab.~\ref{tab:commands}. Our LOD-LiDAR-GS method consistently achieves the best results on large-scale scene datasets, such as our GarageWorld dataset and the KITTI-360 dataset. On \textcolor{orange}{the} small-scale scene dataset ScanNet++, our LOD-LiDAR-GS approach achieves the second-best results. 
From the last two rows of Tab.~\ref{tab:commands}, our web renderer demonstrate comparable performance on lightweight devices (MacBook Air with M2 clip). %{\color{red}{(ATTENTION: plz add, which kind of laptop)}}.

\begin{figure}[t]
    \centering
    \includegraphics[width=\linewidth]{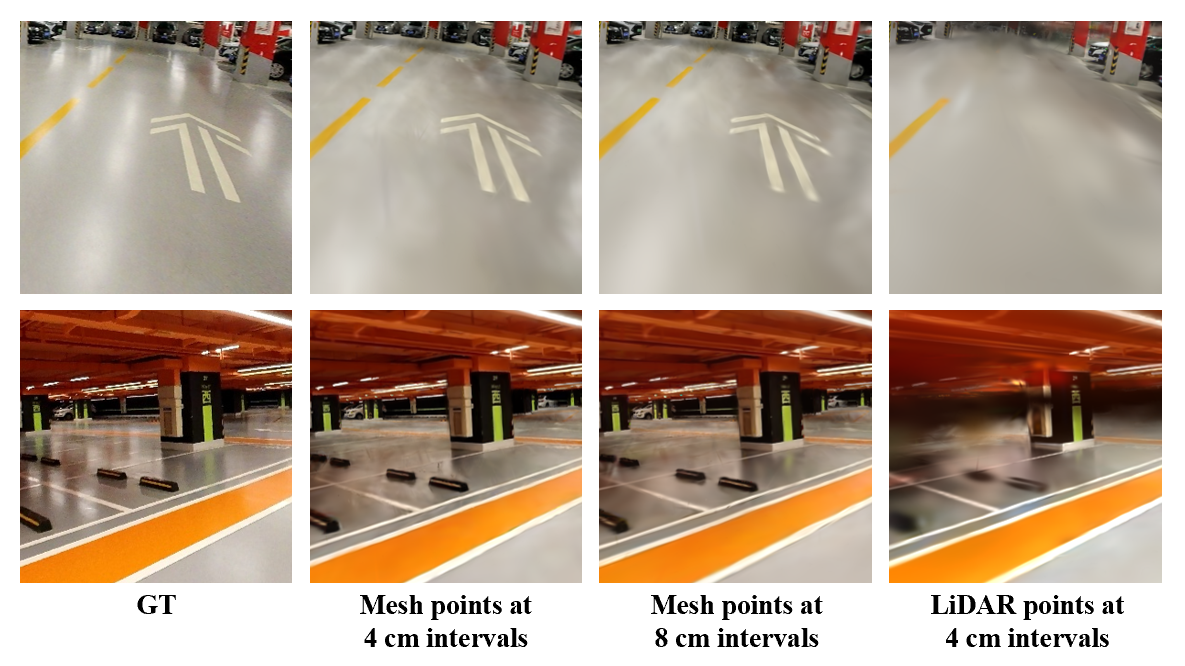}
   \caption{\textcolor{red} {Qualitative ablation study on different input point cloud qualities demonstrates that using point clouds sampled from the mesh significantly enhances rendering quality.
   % compared to LiDAR, and an appropriate density effectively balances processing speed and rendering quality.
   }}
    \label{fig:Ablation_Density}
    \vspace{-2mm}
\end{figure}

\paragraph{Qualitative Comparison Results. }
As shown in Fig.~\ref{fig:comparison}, both 3DGS* and Splactfacto* tend to generate floating Gaussian points around scene surfaces, resulting in blurriness and poor rendering quality. F2-NeRF performs well on the relatively small-scale ScanNet++ dataset but suffers from artifacts in the large-scale KITTI-360 dataset and blurriness in our GarageWorld dataset. Instant-NGP, which utilizes hash encoding, produces notably blurry results in large-scale scenes. In contrast, our method demonstrates the highest rendering quality across various scenarios.
We also compared our method with concurrent works such as OCT-GS* and LoG*. As shown in Fig.~\ref{fig:Concurent work}, our approach achieves the best rendering quality on both our GarageWorld and the UAVD4L datasets (recently proposed and used in LoG*).
% Sam added for READ.
% 我们同样与以Lidar为输入的方法代表做了对比，并在图中展示了qualitative和quantitative的结果对比。对比之下，我们的方法能够恢复更多的高频细节。
% We also conducted a comparison with representative methods that use Lidar as input, presenting both qualitative and quantitative results in the figures. In contrast, our method is able to recover more high-frequency details.
Our web-based results also show comparable visual quality.
\textcolor{red}{Moreover, we evaluate our method against the READ method~\cite{li2023read} which also uses LiDAR as input. Fig.~\ref{fig:READ} presents the qualitative and quantitative results, showing that our method can recover more high-frequency details.}

\paragraph{Rendering Performance of Web-based Renderer. }
Additionally, we evaluate the performance of our web-based renderer on various devices. 
On a high-performance desktop equipped with an i9-10900X CPU and a Samsung SSD T5 Disk, it takes 1.36 seconds to load approximately 2 million 3D Gaussians.
On a MacBook laptop with an Apple M2 CPU and Apple SSD AP0512Z Disk, the load time for the same number of 3D Gaussians is 1.29 seconds. 
Combining these newly loaded 3D Gaussians with an additional 3 million pre-loaded Gaussians, both devices maintain a rendering frame rate of 60 FPS.
\textcolor{red}{
% Because of browser memory limitations, we limit the 3D Gaussian data loaded for rendering to a maximum of 2GB. Additionally, the web renderer relies on CPU-based sorting, which is slower compared to CUDA-based sorting used in the original 3DGS renderer. Therefore, we adopt our designed LOD rendering strategy to trade-off the rendering quality and web performance by controlling the number of loaded 3D Gaussians.
Due to browser memory constraints, we restrict the 3D Gaussian data loaded for rendering to a maximum of \textcolor{orange}{2 GB}. Furthermore, the web renderer uses CPU-based sorting, which is slower than the CUDA-based sorting employed by the original 3DGS renderer. To balance rendering quality and performance, we adopt our LOD rendering strategy that adjusts the number of loaded 3D Gaussians.}

\begin{figure}[t]
    \centering
    \includegraphics[width=\linewidth]{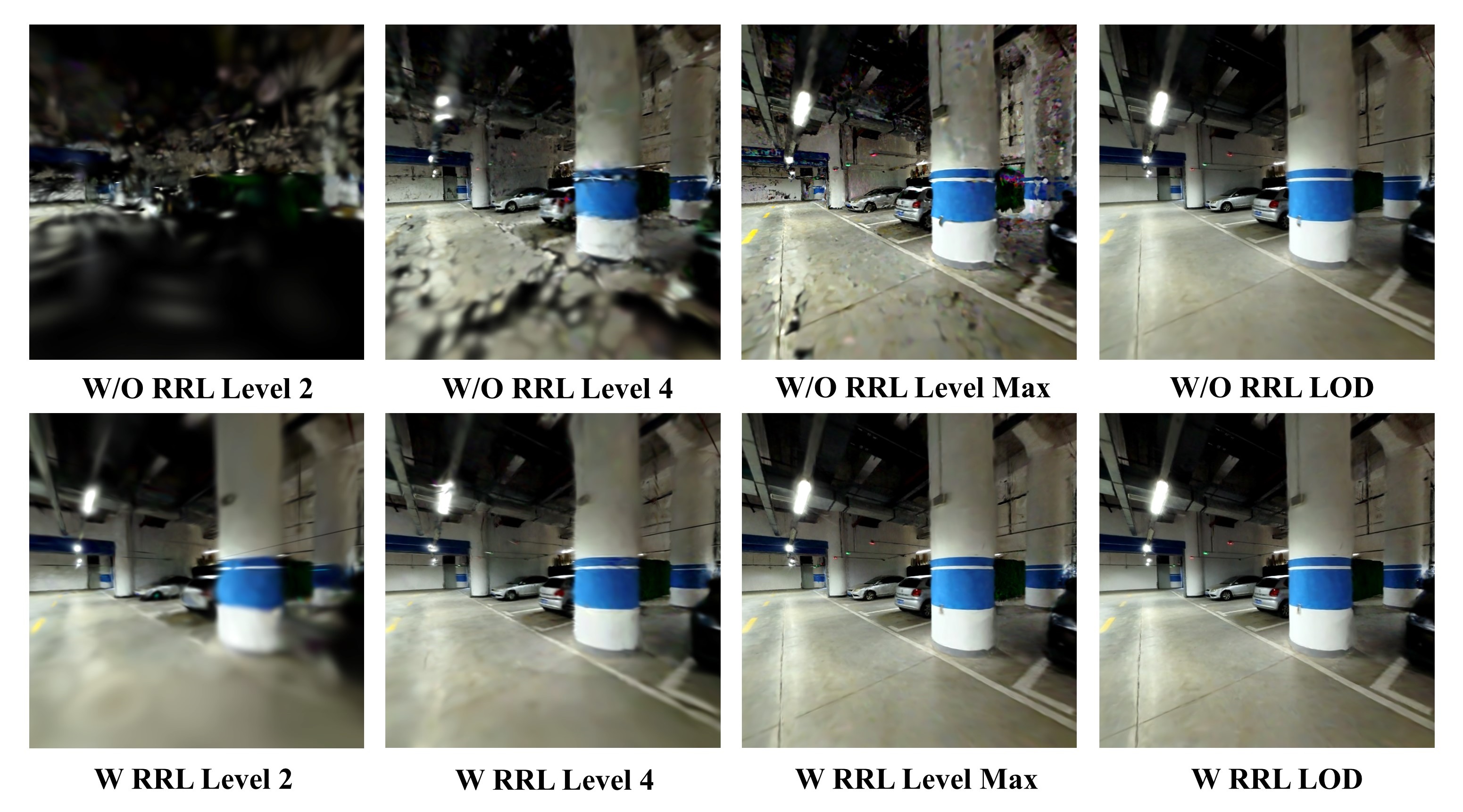}
    \caption{Qualitative ablation study on the random-resolution-level (RRL) training strategy reveals that our RRL approach effectively addresses the overfitting issue at a certain level, influenced by the camera distribution.}
    \label{fig:Ablation_Random_Level}
    \vspace{-2mm}
\end{figure}

\subsection{Ablation Study}
In this section, we perform ablation studies to validate the effectiveness of each component of our approach. 

\begin{table*}[htbp]
\small
\centering
\setlength{\abovecaptionskip}{0pt}
\setlength{\belowcaptionskip}{0pt}
\caption{\textcolor{orange}{Quantitative ablation study on different input point cloud qualities and corresponding training times across two datasets. Point clouds sampled from the mesh at varying intervals and original LiDAR point clouds downsampled to 4 cm intervals. The training times are based on tests performed with an NVIDIA RTX A6000 GPU.}}
\label{tab:ablation}
\setlength{\tabcolsep}{1.6mm}{
    \begin{tabular}{lcccccccccc}
        \toprule
        \multirow{2}{*}{\begin{tabular}[c]{@{}c@{}}Point Cloud\\ Density\end{tabular}} & \multicolumn{5}{c}{Arts Center} & \multicolumn{5}{c}{Shopping Mall 3} \\
        \cline{2-11}
          & PSNR$\uparrow$ & SSIM$\uparrow$ & LPIPS$\downarrow$ & Number of Gaussians & Training Time & PSNR$\uparrow$ & SSIM$\uparrow$ & LPIPS$\downarrow$ & Number of Gaussians & Training Time\\
        \midrule
        2 cm (Mesh)        & \BestCellColor{\textbf{25.87}}  & \BestCellColor{\textbf{0.862}}  & \BestCellColor{\textbf{0.317}}  & 17698464 & 10:55:07 & \BestCellColor{\textbf{25.01}}  & \BestCellColor{\textbf{0.836}}  & \BestCellColor{\textbf{0.320}}  & 36042237 & 31:01:11 \\
        6 cm (Mesh)        & 25.52  & 0.854  & 0.340  & 2300174 & 02:47:02 & 24.73  & 0.827  & 0.340  & 4559012 & 06:35:14 \\
        8 cm (Mesh)        & 25.29  & 0.848  & 0.348  & 1306965 & 02:15:16 & 24.49  & 0.821  & 0.348  & 2566219 & 04:53:08 \\
        10 cm (Mesh)       & 25.08  & 0.843  & 0.356  & 836921  & 02:02:50 & 24.21  & 0.814  & 0.357  & 1627634 & 04:06:29 \\
        4 cm (LiDAR)       & 23.11  & 0.816  & 0.370  & 10024717 & 03:03:03 & 21.52  & 0.764  & 0.399  & 21162613 & 04:39:05 \\

        \midrule
        4 cm (Mesh)        & \SecondBestCellColor{25.70}  & \SecondBestCellColor{0.858}  & \SecondBestCellColor{0.330}  & 4997081 & 02:23:27 & \SecondBestCellColor{24.91}  & \SecondBestCellColor{0.832}  & \SecondBestCellColor{0.331}  & 10015908 & 10:58:39 \\
        \bottomrule
    \end{tabular}}
\vspace{-2mm}
\end{table*}

\paragraph{Depth Regularization.}
Using only photometric loss constraints can easily produce artifacts, such as floaters, in large-scale scene reconstruction. To address this, we introduce an additional depth regularization term to enforce geometric constraints. As shown in Fig.~\ref{fig:Ablation_Depth}, we evaluated the impact of our depth regularization module across various scenes. Our results demonstrate that the \textbf{LOD-LiDAR-GS} method effectively mitigates floaters on the ground when depth regularization is applied.

\paragraph{\textcolor{red}{Input Point Cloud Qualities.}}
\textcolor{orange}{Relying solely on high-density point clouds can lead to computational inefficiencies without significantly enhancing reconstruction quality. To investigate this, we conducted an ablation study assessing the impact of varying point cloud qualities on our reconstruction results. As shown in Fig.~\ref{fig:Ablation_Density} and Tab.~\ref{tab:ablation}, we evaluated the performance of our \textbf{LOD-LiDAR-GS} method using point clouds sampled from the mesh at various intervals, alongside original LiDAR point clouds downsampled to 4 cm intervals. Our findings indicate that while higher densities yield more detailed reconstructions, the benefits plateau beyond a certain threshold. Conversely, lower densities still deliver robust performance with reduced computational costs. To balance training time and reconstruction quality, we selected 4 cm intervals for sampling from the mesh. Notably, using original LiDAR point clouds directly results in poorer outcomes compared to mesh-sampled point clouds, particularly in garage environments where objects are smaller and geometrically simpler. Thus, we opted for point clouds sampled from the mesh at 4 cm intervals, which offer superior reconstruction quality while maintaining efficiency.}

\begin{figure*}[ht]
    \centering
    \setlength{\abovecaptionskip}{0pt}
    \setlength{\belowcaptionskip}{0pt}
    
    \subfigure[Autonomous vehicle parking]{
    \label{fig:autodriving}
    \includegraphics[height=1.28in]{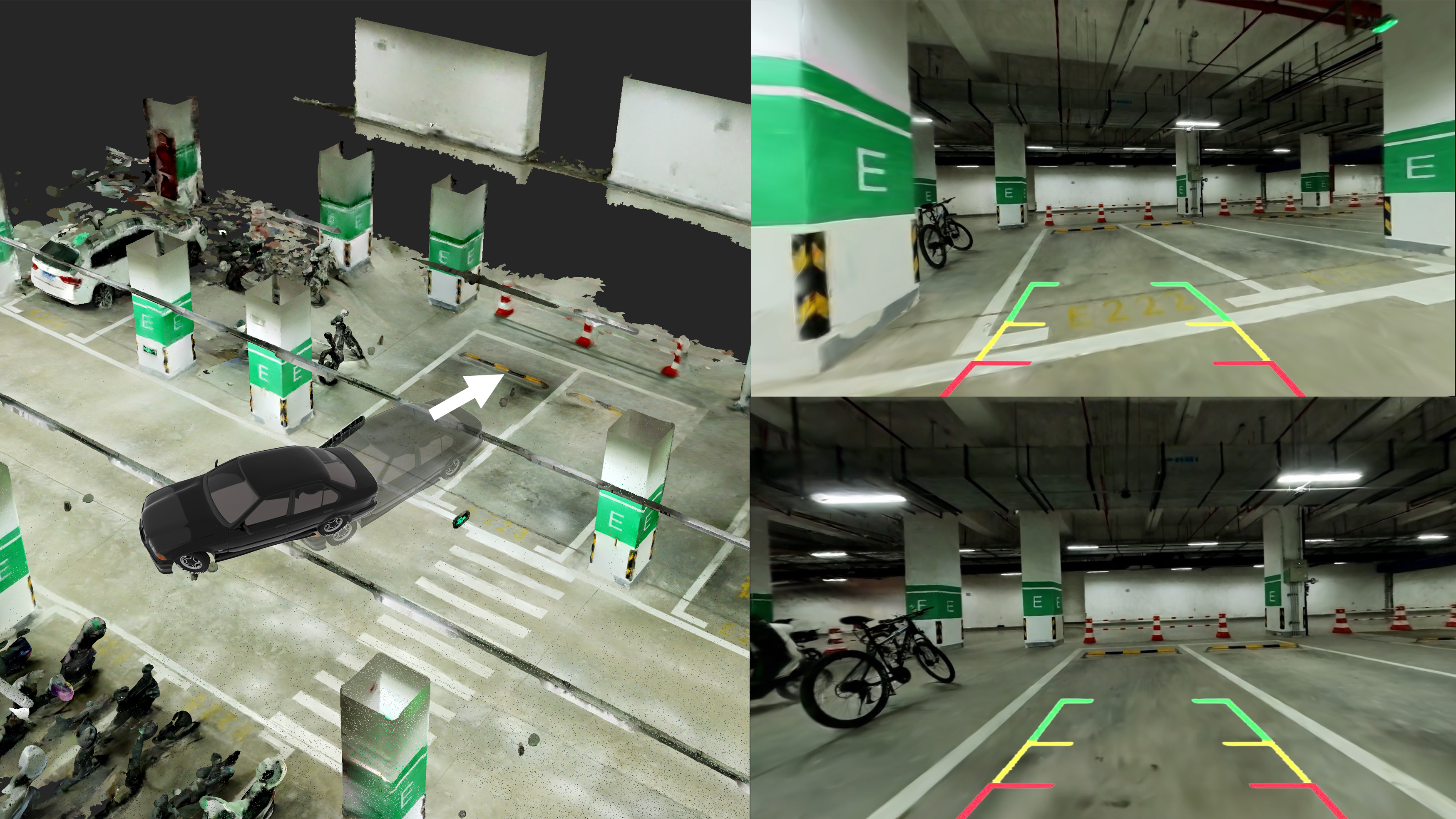}}
    \subfigure[Real-time localization \& navigation]{
    \label{fig:navigation2}
    \includegraphics[height=1.28in]{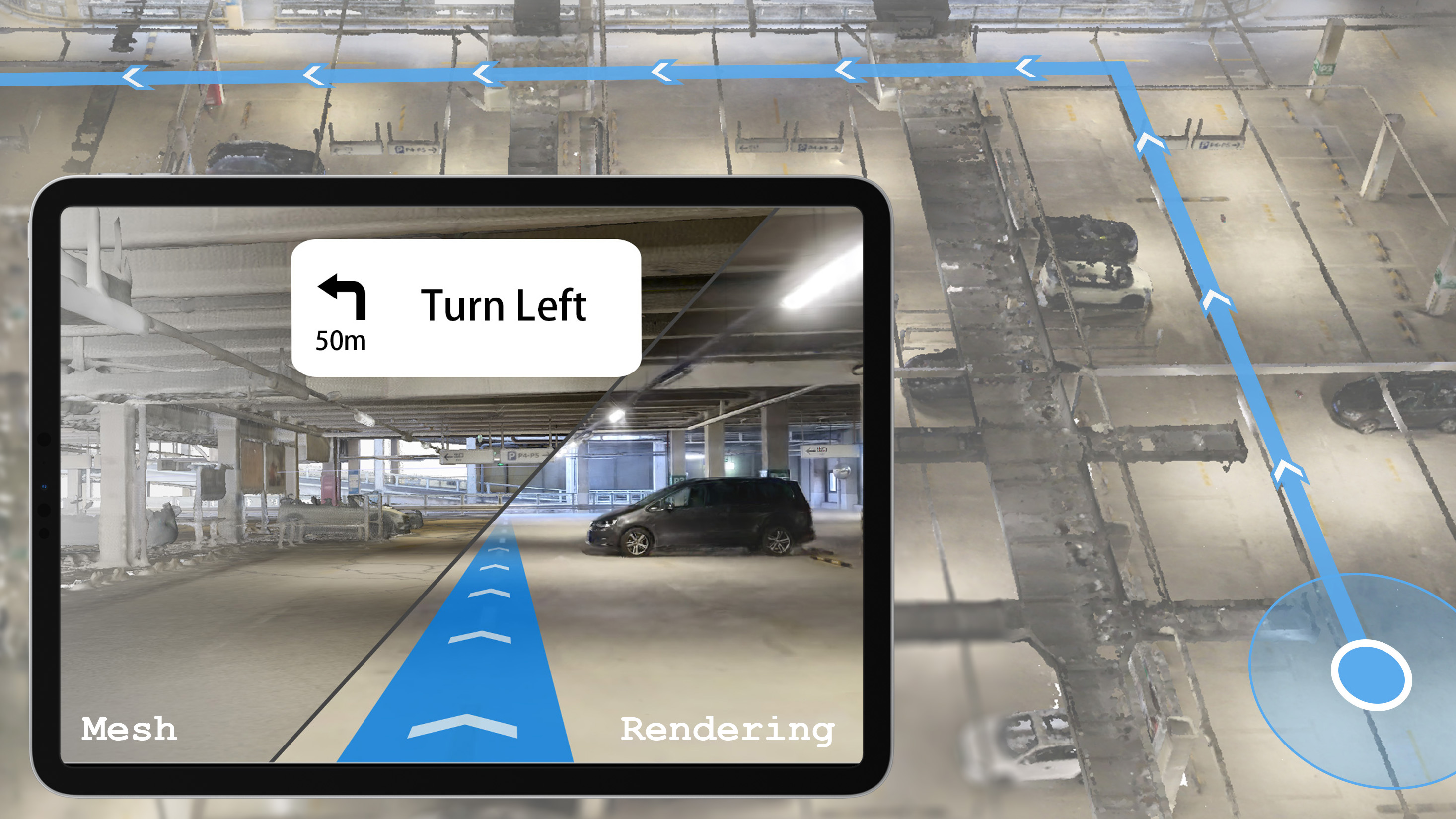}}  
    \subfigure[VFX demonstration]{
    \label{fig:vfx}
    \includegraphics[height=1.28in]{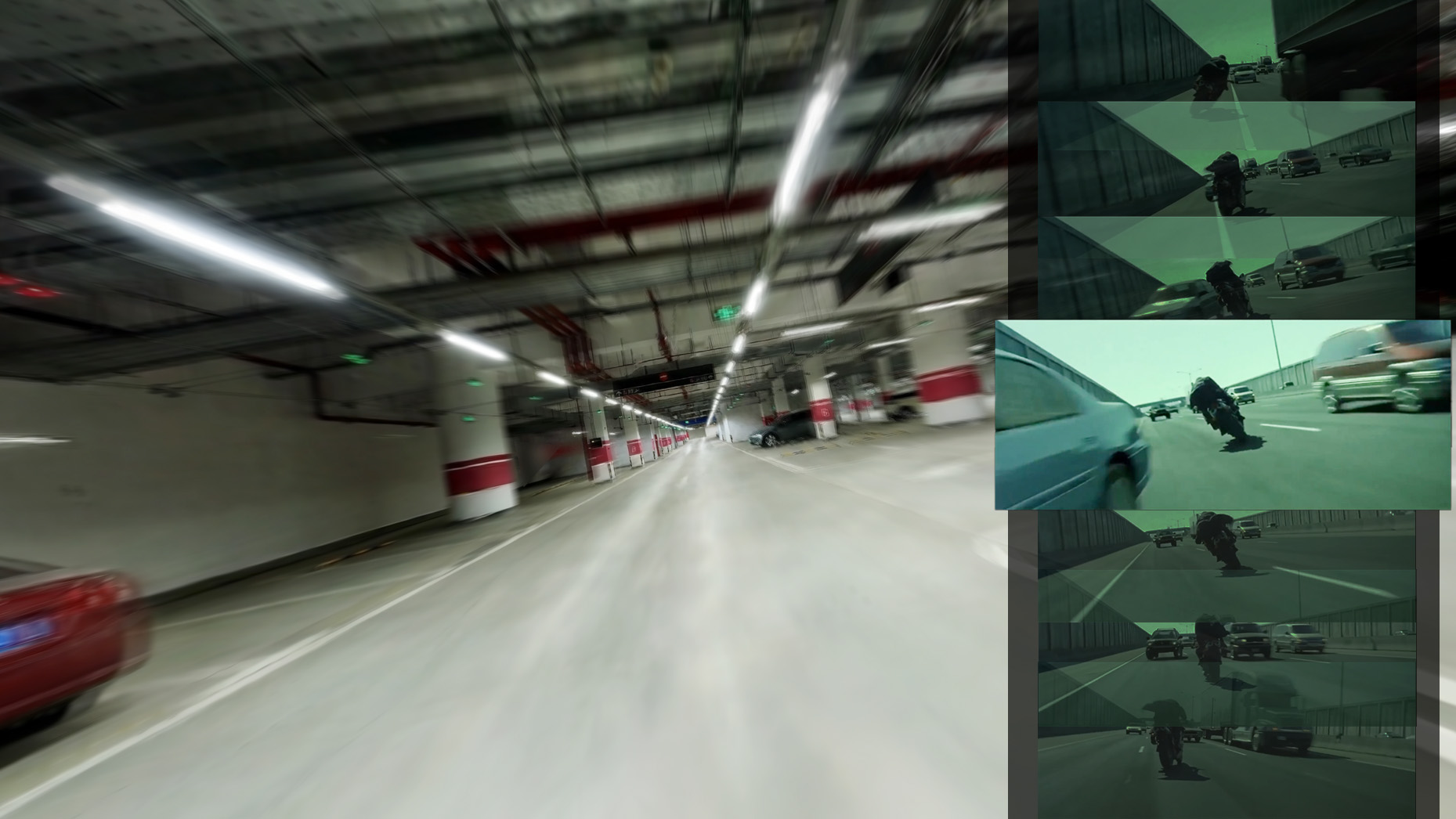}}
    
    \caption{Applications of our system: (a) Autonomous vehicle parking. Our diverse garage scenes facilitate training algorithms for generating parking trajectories under different scenarios;
    % When guiding the vehicle to the parking space, our garage model allows real-time and wide-FOV rendering of the environment, capturing drivable area and obstacles, thereby enhancing safe parking capabilities; 
    (b) Real-time localization \& navigation in challenging garage environments;
    % Our colored 3D model facilitates precise vehicle camera localization and optimal path navigation, particularly in low-light garage conditions, ensuring safe driving. The lightweight web-based rendering ensures deployability in vehicles with limited computing resources
    (c) VFX demonstration. Through an analysis of the animation in the reference video, we extract the poses of several keyframes, which enable our system's renderer to generate corresponding video segments and produce realistic visual effects. 
    % Our 3D garage modeling and rendering also enables motion blur rendering, producing realistic visual effects.
    }
    
    \vspace{-2mm}
    \label{fig:combined}
\end{figure*}

\paragraph{LOD Rendering Strategy.}
Next, we ablate the LOD rendering strategy. Based on our multi-resolution representation, we select Gaussian subsets from various resolution levels according to depth ranges, following the LOD rendering strategy outlined in Equation~\ref{LOD-rendering}. This strategy significantly reduces the complexity of sorting, projection, and accumulation operations involved with 3D Gaussians, thereby accelerating the rendering speed. As shown in Fig.~\ref{fig:Ablation_FPS}, our real-time viewer, integrated with the LOD rendering technique, is about 3-4 times faster than 3DGS's SIBR viewer with similar rendering quality.

\paragraph{Random-Resolution-Level (RRL) Training Strategy.}
% For our large-scale garage scenes, we observed that the distribution of input cameras significantly affects the multi-resolution training strategy for \textbf{LOD-LiDAR-GS}. 
During training, directly applying the LOD rendering strategy to render images and optimize the loss can lead to overfitting at certain resolution levels in specific scene regions. We propose a random-resolution-level (RRL) training strategy to address this issue. As shown in Fig.~\ref{fig:Ablation_Random_Level}, our RRL training strategy mitigates the overfitting problem and remains robust to variations in the distribution of input cameras.

\section{GarageWorld}

We introduce the first large-scale garage dataset, \textbf{GarageWorld}, captured with our Polar device. This dataset comprises eight large garages, featuring six underground garages, one outdoor parking lot, and one multi-floor indoor garage. 
Fig.~\ref{fig:dataset_gallery} visualizes the 3D models and rendered images of a subsection of these garages using our method. Our rendering pipeline and the GarageWorld dataset are designed to support a wide range of applications, which are further detailed in the following sections.

\paragraph{Data Generation and Testbed for Autonomous Driving.} 
Our developed 3D reconstruction of large-scale garage scenes with enhanced real-time rendering can support various autonomous driving algorithms. 
% We have developed a 3D reconstruction of large-scale garage scenes with enhanced real-time rendering to support various autonomous driving algorithms.
The GarageWorld dataset provides diverse driving scenarios, including tight parking spaces and complex geometric layouts, which are instrumental for training algorithms in autonomous parking and navigation path planning. 
Additionally, our LetsGo pipeline further aids vehicle parking by supplementing sensor data, particularly in low-light garage conditions. Our LiDAR-assisted Gaussian primitives allow rigorous testing in a virtual setting, prompting safe and efficient parking solutions. Fig.~\ref{fig:autodriving} illustrates how our approach generates rear-view images, allowing vehicles to accurately estimate distances and navigate tight spaces, thereby avoiding collisions and accelerating the development and deployment of safe autonomous driving technologies.

\paragraph{Real-time Localization and Navigation.}
\textcolor{red}{Our LiDAR-assisted garage modeling and rendering provide accurate 3D references, significantly enhancing autonomous vehicle localization and navigation, particularly in challenging indoor or underground garages where traditional methods struggle. 
% By integrating our Polar scanner with LiDAR-assisted Gaussian primitives, we improve feature recognition and matching, enhancing localization and navigation reliability. 
}
Fig.~\ref{fig:navigation2} presents an example of real-time localization and navigation. 
Additionally, our web-based rendering engine represents expansive 3D garage maps and navigation trajectories on lightweight devices with limited computational capacity, delivering smooth real-time interactions and superior rendering performance.

\paragraph{VFX Production.} Our reconstructed garage datasets and real-time rendering capabilities significantly enhance VFX production by providing a robust foundation for creating realistic backgrounds and seamlessly integrating CGI elements into live-action footage. \textcolor{red}{Inspired by the highway chase in "The Matrix Reloaded", our LetsGo pipeline excels in rendering complex garage scenes, offering precise control over elements like exposure time and motion blur.}
We created a VFX video production to showcase our technology using our reconstructed garage models (Fig.~\ref{fig:vfx}) in the supplementary video. We downloaded a video depicting highway chase scenes with dynamic viewing angles of a rapidly moving motorcycle. By extracting and aligning camera trajectories with our models and synthesizing motion blur, we created immersive visual effects that convincingly depict a chase within a garage environment. Our approach supports the exploration of diverse camera angles and compositions in real-time while ensuring visual consistency between real and virtual components, significantly reducing the need for reshoots and streamlining the production process.

\section{Limitations and Discussions}

% Given that the large texture-less regions and repetitive patterns in garage scenes hinder the 3D reconstruction of expansive garages from images by SfM and MVS algorithms,
% % Consider the 3D point clouds of large-scale garage data are hard to reconstruct from images by SfM algorithms, due to the large texture-less regions and repetitive patterns. 
% this paper designs a Polar device using which the point cloud data of the large-scale garages can be easily captured and calibrated, and demonstrated that the collected LiDAR data successfully assist a suit of 3D Gaussian splatting algorithms, enabling high-quality rendering. 
% % By leveraging the advantages of both mesh-based and state-of-the-art Gaussian-based representations, we introduce the LetsGo method for large-scale garage scenes of various types, enabling high-quality 3D modeling and rendering.
% Moreover, our lightweight renderer, combined with LOD techniques, facilitates real-time rendering of expansive garage environments on lightweight platforms. 
Although our rendering results exhibit realism and we showcase various applications, our workflow still possesses certain limitations. Here we present a detailed analysis and explore further potential applications.

Firstly, our method focuses on large-scale garage scenes and relies on our lightweight 3D scanner to provide high-quality LiDAR point clouds, color images, and corresponding camera information. Although we have also validated the effectiveness of our method on other open-source datasets, it is important to explore the use of even lighter devices for scanning and rendering large-scale scenes, such as smartphones equipped with depth sensors. Additionally, we have currently collected data from eight large-scale garages, but it is necessary to gather more garage data to contribute to the community and facilitate further research on garage modeling and rendering.

As a method based on image rendering, our approach achieves highly realistic rendering effects, almost indistinguishable from real scenes. However, the existing pipeline does not support modifying lighting conditions. This requires us to carefully design the shooting process according to the lighting conditions, limiting the applicability of our method in various scenarios. 
% Recently, Gaussian-based work SuGaR~\cite{guedon2023sugar} has introduced editing operations such as adjusting the size, position, and orientation of objects, but it still does not support modifying lighting information. 
Enabling rendering and editing of large-scale scenes under different lighting conditions is a meaningful direction that deserves further investigation. Our open-source dataset provides a foundation for the community to conduct research in these directions, allowing for advancements in this field.
\textcolor{red}{Also there are switching artifacts primarily caused by loading different Gaussian levels. These artifacts can be mitigated by interpolating between adjacent Gaussian levels during switching.}
\textcolor{red}{\cite{hierarchicalgaussians24} successfully achieves smooth level transitions using a linear interpolation scheme. We will explore this approach and decrease the switching artifacts in future work.}

% CityDreamer, ReconFusion from sparse views
% We have demonstrated the applications of our method in automatic driving data generation, localization and navigation on lightweight devices, and the production of visual effects in films. These applications illustrate the excellent performance of our method in rendering quality. 
Furthermore, there are additional directions and applications worth exploring. Inspired by SMERF~\cite{duckworth2024smerf}, one potential research direction is to investigate streaming transmission methods for Gaussian kernels, enabling the distribution and on-the-fly rendering of large-scale scene data. Additionally, there are city generation methods such as InfiniCity~\cite{lin2023infinicity} and CityDreamer~\cite{xie2024citydreamer}, which leverage Generative Adversarial Networks (GANs)~\cite{goodfellow2014generative} to achieve rapid modeling of large-scale scenes. We intend to study large-scale scene generation based on 3D Gaussian representations. This fully explicit representation can be easily integrated into existing computer graphics workflows and achieve superior rendering effects.

% \begin{figure*}[t]
% \begin{figure*}
%     \centering
% %     \setlength{\abovecaptionskip}{0pt}
% % \setlength{\belowcaptionskip}{0pt}
%     % \includegraphics[width=0.97\linewidth]{figures/gallery_2.jpg}
%     \includegraphics[width=0.91\linewidth]{figures/gallery_2.jpg}
%     \caption{A gallery showcasing the rendering results of our pipeline across various scenes in GarageWorld. Our LiDAR-assisted Gaussian primitives enable photorealistic rendering of expansive garages. The enlarged images illustrate the fidelity of our high-resolution point clouds and captured imagery. }
%     \label{fig:gallery}
% \end{figure*}

\section{Conclusion} 
For many of us, our daily lives begin with a safe departure from a garage and end with a safe arrival. The garage serves as the origin of our journey to innovation. 
This paper has contributed a handheld Polar device for data collection, a GarageWorld dataset, LiDAR-assisted Gaussian splatting for scene representation, and an LOD-based rendering technique that allows web-based rendering on consumer-level devices.  
Benefiting from these innovations, we successfully reconstruct various garages with diverse and challenging environments, allowing real-time rendering from any viewpoint on lightweight devices. 
Experimental results on the collected and two public datasets have demonstrated the effectiveness of our approach. 
Our GarageWorld, along with the reconstructed 3D model and real-time rendering, enables a set of applications, including training data generation and testbed for autonomous driving algorithms, real-time assistance for autonomous vehicle localization, navigation, and parking, as well as VFX production. 
Our current contributions mainly focus on the perception of the world, and enables downstream recognition tasks.
In the future, we will also explore garage generation, continuing to push the boundary of garage modeling and accomplishing a closure from perception, recognition, and generation.

\begin{acks}
	The authors would like to thank Shanghai Municipal Big Data Center for coordinating the collection of garage data.
	This work was supported by National Key R$\&$D Program of China (2022YFF0902301), NSFC programs (61976138, 61977047), STCSM (2015F0203-000-06), and SHMEC (2019-01-07-00-01-E00003). We also acknowledge support from Shanghai Frontiers Science Center of Human-centered Artificial Intelligence (ShangHAI) and MoE Key Lab of Intelligent Perception and Human-Machine Collaboration (ShanghaiTech University).	
\end{acks}

% Bibliography
\bibliographystyle{ACM-Reference-Format}
\bibliography{references.bib}


\begin{thebibliography}{105}

%%% the Web address.

\ifx \showCODEN    \undefined \def \showCODEN     #1{\unskip}     \fi
\ifx \showDOI      \undefined \def \showDOI       #1{#1}\fi
\ifx \showISBNx    \undefined \def \showISBNx     #1{\unskip}     \fi
\ifx \showISBNxiii \undefined \def \showISBNxiii  #1{\unskip}     \fi
\ifx \showISSN     \undefined \def \showISSN      #1{\unskip}     \fi
\ifx \showLCCN     \undefined \def \showLCCN      #1{\unskip}     \fi
\ifx \shownote     \undefined \def \shownote      #1{#1}          \fi
\ifx \showarticletitle \undefined \def \showarticletitle #1{#1}   \fi
\ifx \showURL      \undefined \def \showURL       {\relax}        \fi
% invisible to TeX
\providecommand\bibfield[2]{#2}
\providecommand\bibinfo[2]{#2}
\providecommand\natexlab[1]{#1}
\providecommand\showeprint[2][]{arXiv:#2}

\bibitem[Agarwal et~al\mbox{.}(2011)]%
        {agarwal2011building}
\bibfield{author}{\bibinfo{person}{Sameer Agarwal}, \bibinfo{person}{Yasutaka
  Furukawa}, \bibinfo{person}{Noah Snavely}, \bibinfo{person}{Ian Simon},
  \bibinfo{person}{Brian Curless}, \bibinfo{person}{Steven~M Seitz}, {and}
  \bibinfo{person}{Richard Szeliski}.} \bibinfo{year}{2011}\natexlab{}.
\newblock \showarticletitle{Building Rome in a day}.
\newblock \bibinfo{journal}{\emph{Commun. ACM}} \bibinfo{volume}{54},
  \bibinfo{number}{10} (\bibinfo{year}{2011}), \bibinfo{pages}{105--112}.
\newblock


\bibitem[Aliev et~al\mbox{.}(2020)]%
        {aliev2020neural}
\bibfield{author}{\bibinfo{person}{Kara-Ali Aliev}, \bibinfo{person}{Artem
  Sevastopolsky}, \bibinfo{person}{Maria Kolos}, \bibinfo{person}{Dmitry
  Ulyanov}, {and} \bibinfo{person}{Victor Lempitsky}.}
  \bibinfo{year}{2020}\natexlab{}.
\newblock \showarticletitle{Neural point-based graphics}. In
  \bibinfo{booktitle}{\emph{Computer Vision--ECCV 2020: 16th European
  Conference, Glasgow, UK, August 23--28, 2020, Proceedings, Part XXII 16}}.
  Springer, \bibinfo{pages}{696--712}.
\newblock


\bibitem[Barron et~al\mbox{.}(2021)]%
        {barron2021mip}
\bibfield{author}{\bibinfo{person}{Jonathan~T Barron}, \bibinfo{person}{Ben
  Mildenhall}, \bibinfo{person}{Matthew Tancik}, \bibinfo{person}{Peter
  Hedman}, \bibinfo{person}{Ricardo Martin-Brualla}, {and}
  \bibinfo{person}{Pratul~P Srinivasan}.} \bibinfo{year}{2021}\natexlab{}.
\newblock \showarticletitle{Mip-nerf: A multiscale representation for
  anti-aliasing neural radiance fields}. In
  \bibinfo{booktitle}{\emph{Proceedings of the IEEE/CVF International
  Conference on Computer Vision}}. \bibinfo{pages}{5855--5864}.
\newblock


\bibitem[Barron et~al\mbox{.}(2022)]%
        {barron2022mip}
\bibfield{author}{\bibinfo{person}{Jonathan~T Barron}, \bibinfo{person}{Ben
  Mildenhall}, \bibinfo{person}{Dor Verbin}, \bibinfo{person}{Pratul~P
  Srinivasan}, {and} \bibinfo{person}{Peter Hedman}.}
  \bibinfo{year}{2022}\natexlab{}.
\newblock \showarticletitle{Mip-nerf 360: Unbounded anti-aliased neural
  radiance fields}. In \bibinfo{booktitle}{\emph{Proceedings of the IEEE/CVF
  Conference on Computer Vision and Pattern Recognition}}.
  \bibinfo{pages}{5470--5479}.
\newblock


\bibitem[Bavle et~al\mbox{.}(2023)]%
        {bavle2023slam}
\bibfield{author}{\bibinfo{person}{Hriday Bavle}, \bibinfo{person}{Jose~Luis
  Sanchez-Lopez}, \bibinfo{person}{Claudio Cimarelli}, \bibinfo{person}{Ali
  Tourani}, {and} \bibinfo{person}{Holger Voos}.}
  \bibinfo{year}{2023}\natexlab{}.
\newblock \showarticletitle{From SLAM to situational awareness: Challenges and
  survey}.
\newblock \bibinfo{journal}{\emph{Sensors}} \bibinfo{volume}{23},
  \bibinfo{number}{10} (\bibinfo{year}{2023}), \bibinfo{pages}{4849}.
\newblock


\bibitem[Bujanca et~al\mbox{.}(2021)]%
        {bujanca2021robust}
\bibfield{author}{\bibinfo{person}{Mihai Bujanca}, \bibinfo{person}{Xuesong
  Shi}, \bibinfo{person}{Matthew Spear}, \bibinfo{person}{Pengpeng Zhao},
  \bibinfo{person}{Barry Lennox}, {and} \bibinfo{person}{Mikel Luj{\'a}n}.}
  \bibinfo{year}{2021}\natexlab{}.
\newblock \showarticletitle{Robust SLAM systems: Are we there yet?}. In
  \bibinfo{booktitle}{\emph{2021 IEEE/RSJ International Conference on
  Intelligent Robots and Systems (IROS)}}. IEEE, \bibinfo{pages}{5320--5327}.
\newblock


\bibitem[Ceriani et~al\mbox{.}(2015)]%
        {ceriani2015pose}
\bibfield{author}{\bibinfo{person}{Simone Ceriani}, \bibinfo{person}{Carlos
  S{\'a}nchez}, \bibinfo{person}{Pierluigi Taddei}, \bibinfo{person}{Erik
  Wolfart}, {and} \bibinfo{person}{V{\'\i}tor Sequeira}.}
  \bibinfo{year}{2015}\natexlab{}.
\newblock \showarticletitle{Pose interpolation SLAM for large maps using moving
  3d sensors}. In \bibinfo{booktitle}{\emph{2015 IEEE/RSJ international
  conference on intelligent robots and systems (IROS)}}. IEEE,
  \bibinfo{pages}{750--757}.
\newblock


\bibitem[Charatan et~al\mbox{.}(2024)]%
        {charatan2024pixelsplat}
\bibfield{author}{\bibinfo{person}{David Charatan},
  \bibinfo{person}{Sizhe~Lester Li}, \bibinfo{person}{Andrea Tagliasacchi},
  {and} \bibinfo{person}{Vincent Sitzmann}.} \bibinfo{year}{2024}\natexlab{}.
\newblock \showarticletitle{PixelSplat: 3d Gaussian splats from image pairs for
  scalable generalizable 3d reconstruction}. In
  \bibinfo{booktitle}{\emph{Proceedings of the IEEE/CVF Conference on Computer
  Vision and Pattern Recognition}}. \bibinfo{pages}{19457--19467}.
\newblock


\bibitem[Chen et~al\mbox{.}(2022)]%
        {Chen2022ECCV}
\bibfield{author}{\bibinfo{person}{Anpei Chen}, \bibinfo{person}{Zexiang Xu},
  \bibinfo{person}{Andreas Geiger}, \bibinfo{person}{Jingyi Yu}, {and}
  \bibinfo{person}{Hao Su}.} \bibinfo{year}{2022}\natexlab{}.
\newblock \showarticletitle{TensoRF: Tensorial Radiance Fields}. In
  \bibinfo{booktitle}{\emph{European Conference on Computer Vision (ECCV)}}.
\newblock


\bibitem[Chen et~al\mbox{.}(2021)]%
        {chen2021mvsnerf}
\bibfield{author}{\bibinfo{person}{Anpei Chen}, \bibinfo{person}{Zexiang Xu},
  \bibinfo{person}{Fuqiang Zhao}, \bibinfo{person}{Xiaoshuai Zhang},
  \bibinfo{person}{Fanbo Xiang}, \bibinfo{person}{Jingyi Yu}, {and}
  \bibinfo{person}{Hao Su}.} \bibinfo{year}{2021}\natexlab{}.
\newblock \showarticletitle{Mvsnerf: Fast generalizable radiance field
  reconstruction from multi-view stereo}. In
  \bibinfo{booktitle}{\emph{Proceedings of the IEEE/CVF International
  Conference on Computer Vision}}. \bibinfo{pages}{14124--14133}.
\newblock


\bibitem[Cheng et~al\mbox{.}(2024)]%
        {cheng2024gaussianpro}
\bibfield{author}{\bibinfo{person}{Kai Cheng}, \bibinfo{person}{Xiaoxiao Long},
  \bibinfo{person}{Kaizhi Yang}, \bibinfo{person}{Yao Yao},
  \bibinfo{person}{Wei Yin}, \bibinfo{person}{Yuexin Ma},
  \bibinfo{person}{Wenping Wang}, {and} \bibinfo{person}{Xuejin Chen}.}
  \bibinfo{year}{2024}\natexlab{}.
\newblock \showarticletitle{Gaussianpro: 3d Gaussian Splatting with progressive
  propagation}. In \bibinfo{booktitle}{\emph{Forty-first International
  Conference on Machine Learning}}.
\newblock


\bibitem[Cignoni et~al\mbox{.}(2004)]%
        {10.1145/1015706.1015802}
\bibfield{author}{\bibinfo{person}{Paolo Cignoni}, \bibinfo{person}{Fabio
  Ganovelli}, \bibinfo{person}{Enrico Gobbetti}, \bibinfo{person}{Fabio
  Marton}, \bibinfo{person}{Federico Ponchio}, {and} \bibinfo{person}{Roberto
  Scopigno}.} \bibinfo{year}{2004}\natexlab{}.
\newblock \showarticletitle{Adaptive tetrapuzzles: efficient out-of-core
  construction and visualization of gigantic multiresolution polygonal models}.
\newblock \bibinfo{journal}{\emph{ACM Trans. Graph.}} \bibinfo{volume}{23},
  \bibinfo{number}{3} (\bibinfo{date}{aug} \bibinfo{year}{2004}),
  \bibinfo{pages}{796–803}.
\newblock
\showISSN{0730-0301}
\urldef\tempurl%
\url{https://doi.org/10.1145/1015706.1015802}
\showDOI{\tempurl}


\bibitem[Crandall et~al\mbox{.}(2011)]%
        {crandall2011discrete}
\bibfield{author}{\bibinfo{person}{David Crandall}, \bibinfo{person}{Andrew
  Owens}, \bibinfo{person}{Noah Snavely}, {and} \bibinfo{person}{Dan
  Huttenlocher}.} \bibinfo{year}{2011}\natexlab{}.
\newblock \showarticletitle{Discrete-continuous optimization for large-scale
  structure from motion}. In \bibinfo{booktitle}{\emph{CVPR 2011}}. IEEE,
  \bibinfo{pages}{3001--3008}.
\newblock


\bibitem[Dai et~al\mbox{.}(2017)]%
        {dai2017scannet}
\bibfield{author}{\bibinfo{person}{Angela Dai}, \bibinfo{person}{Angel~X
  Chang}, \bibinfo{person}{Manolis Savva}, \bibinfo{person}{Maciej Halber},
  \bibinfo{person}{Thomas Funkhouser}, {and} \bibinfo{person}{Matthias
  Nie{\ss}ner}.} \bibinfo{year}{2017}\natexlab{}.
\newblock \showarticletitle{Scannet: Richly-annotated 3d reconstructions of
  indoor scenes}. In \bibinfo{booktitle}{\emph{Proceedings of the IEEE
  conference on computer vision and pattern recognition}}.
  \bibinfo{pages}{5828--5839}.
\newblock


\bibitem[Dai et~al\mbox{.}(2024)]%
        {Dai2024GaussianSurfels}
\bibfield{author}{\bibinfo{person}{Pinxuan Dai}, \bibinfo{person}{Jiamin Xu},
  \bibinfo{person}{Wenxiang Xie}, \bibinfo{person}{Xinguo Liu},
  \bibinfo{person}{Huamin Wang}, {and} \bibinfo{person}{Weiwei Xu}.}
  \bibinfo{year}{2024}\natexlab{}.
\newblock \showarticletitle{High-quality Surface Reconstruction using Gaussian
  Surfels}. In \bibinfo{booktitle}{\emph{SIGGRAPH 2024 Conference Papers}}.
  \bibinfo{publisher}{Association for Computing Machinery}.
\newblock
\urldef\tempurl%
\url{https://doi.org/10.1145/3641519.3657441}
\showDOI{\tempurl}


\bibitem[Dai et~al\mbox{.}(2019)]%
        {dai2019mvs2}
\bibfield{author}{\bibinfo{person}{Yuchao Dai}, \bibinfo{person}{Zhidong Zhu},
  \bibinfo{person}{Zhibo Rao}, {and} \bibinfo{person}{Bo Li}.}
  \bibinfo{year}{2019}\natexlab{}.
\newblock \showarticletitle{Mvs2: Deep unsupervised multi-view stereo with
  multi-view symmetry}. In \bibinfo{booktitle}{\emph{2019 International
  Conference on 3D Vision (3DV)}}. Ieee, \bibinfo{pages}{1--8}.
\newblock


\bibitem[Deng et~al\mbox{.}(2022)]%
        {kangle2021dsnerf}
\bibfield{author}{\bibinfo{person}{Kangle Deng}, \bibinfo{person}{Andrew Liu},
  \bibinfo{person}{Jun-Yan Zhu}, {and} \bibinfo{person}{Deva Ramanan}.}
  \bibinfo{year}{2022}\natexlab{}.
\newblock \showarticletitle{Depth-supervised {NeRF}: Fewer Views and Faster
  Training for Free}. In \bibinfo{booktitle}{\emph{Proceedings of the IEEE/CVF
  Conference on Computer Vision and Pattern Recognition (CVPR)}}.
\newblock


\bibitem[Duckworth et~al\mbox{.}(2024)]%
        {duckworth2024smerf}
\bibfield{author}{\bibinfo{person}{Daniel Duckworth}, \bibinfo{person}{Peter
  Hedman}, \bibinfo{person}{Christian Reiser}, \bibinfo{person}{Peter Zhizhin},
  \bibinfo{person}{Jean-Fran{\c{c}}ois Thibert}, \bibinfo{person}{Mario
  Lu{\v{c}}i{\'c}}, \bibinfo{person}{Richard Szeliski}, {and}
  \bibinfo{person}{Jonathan~T Barron}.} \bibinfo{year}{2024}\natexlab{}.
\newblock \showarticletitle{SMERF: Streamable memory efficient radiance fields
  for real-time large-scene exploration}.
\newblock \bibinfo{journal}{\emph{ACM Transactions on Graphics (TOG)}}
  \bibinfo{volume}{43}, \bibinfo{number}{4} (\bibinfo{year}{2024}),
  \bibinfo{pages}{1--13}.
\newblock


\bibitem[Face(2024)]%
        {huggingface_gsplat}
\bibfield{author}{\bibinfo{person}{Hugging Face}.}
  \bibinfo{year}{2024}\natexlab{}.
\newblock \bibinfo{title}{gsplat.js: JavaScript Gaussian Splatting library}.
\newblock
  \bibinfo{howpublished}{\url{https://github.com/huggingface/gsplat.js}}.
\newblock
\newblock
\shownote{Accessed: 2024-05-15}.


\bibitem[Frahm et~al\mbox{.}(2010)]%
        {frahm2010building}
\bibfield{author}{\bibinfo{person}{Jan-Michael Frahm}, \bibinfo{person}{Pierre
  Fite-Georgel}, \bibinfo{person}{David Gallup}, \bibinfo{person}{Tim Johnson},
  \bibinfo{person}{Rahul Raguram}, \bibinfo{person}{Changchang Wu},
  \bibinfo{person}{Yi-Hung Jen}, \bibinfo{person}{Enrique Dunn},
  \bibinfo{person}{Brian Clipp}, \bibinfo{person}{Svetlana Lazebnik},
  {et~al\mbox{.}}} \bibinfo{year}{2010}\natexlab{}.
\newblock \showarticletitle{Building Rome on a cloudless day}. In
  \bibinfo{booktitle}{\emph{Computer Vision--ECCV 2010: 11th European
  Conference on Computer Vision, Heraklion, Crete, Greece, September 5-11,
  2010, Proceedings, Part IV 11}}. Springer, \bibinfo{pages}{368--381}.
\newblock


\bibitem[Franke et~al\mbox{.}(2024)]%
        {franke2024trips}
\bibfield{author}{\bibinfo{person}{Linus Franke}, \bibinfo{person}{Darius
  R{\"u}ckert}, \bibinfo{person}{Laura Fink}, {and} \bibinfo{person}{Marc
  Stamminger}.} \bibinfo{year}{2024}\natexlab{}.
\newblock \showarticletitle{TRIPS: Trilinear Point Splatting for Real-Time
  Radiance Field Rendering}. In \bibinfo{booktitle}{\emph{Computer Graphics
  Forum}}. Wiley Online Library, \bibinfo{pages}{e15012}.
\newblock


\bibitem[Fridovich-Keil et~al\mbox{.}(2022)]%
        {fridovich2022plenoxels}
\bibfield{author}{\bibinfo{person}{Sara Fridovich-Keil}, \bibinfo{person}{Alex
  Yu}, \bibinfo{person}{Matthew Tancik}, \bibinfo{person}{Qinhong Chen},
  \bibinfo{person}{Benjamin Recht}, {and} \bibinfo{person}{Angjoo Kanazawa}.}
  \bibinfo{year}{2022}\natexlab{}.
\newblock \showarticletitle{Plenoxels: Radiance fields without neural
  networks}. In \bibinfo{booktitle}{\emph{Proceedings of the IEEE/CVF
  Conference on Computer Vision and Pattern Recognition}}.
  \bibinfo{pages}{5501--5510}.
\newblock


\bibitem[Furgale et~al\mbox{.}(2012)]%
        {furgale2012continuous}
\bibfield{author}{\bibinfo{person}{Paul Furgale}, \bibinfo{person}{Timothy~D
  Barfoot}, {and} \bibinfo{person}{Gabe Sibley}.}
  \bibinfo{year}{2012}\natexlab{}.
\newblock \showarticletitle{Continuous-time batch estimation using temporal
  basis functions}. In \bibinfo{booktitle}{\emph{2012 IEEE International
  Conference on Robotics and Automation}}. IEEE, \bibinfo{pages}{2088--2095}.
\newblock


\bibitem[Furgale et~al\mbox{.}(2013)]%
        {furgale2013unified}
\bibfield{author}{\bibinfo{person}{Paul Furgale}, \bibinfo{person}{Joern
  Rehder}, {and} \bibinfo{person}{Roland Siegwart}.}
  \bibinfo{year}{2013}\natexlab{}.
\newblock \showarticletitle{Unified temporal and spatial calibration for
  multi-sensor systems}. In \bibinfo{booktitle}{\emph{2013 IEEE/RSJ
  International Conference on Intelligent Robots and Systems}}. IEEE,
  \bibinfo{pages}{1280--1286}.
\newblock


\bibitem[Furukawa et~al\mbox{.}(2010)]%
        {5539802}
\bibfield{author}{\bibinfo{person}{Yasutaka Furukawa}, \bibinfo{person}{Brian
  Curless}, \bibinfo{person}{Steven~M. Seitz}, {and} \bibinfo{person}{Richard
  Szeliski}.} \bibinfo{year}{2010}\natexlab{}.
\newblock \showarticletitle{Towards Internet-scale multi-view stereo}. In
  \bibinfo{booktitle}{\emph{2010 IEEE Computer Society Conference on Computer
  Vision and Pattern Recognition}}. \bibinfo{pages}{1434--1441}.
\newblock
\urldef\tempurl%
\url{https://doi.org/10.1109/CVPR.2010.5539802}
\showDOI{\tempurl}


\bibitem[Furukawa and Ponce(2010)]%
        {5226635}
\bibfield{author}{\bibinfo{person}{Yasutaka Furukawa} {and}
  \bibinfo{person}{Jean Ponce}.} \bibinfo{year}{2010}\natexlab{}.
\newblock \showarticletitle{Accurate, Dense, and Robust Multiview Stereopsis}.
\newblock \bibinfo{journal}{\emph{IEEE Transactions on Pattern Analysis and
  Machine Intelligence}} \bibinfo{volume}{32}, \bibinfo{number}{8}
  (\bibinfo{year}{2010}), \bibinfo{pages}{1362--1376}.
\newblock
\urldef\tempurl%
\url{https://doi.org/10.1109/TPAMI.2009.161}
\showDOI{\tempurl}


\bibitem[Goesele et~al\mbox{.}(2007)]%
        {goesele2007multi}
\bibfield{author}{\bibinfo{person}{Michael Goesele}, \bibinfo{person}{Noah
  Snavely}, \bibinfo{person}{Brian Curless}, \bibinfo{person}{Hugues Hoppe},
  {and} \bibinfo{person}{Steven~M Seitz}.} \bibinfo{year}{2007}\natexlab{}.
\newblock \showarticletitle{Multi-view stereo for community photo collections}.
  In \bibinfo{booktitle}{\emph{2007 IEEE 11th International Conference on
  Computer Vision}}. IEEE, \bibinfo{pages}{1--8}.
\newblock


\bibitem[Goodfellow et~al\mbox{.}(2014)]%
        {goodfellow2014generative}
\bibfield{author}{\bibinfo{person}{Ian Goodfellow}, \bibinfo{person}{Jean
  Pouget-Abadie}, \bibinfo{person}{Mehdi Mirza}, \bibinfo{person}{Bing Xu},
  \bibinfo{person}{David Warde-Farley}, \bibinfo{person}{Sherjil Ozair},
  \bibinfo{person}{Aaron Courville}, {and} \bibinfo{person}{Yoshua Bengio}.}
  \bibinfo{year}{2014}\natexlab{}.
\newblock \showarticletitle{Generative adversarial nets}.
\newblock \bibinfo{journal}{\emph{Advances in neural information processing
  systems}}  \bibinfo{volume}{27} (\bibinfo{year}{2014}).
\newblock


\bibitem[Gu{\'e}don and Lepetit(2024)]%
        {guedon2024sugar}
\bibfield{author}{\bibinfo{person}{Antoine Gu{\'e}don} {and}
  \bibinfo{person}{Vincent Lepetit}.} \bibinfo{year}{2024}\natexlab{}.
\newblock \showarticletitle{Sugar: Surface-aligned Gaussian Splatting for
  efficient 3d mesh reconstruction and high-quality mesh rendering}. In
  \bibinfo{booktitle}{\emph{Proceedings of the IEEE/CVF Conference on Computer
  Vision and Pattern Recognition}}. \bibinfo{pages}{5354--5363}.
\newblock


\bibitem[Heinly et~al\mbox{.}(2015)]%
        {heinly2015reconstructing}
\bibfield{author}{\bibinfo{person}{Jared Heinly}, \bibinfo{person}{Johannes~L
  Schonberger}, \bibinfo{person}{Enrique Dunn}, {and}
  \bibinfo{person}{Jan-Michael Frahm}.} \bibinfo{year}{2015}\natexlab{}.
\newblock \showarticletitle{Reconstructing the world* in six days*(as captured
  by the yahoo 100 million image dataset)}. In
  \bibinfo{booktitle}{\emph{Proceedings of the IEEE conference on computer
  vision and pattern recognition}}. \bibinfo{pages}{3287--3295}.
\newblock


\bibitem[Huang et~al\mbox{.}(2021)]%
        {huang2021m3vsnet}
\bibfield{author}{\bibinfo{person}{Baichuan Huang}, \bibinfo{person}{Hongwei
  Yi}, \bibinfo{person}{Can Huang}, \bibinfo{person}{Yijia He},
  \bibinfo{person}{Jingbin Liu}, {and} \bibinfo{person}{Xiao Liu}.}
  \bibinfo{year}{2021}\natexlab{}.
\newblock \showarticletitle{M3VSNet: Unsupervised multi-metric multi-view
  stereo network}. In \bibinfo{booktitle}{\emph{2021 IEEE International
  Conference on Image Processing (ICIP)}}. IEEE, \bibinfo{pages}{3163--3167}.
\newblock


\bibitem[Huang et~al\mbox{.}(2024)]%
        {Huang2DGS2024}
\bibfield{author}{\bibinfo{person}{Binbin Huang}, \bibinfo{person}{Zehao Yu},
  \bibinfo{person}{Anpei Chen}, \bibinfo{person}{Andreas Geiger}, {and}
  \bibinfo{person}{Shenghua Gao}.} \bibinfo{year}{2024}\natexlab{}.
\newblock \showarticletitle{2D Gaussian Splatting for Geometrically Accurate
  Radiance Fields}.
\newblock \bibinfo{journal}{\emph{SIGGRAPH}} (\bibinfo{year}{2024}).
\newblock


\bibitem[Jiang et~al\mbox{.}(2024)]%
        {jiang2024hifi4g}
\bibfield{author}{\bibinfo{person}{Yuheng Jiang}, \bibinfo{person}{Zhehao
  Shen}, \bibinfo{person}{Penghao Wang}, \bibinfo{person}{Zhuo Su},
  \bibinfo{person}{Yu Hong}, \bibinfo{person}{Yingliang Zhang},
  \bibinfo{person}{Jingyi Yu}, {and} \bibinfo{person}{Lan Xu}.}
  \bibinfo{year}{2024}\natexlab{}.
\newblock \showarticletitle{Hifi4g: High-fidelity human performance rendering
  via compact Gaussian Splatting}. In \bibinfo{booktitle}{\emph{Proceedings of
  the IEEE/CVF Conference on Computer Vision and Pattern Recognition}}.
  \bibinfo{pages}{19734--19745}.
\newblock


\bibitem[Kazhdan et~al\mbox{.}(2006)]%
        {kazhdan2006poisson}
\bibfield{author}{\bibinfo{person}{Michael Kazhdan}, \bibinfo{person}{Matthew
  Bolitho}, {and} \bibinfo{person}{Hugues Hoppe}.}
  \bibinfo{year}{2006}\natexlab{}.
\newblock \showarticletitle{Poisson surface reconstruction}. In
  \bibinfo{booktitle}{\emph{Proceedings of the fourth Eurographics symposium on
  Geometry processing}}, Vol.~\bibinfo{volume}{7}. \bibinfo{pages}{0}.
\newblock


\bibitem[Keetha et~al\mbox{.}(2024)]%
        {keetha2024splatam}
\bibfield{author}{\bibinfo{person}{Nikhil Keetha}, \bibinfo{person}{Jay
  Karhade}, \bibinfo{person}{Krishna~Murthy Jatavallabhula},
  \bibinfo{person}{Gengshan Yang}, \bibinfo{person}{Sebastian Scherer},
  \bibinfo{person}{Deva Ramanan}, {and} \bibinfo{person}{Jonathon Luiten}.}
  \bibinfo{year}{2024}\natexlab{}.
\newblock \showarticletitle{SplaTAM: Splat, Track \& Map 3D Gaussians for Dense
  RGB-D SLAM}. In \bibinfo{booktitle}{\emph{Proceedings of the IEEE/CVF
  Conference on Computer Vision and Pattern Recognition}}.
\newblock


\bibitem[Kerbl et~al\mbox{.}(2023)]%
        {kerbl3Dgaussians}
\bibfield{author}{\bibinfo{person}{Bernhard Kerbl}, \bibinfo{person}{Georgios
  Kopanas}, \bibinfo{person}{Thomas Leimk{\"u}hler}, {and}
  \bibinfo{person}{George Drettakis}.} \bibinfo{year}{2023}\natexlab{}.
\newblock \showarticletitle{3D Gaussian Splatting for Real-Time Radiance Field
  Rendering}.
\newblock \bibinfo{journal}{\emph{ACM Transactions on Graphics}}
  \bibinfo{volume}{42}, \bibinfo{number}{4} (\bibinfo{date}{July}
  \bibinfo{year}{2023}).
\newblock
\urldef\tempurl%
\url{https://repo-sam.inria.fr/fungraph/3d-gaussian-splatting/}
\showURL{%
\tempurl}


\bibitem[Kerbl et~al\mbox{.}(2024)]%
        {hierarchicalgaussians24}
\bibfield{author}{\bibinfo{person}{Bernhard Kerbl}, \bibinfo{person}{Andreas
  Meuleman}, \bibinfo{person}{Georgios Kopanas}, \bibinfo{person}{Michael
  Wimmer}, \bibinfo{person}{Alexandre Lanvin}, {and} \bibinfo{person}{George
  Drettakis}.} \bibinfo{year}{2024}\natexlab{}.
\newblock \showarticletitle{A Hierarchical 3D Gaussian Representation for
  Real-Time Rendering of Very Large Datasets}.
\newblock \bibinfo{journal}{\emph{ACM Transactions on Graphics}}
  \bibinfo{volume}{43}, \bibinfo{number}{4} (\bibinfo{date}{July}
  \bibinfo{year}{2024}).
\newblock
\urldef\tempurl%
\url{https://repo-sam.inria.fr/fungraph/hierarchical-3d-gaussians/}
\showURL{%
\tempurl}


\bibitem[Kerr et~al\mbox{.}(2023)]%
        {kerr2023lerf}
\bibfield{author}{\bibinfo{person}{Justin Kerr}, \bibinfo{person}{Chung~Min
  Kim}, \bibinfo{person}{Ken Goldberg}, \bibinfo{person}{Angjoo Kanazawa},
  {and} \bibinfo{person}{Matthew Tancik}.} \bibinfo{year}{2023}\natexlab{}.
\newblock \showarticletitle{Lerf: Language embedded radiance fields}. In
  \bibinfo{booktitle}{\emph{Proceedings of the IEEE/CVF International
  Conference on Computer Vision}}. \bibinfo{pages}{19729--19739}.
\newblock


\bibitem[Kirillov et~al\mbox{.}(2023)]%
        {Kirillov_2023_ICCV}
\bibfield{author}{\bibinfo{person}{Alexander Kirillov}, \bibinfo{person}{Eric
  Mintun}, \bibinfo{person}{Nikhila Ravi}, \bibinfo{person}{Hanzi Mao},
  \bibinfo{person}{Chloe Rolland}, \bibinfo{person}{Laura Gustafson},
  \bibinfo{person}{Tete Xiao}, \bibinfo{person}{Spencer Whitehead},
  \bibinfo{person}{Alexander~C. Berg}, \bibinfo{person}{Wan-Yen Lo},
  \bibinfo{person}{Piotr Dollar}, {and} \bibinfo{person}{Ross Girshick}.}
  \bibinfo{year}{2023}\natexlab{}.
\newblock \showarticletitle{Segment Anything}. In
  \bibinfo{booktitle}{\emph{Proceedings of the IEEE/CVF International
  Conference on Computer Vision (ICCV)}}. \bibinfo{pages}{4015--4026}.
\newblock


\bibitem[Kopanas et~al\mbox{.}(2021)]%
        {kopanas2021point}
\bibfield{author}{\bibinfo{person}{Georgios Kopanas}, \bibinfo{person}{Julien
  Philip}, \bibinfo{person}{Thomas Leimk{\"u}hler}, {and}
  \bibinfo{person}{George Drettakis}.} \bibinfo{year}{2021}\natexlab{}.
\newblock \showarticletitle{Point-Based Neural Rendering with Per-View
  Optimization}. In \bibinfo{booktitle}{\emph{Computer Graphics Forum}},
  Vol.~\bibinfo{volume}{40}. Wiley Online Library, \bibinfo{pages}{29--43}.
\newblock


\bibitem[Kulhanek and Sattler(2023)]%
        {kulhanek2023tetra}
\bibfield{author}{\bibinfo{person}{Jonas Kulhanek} {and}
  \bibinfo{person}{Torsten Sattler}.} \bibinfo{year}{2023}\natexlab{}.
\newblock \showarticletitle{Tetra-nerf: Representing neural radiance fields
  using tetrahedra}. In \bibinfo{booktitle}{\emph{Proceedings of the IEEE/CVF
  International Conference on Computer Vision}}. \bibinfo{pages}{18458--18469}.
\newblock


\bibitem[Kwok(2023)]%
        {antimatter15_splat}
\bibfield{author}{\bibinfo{person}{Kevin Kwok}.}
  \bibinfo{year}{2023}\natexlab{}.
\newblock \bibinfo{title}{splat}.
\newblock \bibinfo{howpublished}{\url{https://github.com/antimatter15/splat}}.
\newblock
\newblock
\shownote{Accessed: 2024-05-15}.


\bibitem[Leonard and Durrant-Whyte(1991)]%
        {leonard1991simultaneous}
\bibfield{author}{\bibinfo{person}{John~J Leonard} {and}
  \bibinfo{person}{Hugh~F Durrant-Whyte}.} \bibinfo{year}{1991}\natexlab{}.
\newblock \showarticletitle{Simultaneous map building and localization for an
  autonomous mobile robot.}. In \bibinfo{booktitle}{\emph{IROS}},
  Vol.~\bibinfo{volume}{3}. \bibinfo{pages}{1442--1447}.
\newblock


\bibitem[Li et~al\mbox{.}(2022)]%
        {li2022ds}
\bibfield{author}{\bibinfo{person}{Jingliang Li}, \bibinfo{person}{Zhengda Lu},
  \bibinfo{person}{Yiqun Wang}, \bibinfo{person}{Ying Wang}, {and}
  \bibinfo{person}{Jun Xiao}.} \bibinfo{year}{2022}\natexlab{}.
\newblock \showarticletitle{DS-MVSNet: Unsupervised Multi-view Stereo via Depth
  Synthesis}. In \bibinfo{booktitle}{\emph{Proceedings of the 30th ACM
  International Conference on Multimedia}}. \bibinfo{pages}{5593--5601}.
\newblock


\bibitem[Li et~al\mbox{.}(2020)]%
        {li2020attention}
\bibfield{author}{\bibinfo{person}{Jinquan Li}, \bibinfo{person}{Ling Pei},
  \bibinfo{person}{Danping Zou}, \bibinfo{person}{Songpengcheng Xia},
  \bibinfo{person}{Qi Wu}, \bibinfo{person}{Tao Li}, \bibinfo{person}{Zhen
  Sun}, {and} \bibinfo{person}{Wenxian Yu}.} \bibinfo{year}{2020}\natexlab{}.
\newblock \showarticletitle{Attention-SLAM: A visual monocular SLAM learning
  from human gaze}.
\newblock \bibinfo{journal}{\emph{IEEE Sensors Journal}} \bibinfo{volume}{21},
  \bibinfo{number}{5} (\bibinfo{year}{2020}), \bibinfo{pages}{6408--6420}.
\newblock


\bibitem[Li et~al\mbox{.}(2024)]%
        {li2024dngaussian}
\bibfield{author}{\bibinfo{person}{Jiahe Li}, \bibinfo{person}{Jiawei Zhang},
  \bibinfo{person}{Xiao Bai}, \bibinfo{person}{Jin Zheng}, \bibinfo{person}{Xin
  Ning}, \bibinfo{person}{Jun Zhou}, {and} \bibinfo{person}{Lin Gu}.}
  \bibinfo{year}{2024}\natexlab{}.
\newblock \showarticletitle{DnGaussian: Optimizing sparse-view 3d Gaussian
  radiance fields with global-local depth normalization}. In
  \bibinfo{booktitle}{\emph{Proceedings of the IEEE/CVF Conference on Computer
  Vision and Pattern Recognition}}. \bibinfo{pages}{20775--20785}.
\newblock


\bibitem[Li et~al\mbox{.}(2023)]%
        {li2023read}
\bibfield{author}{\bibinfo{person}{Zhuopeng Li}, \bibinfo{person}{Lu Li}, {and}
  \bibinfo{person}{Jianke Zhu}.} \bibinfo{year}{2023}\natexlab{}.
\newblock \showarticletitle{Read: Large-scale neural scene rendering for
  autonomous driving}. In \bibinfo{booktitle}{\emph{Proceedings of the AAAI
  Conference on Artificial Intelligence}}, Vol.~\bibinfo{volume}{37}.
  \bibinfo{pages}{1522--1529}.
\newblock


\bibitem[Liao et~al\mbox{.}(2022)]%
        {liao2022kitti}
\bibfield{author}{\bibinfo{person}{Yiyi Liao}, \bibinfo{person}{Jun Xie}, {and}
  \bibinfo{person}{Andreas Geiger}.} \bibinfo{year}{2022}\natexlab{}.
\newblock \showarticletitle{KITTI-360: A novel dataset and benchmarks for urban
  scene understanding in 2d and 3d}.
\newblock \bibinfo{journal}{\emph{IEEE Transactions on Pattern Analysis and
  Machine Intelligence}} \bibinfo{volume}{45}, \bibinfo{number}{3}
  (\bibinfo{year}{2022}), \bibinfo{pages}{3292--3310}.
\newblock


\bibitem[Lin et~al\mbox{.}(2023)]%
        {lin2023infinicity}
\bibfield{author}{\bibinfo{person}{Chieh~Hubert Lin},
  \bibinfo{person}{Hsin-Ying Lee}, \bibinfo{person}{Willi Menapace},
  \bibinfo{person}{Menglei Chai}, \bibinfo{person}{Aliaksandr Siarohin},
  \bibinfo{person}{Ming-Hsuan Yang}, {and} \bibinfo{person}{Sergey Tulyakov}.}
  \bibinfo{year}{2023}\natexlab{}.
\newblock \showarticletitle{Infini{C}ity: Infinite-Scale City Synthesis}. In
  \bibinfo{booktitle}{\emph{Proceedings of the IEEE/CVF international
  conference on computer vision}}.
\newblock


\bibitem[Lin et~al\mbox{.}(2024)]%
        {lin2024vastgaussian}
\bibfield{author}{\bibinfo{person}{Jiaqi Lin}, \bibinfo{person}{Zhihao Li},
  \bibinfo{person}{Xiao Tang}, \bibinfo{person}{Jianzhuang Liu},
  \bibinfo{person}{Shiyong Liu}, \bibinfo{person}{Jiayue Liu},
  \bibinfo{person}{Yangdi Lu}, \bibinfo{person}{Xiaofei Wu},
  \bibinfo{person}{Songcen Xu}, \bibinfo{person}{Youliang Yan}, {and}
  \bibinfo{person}{Wenming Yang}.} \bibinfo{year}{2024}\natexlab{}.
\newblock \showarticletitle{VastGaussian: Vast 3D Gaussians for Large Scene
  Reconstruction}. In \bibinfo{booktitle}{\emph{CVPR}}.
\newblock


\bibitem[Liu et~al\mbox{.}(2024)]%
        {liu2024citygaussian}
\bibfield{author}{\bibinfo{person}{Yang Liu}, \bibinfo{person}{He Guan},
  \bibinfo{person}{Chuanchen Luo}, \bibinfo{person}{Lue Fan},
  \bibinfo{person}{Junran Peng}, {and} \bibinfo{person}{Zhaoxiang Zhang}.}
  \bibinfo{year}{2024}\natexlab{}.
\newblock \showarticletitle{CityGaussian: Real-time high-quality large-scale
  scene rendering with Gaussians}.
\newblock \bibinfo{journal}{\emph{arXiv preprint arXiv:2404.01133}}
  (\bibinfo{year}{2024}).
\newblock


\bibitem[Lu and Dai.(2024)]%
        {scaffoldgs}
\bibfield{author}{\bibinfo{person}{Mulin Yu Linning Xu Yuanbo Xiangli Limin
  Wang Dahua~Lin Lu, Tao} {and} \bibinfo{person}{Bo Dai.}}
  \bibinfo{year}{2024}\natexlab{}.
\newblock \showarticletitle{Scaffold-GS: Structured 3D Gaussians for
  View-Adaptive Rendering}.
\newblock \bibinfo{journal}{\emph{Conference on Computer Vision and Pattern
  Recognition (CVPR)}} (\bibinfo{year}{2024}).
\newblock


\bibitem[Matsuki et~al\mbox{.}(2024)]%
        {MatsukiCVPR2024}
\bibfield{author}{\bibinfo{person}{Hidenobu Matsuki}, \bibinfo{person}{Riku
  Murai}, \bibinfo{person}{Paul H.~J. Kelly}, {and} \bibinfo{person}{Andrew~J.
  Davison}.} \bibinfo{year}{2024}\natexlab{}.
\newblock \showarticletitle{{G}aussian {S}platting {SLAM}}.
\newblock  (\bibinfo{year}{2024}).
\newblock


\bibitem[Maye et~al\mbox{.}(2013)]%
        {maye2013self}
\bibfield{author}{\bibinfo{person}{J{\'e}r{\^o}me Maye}, \bibinfo{person}{Paul
  Furgale}, {and} \bibinfo{person}{Roland Siegwart}.}
  \bibinfo{year}{2013}\natexlab{}.
\newblock \showarticletitle{Self-supervised calibration for robotic systems}.
  In \bibinfo{booktitle}{\emph{2013 IEEE Intelligent Vehicles Symposium (IV)}}.
  IEEE, \bibinfo{pages}{473--480}.
\newblock


\bibitem[Mildenhall et~al\mbox{.}(2021)]%
        {mildenhall2021nerf}
\bibfield{author}{\bibinfo{person}{Ben Mildenhall}, \bibinfo{person}{Pratul~P
  Srinivasan}, \bibinfo{person}{Matthew Tancik}, \bibinfo{person}{Jonathan~T
  Barron}, \bibinfo{person}{Ravi Ramamoorthi}, {and} \bibinfo{person}{Ren Ng}.}
  \bibinfo{year}{2021}\natexlab{}.
\newblock \showarticletitle{Nerf: Representing scenes as neural radiance fields
  for view synthesis}.
\newblock \bibinfo{journal}{\emph{Commun. ACM}} \bibinfo{volume}{65},
  \bibinfo{number}{1} (\bibinfo{year}{2021}), \bibinfo{pages}{99--106}.
\newblock


\bibitem[Moulon et~al\mbox{.}(2013)]%
        {moulon2013adaptive}
\bibfield{author}{\bibinfo{person}{Pierre Moulon}, \bibinfo{person}{Pascal
  Monasse}, {and} \bibinfo{person}{Renaud Marlet}.}
  \bibinfo{year}{2013}\natexlab{}.
\newblock \showarticletitle{Adaptive structure from motion with a contrario
  model estimation}. In \bibinfo{booktitle}{\emph{Computer Vision--ACCV 2012:
  11th Asian Conference on Computer Vision, Daejeon, Korea, November 5-9, 2012,
  Revised Selected Papers, Part IV 11}}. Springer, \bibinfo{pages}{257--270}.
\newblock


\bibitem[M{\"u}ller et~al\mbox{.}(2022)]%
        {muller2022instant}
\bibfield{author}{\bibinfo{person}{Thomas M{\"u}ller}, \bibinfo{person}{Alex
  Evans}, \bibinfo{person}{Christoph Schied}, {and} \bibinfo{person}{Alexander
  Keller}.} \bibinfo{year}{2022}\natexlab{}.
\newblock \showarticletitle{Instant neural graphics primitives with a
  multiresolution hash encoding}.
\newblock \bibinfo{journal}{\emph{ACM Transactions on Graphics (ToG)}}
  \bibinfo{volume}{41}, \bibinfo{number}{4} (\bibinfo{year}{2022}),
  \bibinfo{pages}{1--15}.
\newblock


\bibitem[Niedermayr et~al\mbox{.}(2024)]%
        {niedermayr2024compressed}
\bibfield{author}{\bibinfo{person}{Simon Niedermayr}, \bibinfo{person}{Josef
  Stumpfegger}, {and} \bibinfo{person}{R{\"u}diger Westermann}.}
  \bibinfo{year}{2024}\natexlab{}.
\newblock \showarticletitle{Compressed 3d Gaussian splatting for accelerated
  novel view synthesis}. In \bibinfo{booktitle}{\emph{Proceedings of the
  IEEE/CVF Conference on Computer Vision and Pattern Recognition}}.
  \bibinfo{pages}{10349--10358}.
\newblock


\bibitem[Noguchi et~al\mbox{.}(2021)]%
        {noguchi2021neural}
\bibfield{author}{\bibinfo{person}{Atsuhiro Noguchi}, \bibinfo{person}{Xiao
  Sun}, \bibinfo{person}{Stephen Lin}, {and} \bibinfo{person}{Tatsuya Harada}.}
  \bibinfo{year}{2021}\natexlab{}.
\newblock \showarticletitle{Neural articulated radiance field}. In
  \bibinfo{booktitle}{\emph{Proceedings of the IEEE/CVF International
  Conference on Computer Vision}}. \bibinfo{pages}{5762--5772}.
\newblock


\bibitem[Oth et~al\mbox{.}(2013)]%
        {oth2013rolling}
\bibfield{author}{\bibinfo{person}{Luc Oth}, \bibinfo{person}{Paul Furgale},
  \bibinfo{person}{Laurent Kneip}, {and} \bibinfo{person}{Roland Siegwart}.}
  \bibinfo{year}{2013}\natexlab{}.
\newblock \showarticletitle{Rolling shutter camera calibration}. In
  \bibinfo{booktitle}{\emph{Proceedings of the IEEE Conference on Computer
  Vision and Pattern Recognition}}. \bibinfo{pages}{1360--1367}.
\newblock


\bibitem[Ponchio and Dellepiane(2016)]%
        {Ponchio2016MultiresolutionAF}
\bibfield{author}{\bibinfo{person}{Federico Ponchio} {and}
  \bibinfo{person}{Matteo Dellepiane}.} \bibinfo{year}{2016}\natexlab{}.
\newblock \showarticletitle{Multiresolution and fast decompression for optimal
  web-based rendering}.
\newblock \bibinfo{journal}{\emph{Graph. Model.}}  \bibinfo{volume}{88}
  (\bibinfo{year}{2016}), \bibinfo{pages}{1--11}.
\newblock
\urldef\tempurl%
\url{https://api.semanticscholar.org/CorpusID:8770708}
\showURL{%
\tempurl}


\bibitem[Rebain et~al\mbox{.}(2021)]%
        {rebain2021derf}
\bibfield{author}{\bibinfo{person}{Daniel Rebain}, \bibinfo{person}{Wei Jiang},
  \bibinfo{person}{Soroosh Yazdani}, \bibinfo{person}{Ke Li},
  \bibinfo{person}{Kwang~Moo Yi}, {and} \bibinfo{person}{Andrea Tagliasacchi}.}
  \bibinfo{year}{2021}\natexlab{}.
\newblock \showarticletitle{Derf: Decomposed radiance fields}. In
  \bibinfo{booktitle}{\emph{Proceedings of the IEEE/CVF Conference on Computer
  Vision and Pattern Recognition}}. \bibinfo{pages}{14153--14161}.
\newblock


\bibitem[Rehder et~al\mbox{.}(2016)]%
        {rehder2016extending}
\bibfield{author}{\bibinfo{person}{Joern Rehder}, \bibinfo{person}{Janosch
  Nikolic}, \bibinfo{person}{Thomas Schneider}, \bibinfo{person}{Timo
  Hinzmann}, {and} \bibinfo{person}{Roland Siegwart}.}
  \bibinfo{year}{2016}\natexlab{}.
\newblock \showarticletitle{Extending kalibr: Calibrating the extrinsics of
  multiple IMUs and of individual axes}. In \bibinfo{booktitle}{\emph{2016 IEEE
  International Conference on Robotics and Automation (ICRA)}}. IEEE,
  \bibinfo{pages}{4304--4311}.
\newblock


\bibitem[Rematas et~al\mbox{.}(2022)]%
        {rematas2022urban}
\bibfield{author}{\bibinfo{person}{Konstantinos Rematas},
  \bibinfo{person}{Andrew Liu}, \bibinfo{person}{Pratul~P Srinivasan},
  \bibinfo{person}{Jonathan~T Barron}, \bibinfo{person}{Andrea Tagliasacchi},
  \bibinfo{person}{Thomas Funkhouser}, {and} \bibinfo{person}{Vittorio
  Ferrari}.} \bibinfo{year}{2022}\natexlab{}.
\newblock \showarticletitle{Urban radiance fields}. In
  \bibinfo{booktitle}{\emph{Proceedings of the IEEE/CVF Conference on Computer
  Vision and Pattern Recognition}}. \bibinfo{pages}{12932--12942}.
\newblock


\bibitem[Ren et~al\mbox{.}(2024)]%
        {ren2024octree}
\bibfield{author}{\bibinfo{person}{Kerui Ren}, \bibinfo{person}{Lihan Jiang},
  \bibinfo{person}{Tao Lu}, \bibinfo{person}{Mulin Yu},
  \bibinfo{person}{Linning Xu}, \bibinfo{person}{Zhangkai Ni}, {and}
  \bibinfo{person}{Bo Dai}.} \bibinfo{year}{2024}\natexlab{}.
\newblock \showarticletitle{Octree-GS: Towards Consistent Real-time Rendering
  with LOD-Structured 3D Gaussians}.
\newblock \bibinfo{journal}{\emph{arXiv preprint arXiv:2403.17898}}
  (\bibinfo{year}{2024}).
\newblock


\bibitem[Roessle et~al\mbox{.}(2022)]%
        {roessle2022depthpriorsnerf}
\bibfield{author}{\bibinfo{person}{Barbara Roessle},
  \bibinfo{person}{Jonathan~T. Barron}, \bibinfo{person}{Ben Mildenhall},
  \bibinfo{person}{Pratul~P. Srinivasan}, {and} \bibinfo{person}{Matthias
  Nie{\ss}ner}.} \bibinfo{year}{2022}\natexlab{}.
\newblock \showarticletitle{Dense Depth Priors for Neural Radiance Fields from
  Sparse Input Views}. In \bibinfo{booktitle}{\emph{Proceedings of the IEEE/CVF
  Conference on Computer Vision and Pattern Recognition (CVPR)}}.
\newblock


\bibitem[Schauer and N{\"u}chter(2018)]%
        {schauer2018peopleremover}
\bibfield{author}{\bibinfo{person}{Johannes Schauer} {and}
  \bibinfo{person}{Andreas N{\"u}chter}.} \bibinfo{year}{2018}\natexlab{}.
\newblock \showarticletitle{The peopleremover—removing dynamic objects from
  3-d point cloud data by traversing a voxel occupancy grid}.
\newblock \bibinfo{journal}{\emph{IEEE robotics and automation letters}}
  \bibinfo{volume}{3}, \bibinfo{number}{3} (\bibinfo{year}{2018}),
  \bibinfo{pages}{1679--1686}.
\newblock


\bibitem[Schmuck et~al\mbox{.}(2021)]%
        {schmuck2021covins}
\bibfield{author}{\bibinfo{person}{Patrik Schmuck}, \bibinfo{person}{Thomas
  Ziegler}, \bibinfo{person}{Marco Karrer}, \bibinfo{person}{Jonathan
  Perraudin}, {and} \bibinfo{person}{Margarita Chli}.}
  \bibinfo{year}{2021}\natexlab{}.
\newblock \showarticletitle{Covins: Visual-inertial SLAM for centralized
  collaboration}. In \bibinfo{booktitle}{\emph{2021 IEEE International
  Symposium on Mixed and Augmented Reality Adjunct (ISMAR-Adjunct)}}. IEEE,
  \bibinfo{pages}{171--176}.
\newblock


\bibitem[Sch{\"u}tz et~al\mbox{.}(2016)]%
        {schutz2016potree}
\bibfield{author}{\bibinfo{person}{Markus Sch{\"u}tz} {et~al\mbox{.}}}
  \bibinfo{year}{2016}\natexlab{}.
\newblock \showarticletitle{Potree: Rendering large point clouds in web
  browsers}.
\newblock \bibinfo{journal}{\emph{Technische Universit{\"a}t Wien, Wiede{\'n}}}
  (\bibinfo{year}{2016}).
\newblock


\bibitem[Sch{\"u}tz et~al\mbox{.}(2020)]%
        {schutz2020fast}
\bibfield{author}{\bibinfo{person}{Markus Sch{\"u}tz}, \bibinfo{person}{Stefan
  Ohrhallinger}, {and} \bibinfo{person}{Michael Wimmer}.}
  \bibinfo{year}{2020}\natexlab{}.
\newblock \showarticletitle{Fast Out-of-Core Octree Generation for Massive
  Point Clouds}. In \bibinfo{booktitle}{\emph{Computer Graphics Forum}},
  Vol.~\bibinfo{volume}{39}. Wiley Online Library, \bibinfo{pages}{155--167}.
\newblock


\bibitem[Schönberger and Frahm(2016)]%
        {schonberger2016structure}
\bibfield{author}{\bibinfo{person}{Johannes~L Schönberger} {and}
  \bibinfo{person}{Jan-Michael Frahm}.} \bibinfo{year}{2016}\natexlab{}.
\newblock \showarticletitle{Structure-from-motion revisited}. In
  \bibinfo{booktitle}{\emph{Proceedings of the IEEE conference on computer
  vision and pattern recognition}}. \bibinfo{pages}{4104--4113}.
\newblock


\bibitem[Seitz et~al\mbox{.}(2006)]%
        {seitz2006comparison}
\bibfield{author}{\bibinfo{person}{Steven~M Seitz}, \bibinfo{person}{Brian
  Curless}, \bibinfo{person}{James Diebel}, \bibinfo{person}{Daniel
  Scharstein}, {and} \bibinfo{person}{Richard Szeliski}.}
  \bibinfo{year}{2006}\natexlab{}.
\newblock \showarticletitle{A comparison and evaluation of multi-view stereo
  reconstruction algorithms}. In \bibinfo{booktitle}{\emph{2006 IEEE computer
  society conference on computer vision and pattern recognition (CVPR'06)}},
  Vol.~\bibinfo{volume}{1}. IEEE, \bibinfo{pages}{519--528}.
\newblock


\bibitem[Shan et~al\mbox{.}(2021)]%
        {shan2021lvi}
\bibfield{author}{\bibinfo{person}{Tixiao Shan}, \bibinfo{person}{Brendan
  Englot}, \bibinfo{person}{Carlo Ratti}, {and} \bibinfo{person}{Daniela Rus}.}
  \bibinfo{year}{2021}\natexlab{}.
\newblock \showarticletitle{Lvi-sam: Tightly-coupled lidar-visual-inertial
  odometry via smoothing and mapping}. In \bibinfo{booktitle}{\emph{2021 IEEE
  international conference on robotics and automation (ICRA)}}. IEEE,
  \bibinfo{pages}{5692--5698}.
\newblock


\bibitem[Shuai et~al\mbox{.}(2024)]%
        {LoG}
\bibfield{author}{\bibinfo{person}{Qing Shuai}, \bibinfo{person}{Haoyu Guo},
  \bibinfo{person}{Zhen Xu}, \bibinfo{person}{Haotong Lin},
  \bibinfo{person}{Sida Peng}, \bibinfo{person}{Hujun Bao}, {and}
  \bibinfo{person}{Xiaowei Zhou}.} \bibinfo{year}{2024}\natexlab{}.
\newblock \showarticletitle{Real-Time View Synthesis for Large Scenes with
  Millions of Square Meters}.
\newblock


\bibitem[Snavely et~al\mbox{.}(2008)]%
        {snavely2008skeletal}
\bibfield{author}{\bibinfo{person}{Noah Snavely}, \bibinfo{person}{Steven~M
  Seitz}, {and} \bibinfo{person}{Richard Szeliski}.}
  \bibinfo{year}{2008}\natexlab{}.
\newblock \showarticletitle{Skeletal graphs for efficient structure from
  motion}. In \bibinfo{booktitle}{\emph{2008 IEEE Conference on Computer Vision
  and Pattern Recognition}}. IEEE, \bibinfo{pages}{1--8}.
\newblock


\bibitem[Sun et~al\mbox{.}(2022)]%
        {SunSC22}
\bibfield{author}{\bibinfo{person}{Cheng Sun}, \bibinfo{person}{Min Sun}, {and}
  \bibinfo{person}{Hwann{-}Tzong Chen}.} \bibinfo{year}{2022}\natexlab{}.
\newblock \showarticletitle{Direct Voxel Grid Optimization: Super-fast
  Convergence for Radiance Fields Reconstruction}. In
  \bibinfo{booktitle}{\emph{CVPR}}.
\newblock


\bibitem[Sun et~al\mbox{.}(2023)]%
        {sun2023pointnerf++}
\bibfield{author}{\bibinfo{person}{Weiwei Sun}, \bibinfo{person}{Eduard
  Trulls}, \bibinfo{person}{Yang-Che Tseng}, \bibinfo{person}{Sneha Sambandam},
  \bibinfo{person}{Gopal Sharma}, \bibinfo{person}{Andrea Tagliasacchi}, {and}
  \bibinfo{person}{Kwang~Moo Yi}.} \bibinfo{year}{2023}\natexlab{}.
\newblock \showarticletitle{PointNeRF++: A multi-scale, point-based Neural
  Radiance Field}.
\newblock \bibinfo{journal}{\emph{arXiv preprint arXiv:2312.02362}}
  (\bibinfo{year}{2023}).
\newblock


\bibitem[Sweeney et~al\mbox{.}(2016)]%
        {sweeney2016large}
\bibfield{author}{\bibinfo{person}{Chris Sweeney}, \bibinfo{person}{Victor
  Fragoso}, \bibinfo{person}{Tobias H{\"o}llerer}, {and}
  \bibinfo{person}{Matthew Turk}.} \bibinfo{year}{2016}\natexlab{}.
\newblock \showarticletitle{Large scale sfm with the distributed camera model}.
  In \bibinfo{booktitle}{\emph{2016 Fourth International Conference on 3D
  Vision (3DV)}}. IEEE, \bibinfo{pages}{230--238}.
\newblock


\bibitem[Tancik et~al\mbox{.}(2022)]%
        {tancik2022block}
\bibfield{author}{\bibinfo{person}{Matthew Tancik}, \bibinfo{person}{Vincent
  Casser}, \bibinfo{person}{Xinchen Yan}, \bibinfo{person}{Sabeek Pradhan},
  \bibinfo{person}{Ben Mildenhall}, \bibinfo{person}{Pratul~P Srinivasan},
  \bibinfo{person}{Jonathan~T Barron}, {and} \bibinfo{person}{Henrik
  Kretzschmar}.} \bibinfo{year}{2022}\natexlab{}.
\newblock \showarticletitle{Block-nerf: Scalable large scene neural view
  synthesis}. In \bibinfo{booktitle}{\emph{Proceedings of the IEEE/CVF
  Conference on Computer Vision and Pattern Recognition}}.
  \bibinfo{pages}{8248--8258}.
\newblock


\bibitem[Tancik et~al\mbox{.}(2023)]%
        {nerfstudio}
\bibfield{author}{\bibinfo{person}{Matthew Tancik}, \bibinfo{person}{Ethan
  Weber}, \bibinfo{person}{Evonne Ng}, \bibinfo{person}{Ruilong Li},
  \bibinfo{person}{Brent Yi}, \bibinfo{person}{Justin Kerr},
  \bibinfo{person}{Terrance Wang}, \bibinfo{person}{Alexander Kristoffersen},
  \bibinfo{person}{Jake Austin}, \bibinfo{person}{Kamyar Salahi},
  \bibinfo{person}{Abhik Ahuja}, \bibinfo{person}{David McAllister}, {and}
  \bibinfo{person}{Angjoo Kanazawa}.} \bibinfo{year}{2023}\natexlab{}.
\newblock \showarticletitle{Nerfstudio: A Modular Framework for Neural Radiance
  Field Development}. In \bibinfo{booktitle}{\emph{ACM SIGGRAPH 2023 Conference
  Proceedings}} \emph{(\bibinfo{series}{SIGGRAPH '23})}.
\newblock


\bibitem[Tang et~al\mbox{.}(2023)]%
        {tang2023dreamgaussian}
\bibfield{author}{\bibinfo{person}{Jiaxiang Tang}, \bibinfo{person}{Jiawei
  Ren}, \bibinfo{person}{Hang Zhou}, \bibinfo{person}{Ziwei Liu}, {and}
  \bibinfo{person}{Gang Zeng}.} \bibinfo{year}{2023}\natexlab{}.
\newblock \showarticletitle{DreamGaussian: Generative Gaussian splatting for
  efficient 3d content creation}.
\newblock \bibinfo{journal}{\emph{arXiv preprint arXiv:2309.16653}}
  (\bibinfo{year}{2023}).
\newblock


\bibitem[Teed and Deng(2021)]%
        {teed2021droid}
\bibfield{author}{\bibinfo{person}{Zachary Teed} {and} \bibinfo{person}{Jia
  Deng}.} \bibinfo{year}{2021}\natexlab{}.
\newblock \showarticletitle{Droid-SLAM: Deep visual SLAM for monocular, stereo,
  and rgb-d cameras}.
\newblock \bibinfo{journal}{\emph{Advances in neural information processing
  systems}}  \bibinfo{volume}{34} (\bibinfo{year}{2021}),
  \bibinfo{pages}{16558--16569}.
\newblock


\bibitem[Turki et~al\mbox{.}(2022)]%
        {turki2022mega}
\bibfield{author}{\bibinfo{person}{Haithem Turki}, \bibinfo{person}{Deva
  Ramanan}, {and} \bibinfo{person}{Mahadev Satyanarayanan}.}
  \bibinfo{year}{2022}\natexlab{}.
\newblock \showarticletitle{Mega-nerf: Scalable construction of large-scale
  nerfs for virtual fly-throughs}. In \bibinfo{booktitle}{\emph{Proceedings of
  the IEEE/CVF Conference on Computer Vision and Pattern Recognition}}.
  \bibinfo{pages}{12922--12931}.
\newblock


\bibitem[Turkulainen et~al\mbox{.}(2024)]%
        {turkulainen2024dnsplatter}
\bibfield{author}{\bibinfo{person}{Matias Turkulainen}, \bibinfo{person}{Xuqian
  Ren}, \bibinfo{person}{Iaroslav Melekhov}, \bibinfo{person}{Otto Seiskari},
  \bibinfo{person}{Esa Rahtu}, {and} \bibinfo{person}{Juho Kannala}.}
  \bibinfo{year}{2024}\natexlab{}.
\newblock \bibinfo{title}{DN-Splatter: Depth and Normal Priors for Gaussian
  Splatting and Meshing}.
\newblock
\newblock
\showeprint[arxiv]{2403.17822}~[cs.CV]


\bibitem[Wang et~al\mbox{.}(2023)]%
        {wang2023f2}
\bibfield{author}{\bibinfo{person}{Peng Wang}, \bibinfo{person}{Yuan Liu},
  \bibinfo{person}{Zhaoxi Chen}, \bibinfo{person}{Lingjie Liu},
  \bibinfo{person}{Ziwei Liu}, \bibinfo{person}{Taku Komura},
  \bibinfo{person}{Christian Theobalt}, {and} \bibinfo{person}{Wenping Wang}.}
  \bibinfo{year}{2023}\natexlab{}.
\newblock \showarticletitle{F2-NeRF: Fast Neural Radiance Field Training with
  Free Camera Trajectories}. In \bibinfo{booktitle}{\emph{Proceedings of the
  IEEE/CVF Conference on Computer Vision and Pattern Recognition}}.
  \bibinfo{pages}{4150--4159}.
\newblock


\bibitem[Wu(2013)]%
        {wu2013towards}
\bibfield{author}{\bibinfo{person}{Changchang Wu}.}
  \bibinfo{year}{2013}\natexlab{}.
\newblock \showarticletitle{Towards linear-time incremental structure from
  motion}. In \bibinfo{booktitle}{\emph{2013 International Conference on 3D
  Vision-3DV 2013}}. IEEE, \bibinfo{pages}{127--134}.
\newblock


\bibitem[Wu et~al\mbox{.}(2024)]%
        {wu20244d}
\bibfield{author}{\bibinfo{person}{Guanjun Wu}, \bibinfo{person}{Taoran Yi},
  \bibinfo{person}{Jiemin Fang}, \bibinfo{person}{Lingxi Xie},
  \bibinfo{person}{Xiaopeng Zhang}, \bibinfo{person}{Wei Wei},
  \bibinfo{person}{Wenyu Liu}, \bibinfo{person}{Qi Tian}, {and}
  \bibinfo{person}{Xinggang Wang}.} \bibinfo{year}{2024}\natexlab{}.
\newblock \showarticletitle{4d Gaussian Splatting for real-time dynamic scene
  rendering}. In \bibinfo{booktitle}{\emph{Proceedings of the IEEE/CVF
  Conference on Computer Vision and Pattern Recognition}}.
  \bibinfo{pages}{20310--20320}.
\newblock


\bibitem[Wu et~al\mbox{.}(2023)]%
        {wu2023scanerf}
\bibfield{author}{\bibinfo{person}{Xiuchao Wu}, \bibinfo{person}{Jiamin Xu},
  \bibinfo{person}{Xin Zhang}, \bibinfo{person}{Hujun Bao},
  \bibinfo{person}{Qixing Huang}, \bibinfo{person}{Yujun Shen},
  \bibinfo{person}{James Tompkin}, {and} \bibinfo{person}{Weiwei Xu}.}
  \bibinfo{year}{2023}\natexlab{}.
\newblock \showarticletitle{ScaNeRF: Scalable Bundle-Adjusting Neural Radiance
  Fields for Large-Scale Scene Rendering}.
\newblock \bibinfo{journal}{\emph{ACM Transactions on Graphics (TOG)}}
  \bibinfo{volume}{42}, \bibinfo{number}{6} (\bibinfo{year}{2023}),
  \bibinfo{pages}{1--18}.
\newblock


\bibitem[Xie et~al\mbox{.}(2024)]%
        {xie2024citydreamer}
\bibfield{author}{\bibinfo{person}{Haozhe Xie}, \bibinfo{person}{Zhaoxi Chen},
  \bibinfo{person}{Fangzhou Hong}, {and} \bibinfo{person}{Ziwei Liu}.}
  \bibinfo{year}{2024}\natexlab{}.
\newblock \showarticletitle{Citydreamer: Compositional generative model of
  unbounded 3d cities}. In \bibinfo{booktitle}{\emph{Proceedings of the
  IEEE/CVF Conference on Computer Vision and Pattern Recognition}}.
  \bibinfo{pages}{9666--9675}.
\newblock


\bibitem[Xu et~al\mbox{.}(2021a)]%
        {xu2021self}
\bibfield{author}{\bibinfo{person}{Hongbin Xu}, \bibinfo{person}{Zhipeng Zhou},
  \bibinfo{person}{Yu Qiao}, \bibinfo{person}{Wenxiong Kang}, {and}
  \bibinfo{person}{Qiuxia Wu}.} \bibinfo{year}{2021}\natexlab{a}.
\newblock \showarticletitle{Self-supervised multi-view stereo via effective
  co-segmentation and data-augmentation}. In
  \bibinfo{booktitle}{\emph{Proceedings of the AAAI Conference on Artificial
  Intelligence}}, Vol.~\bibinfo{volume}{35}. \bibinfo{pages}{3030--3038}.
\newblock


\bibitem[Xu et~al\mbox{.}(2021b)]%
        {xu2021digging}
\bibfield{author}{\bibinfo{person}{Hongbin Xu}, \bibinfo{person}{Zhipeng Zhou},
  \bibinfo{person}{Yali Wang}, \bibinfo{person}{Wenxiong Kang},
  \bibinfo{person}{Baigui Sun}, \bibinfo{person}{Hao Li}, {and}
  \bibinfo{person}{Yu Qiao}.} \bibinfo{year}{2021}\natexlab{b}.
\newblock \showarticletitle{Digging into uncertainty in self-supervised
  multi-view stereo}. In \bibinfo{booktitle}{\emph{Proceedings of the IEEE/CVF
  International Conference on Computer Vision}}. \bibinfo{pages}{6078--6087}.
\newblock


\bibitem[Xu et~al\mbox{.}(2022)]%
        {xu2022point}
\bibfield{author}{\bibinfo{person}{Qiangeng Xu}, \bibinfo{person}{Zexiang Xu},
  \bibinfo{person}{Julien Philip}, \bibinfo{person}{Sai Bi},
  \bibinfo{person}{Zhixin Shu}, \bibinfo{person}{Kalyan Sunkavalli}, {and}
  \bibinfo{person}{Ulrich Neumann}.} \bibinfo{year}{2022}\natexlab{}.
\newblock \showarticletitle{Point-nerf: Point-based neural radiance fields}. In
  \bibinfo{booktitle}{\emph{Proceedings of the IEEE/CVF Conference on Computer
  Vision and Pattern Recognition}}. \bibinfo{pages}{5438--5448}.
\newblock


\bibitem[Yan et~al\mbox{.}(2024)]%
        {yan2024street}
\bibfield{author}{\bibinfo{person}{Yunzhi Yan}, \bibinfo{person}{Haotong Lin},
  \bibinfo{person}{Chenxu Zhou}, \bibinfo{person}{Weijie Wang},
  \bibinfo{person}{Haiyang Sun}, \bibinfo{person}{Kun Zhan},
  \bibinfo{person}{Xianpeng Lang}, \bibinfo{person}{Xiaowei Zhou}, {and}
  \bibinfo{person}{Sida Peng}.} \bibinfo{year}{2024}\natexlab{}.
\newblock \bibinfo{title}{Street Gaussians for Modeling Dynamic Urban Scenes}.
\newblock
\newblock
\showeprint[arxiv]{2401.01339}~[cs.CV]


\bibitem[Yang et~al\mbox{.}(2021)]%
        {yang2021self}
\bibfield{author}{\bibinfo{person}{Jiayu Yang}, \bibinfo{person}{Jose~M
  Alvarez}, {and} \bibinfo{person}{Miaomiao Liu}.}
  \bibinfo{year}{2021}\natexlab{}.
\newblock \showarticletitle{Self-supervised learning of depth inference for
  multi-view stereo}. In \bibinfo{booktitle}{\emph{Proceedings of the IEEE/CVF
  Conference on Computer Vision and Pattern Recognition}}.
  \bibinfo{pages}{7526--7534}.
\newblock


\bibitem[Yao et~al\mbox{.}(2018)]%
        {yao2018mvsnet}
\bibfield{author}{\bibinfo{person}{Yao Yao}, \bibinfo{person}{Zixin Luo},
  \bibinfo{person}{Shiwei Li}, \bibinfo{person}{Tian Fang}, {and}
  \bibinfo{person}{Long Quan}.} \bibinfo{year}{2018}\natexlab{}.
\newblock \showarticletitle{Mvsnet: Depth inference for unstructured multi-view
  stereo}. In \bibinfo{booktitle}{\emph{Proceedings of the European conference
  on computer vision (ECCV)}}. \bibinfo{pages}{767--783}.
\newblock


\bibitem[Yeshwanth et~al\mbox{.}(2023)]%
        {yeshwanth2023scannet++}
\bibfield{author}{\bibinfo{person}{Chandan Yeshwanth},
  \bibinfo{person}{Yueh-Cheng Liu}, \bibinfo{person}{Matthias Nie{\ss}ner},
  {and} \bibinfo{person}{Angela Dai}.} \bibinfo{year}{2023}\natexlab{}.
\newblock \showarticletitle{Scannet++: A high-fidelity dataset of 3d indoor
  scenes}. In \bibinfo{booktitle}{\emph{Proceedings of the IEEE/CVF
  International Conference on Computer Vision}}. \bibinfo{pages}{12--22}.
\newblock


\bibitem[Yu et~al\mbox{.}(2021)]%
        {yu2021pixelnerf}
\bibfield{author}{\bibinfo{person}{Alex Yu}, \bibinfo{person}{Vickie Ye},
  \bibinfo{person}{Matthew Tancik}, {and} \bibinfo{person}{Angjoo Kanazawa}.}
  \bibinfo{year}{2021}\natexlab{}.
\newblock \showarticletitle{pixelnerf: Neural radiance fields from one or few
  images}. In \bibinfo{booktitle}{\emph{Proceedings of the IEEE/CVF Conference
  on Computer Vision and Pattern Recognition}}. \bibinfo{pages}{4578--4587}.
\newblock


\bibitem[Yu et~al\mbox{.}(2024)]%
        {yu2024mip}
\bibfield{author}{\bibinfo{person}{Zehao Yu}, \bibinfo{person}{Anpei Chen},
  \bibinfo{person}{Binbin Huang}, \bibinfo{person}{Torsten Sattler}, {and}
  \bibinfo{person}{Andreas Geiger}.} \bibinfo{year}{2024}\natexlab{}.
\newblock \showarticletitle{Mip-splatting: Alias-free 3d Gaussian Splatting}.
  In \bibinfo{booktitle}{\emph{Proceedings of the IEEE/CVF Conference on
  Computer Vision and Pattern Recognition}}. \bibinfo{pages}{19447--19456}.
\newblock


\bibitem[Zhang et~al\mbox{.}(2020)]%
        {zhang2020nerf++}
\bibfield{author}{\bibinfo{person}{Kai Zhang}, \bibinfo{person}{Gernot
  Riegler}, \bibinfo{person}{Noah Snavely}, {and} \bibinfo{person}{Vladlen
  Koltun}.} \bibinfo{year}{2020}\natexlab{}.
\newblock \showarticletitle{Nerf++: Analyzing and improving neural radiance
  fields}.
\newblock \bibinfo{journal}{\emph{arXiv preprint arXiv:2010.07492}}
  (\bibinfo{year}{2020}).
\newblock


\bibitem[Zhang et~al\mbox{.}(2018)]%
        {zhang2018unreasonable}
\bibfield{author}{\bibinfo{person}{Richard Zhang}, \bibinfo{person}{Phillip
  Isola}, \bibinfo{person}{Alexei~A Efros}, \bibinfo{person}{Eli Shechtman},
  {and} \bibinfo{person}{Oliver Wang}.} \bibinfo{year}{2018}\natexlab{}.
\newblock \showarticletitle{The unreasonable effectiveness of deep features as
  a perceptual metric}. In \bibinfo{booktitle}{\emph{Proceedings of the IEEE
  conference on computer vision and pattern recognition}}.
  \bibinfo{pages}{586--595}.
\newblock


\bibitem[Zhao et~al\mbox{.}(2022a)]%
        {zhao2022human}
\bibfield{author}{\bibinfo{person}{Fuqiang Zhao}, \bibinfo{person}{Yuheng
  Jiang}, \bibinfo{person}{Kaixin Yao}, \bibinfo{person}{Jiakai Zhang},
  \bibinfo{person}{Liao Wang}, \bibinfo{person}{Haizhao Dai},
  \bibinfo{person}{Yuhui Zhong}, \bibinfo{person}{Yingliang Zhang},
  \bibinfo{person}{Minye Wu}, \bibinfo{person}{Lan Xu}, {et~al\mbox{.}}}
  \bibinfo{year}{2022}\natexlab{a}.
\newblock \showarticletitle{Human performance modeling and rendering via neural
  animated mesh}.
\newblock \bibinfo{journal}{\emph{ACM Transactions on Graphics (TOG)}}
  \bibinfo{volume}{41}, \bibinfo{number}{6} (\bibinfo{year}{2022}),
  \bibinfo{pages}{1--17}.
\newblock


\bibitem[Zhao et~al\mbox{.}(2022b)]%
        {zhao2022humannerf}
\bibfield{author}{\bibinfo{person}{Fuqiang Zhao}, \bibinfo{person}{Wei Yang},
  \bibinfo{person}{Jiakai Zhang}, \bibinfo{person}{Pei Lin},
  \bibinfo{person}{Yingliang Zhang}, \bibinfo{person}{Jingyi Yu}, {and}
  \bibinfo{person}{Lan Xu}.} \bibinfo{year}{2022}\natexlab{b}.
\newblock \showarticletitle{Humannerf: Efficiently generated human radiance
  field from sparse inputs}. In \bibinfo{booktitle}{\emph{Proceedings of the
  IEEE/CVF Conference on Computer Vision and Pattern Recognition}}.
  \bibinfo{pages}{7743--7753}.
\newblock


\bibitem[Zhou et~al\mbox{.}(2024)]%
        {zhou2024drivinggaussian}
\bibfield{author}{\bibinfo{person}{Xiaoyu Zhou}, \bibinfo{person}{Zhiwei Lin},
  \bibinfo{person}{Xiaojun Shan}, \bibinfo{person}{Yongtao Wang},
  \bibinfo{person}{Deqing Sun}, {and} \bibinfo{person}{Ming-Hsuan Yang}.}
  \bibinfo{year}{2024}\natexlab{}.
\newblock \showarticletitle{DrivingGaussian: Composite Gaussian Splatting for
  surrounding dynamic autonomous driving scenes}. In
  \bibinfo{booktitle}{\emph{Proceedings of the IEEE/CVF Conference on Computer
  Vision and Pattern Recognition}}. \bibinfo{pages}{21634--21643}.
\newblock


\bibitem[Zwicker et~al\mbox{.}(2001a)]%
        {ewa}
\bibfield{author}{\bibinfo{person}{M. Zwicker}, \bibinfo{person}{H. Pfister},
  \bibinfo{person}{J. van Baar}, {and} \bibinfo{person}{M. Gross}.}
  \bibinfo{year}{2001}\natexlab{a}.
\newblock \showarticletitle{EWA volume splatting}. In
  \bibinfo{booktitle}{\emph{Proceedings Visualization, 2001. VIS '01.}}
  \bibinfo{pages}{29--538}.
\newblock
\urldef\tempurl%
\url{https://doi.org/10.1109/VISUAL.2001.964490}
\showDOI{\tempurl}


\bibitem[Zwicker et~al\mbox{.}(2001b)]%
        {zwicker2001surface}
\bibfield{author}{\bibinfo{person}{Matthias Zwicker},
  \bibinfo{person}{Hanspeter Pfister}, \bibinfo{person}{Jeroen Van~Baar}, {and}
  \bibinfo{person}{Markus Gross}.} \bibinfo{year}{2001}\natexlab{b}.
\newblock \showarticletitle{Surface splatting}. In
  \bibinfo{booktitle}{\emph{Proceedings of the 28th annual conference on
  Computer graphics and interactive techniques}}. \bibinfo{pages}{371--378}.
\newblock


\end{thebibliography}

% \section{Discussion & Future Work}

% Appendix
\appendix

\balance
%之后有空验算一下 e^C
\section{Specifications and Calibration of Polar Device}
\label{app_polar_specs}

% To ensure replicability, we meticulously detail the components of our Polar scanner. 
Below, we provide the configuration and calibration details of our Polar scanner. 
The data acquisition unit comprises an Ouster OS0-128 REV6 LiDAR, an Insta360 ONE RS 1-inch 360 Edition Camera, and an Xsens MTi-630 IMU. Time synchronization across sensors is achieved using DEITY TC-1 timecode generators and simulated analog GPS NMEA time signals. The data collection unit and processing unit weigh 1250g and 850g, respectively, excluding the battery. The total apparatus weight does not exceed 3.2 kg, making it ideal for both handheld and wearable scanning applications. Additionally, we have engineered an over-shoulder support system to enhance operator stability during scans. 

For the intrinsic parameter calibration of the fisheye camera and the IMU sensor, we leverage the program provided by OpenCV and the Allan Variance ROS toolbox~\cite{rehder2016extending, furgale2013unified, furgale2012continuous, maye2013self, oth2013rolling}, respectively. 
The extrinsic calibration between the fisheye camera and the IMU uses the Kalibr calibration program~\cite{rehder2016extending}.
% First, we calibrate the intrinsic parameters of the fish-eye Camera using the program provided by OpenCV and those of the IMU by the Allan Variance ROS toolbox\cite{}. 
Due to the sparsity, noisiness, and uncolored nature of the point cloud data collected by our LiDAR sensor, it is hard to establish correspondences between the LiDAR data and the images captured by the fisheye camera, making relative pose estimation between the sensors difficult.
% Due to the different modalities (colored images Vs. uncolored sparse point cloud) between the data captured by the camera and Lidar sensors, the direct relative pose estimation between the sensors is difficult. 
To address this issue, we introduce an additional sensor, a FARO laser scanner. This scanner provides dense, accurate, and colored point clouds and thus functions as a bridge for the relative pose estimation between the LiDAR sensor and the fisheye camera. 
Specifically, we adopt a checkerboard as our calibration scene and capture data from it using the FARO laser scanner and Polar device. 
The relative pose between the FARO laser scanner and the LiDAR sensor in our Polar device is achieved by point cloud registration, and that between the FARO laser scanner and the fisheye camera is obtained by establishing color correspondences between the point cloud captured by the FARO sensor and the fisheye image. 
By connecting the transformation from FARO to LiDAR and from FARO to the fisheye camera, we can finally calculate the extrinsic parameters between the LiDAR and the fisheye camera. 
% The fisheye camera and IMU are calibrated using the Kalibr calibration program \cite{rehder2016extending}.
We repeat this process twelve times in our implementation to reduce experimental error. It is important to note that FARO was only used to calibrate between the fisheye camera and the LiDAR \textbf{before} we start scanning a garage, and the process only needs to be conducted once.

\section{Data Preprocessing and Mesh cleanup}
\label{app_data_clean}
To ensure data privacy, we anonymized vehicle license plate information recorded in the underground garage data collection. Specifically, we used image blurring to hide license plate numbers from identification. This step is essential to protect individual privacy and comply with data protection regulations.
To remove the dynamic objects in the collected data, we apply Segment Anything~\cite{Kirillov_2023_ICCV} to the images.  
For the point cloud data, we estimate whether the points scanned at one LiDAR frame are also observed at other frames, similar to Schauer and N{\"u}chter\textcolor{orange}{\shortcite{schauer2018peopleremover}}.
For mesh cleaning, we construct a kd-tree from the point cloud of LIV-SLAM and calculate the nearest neighbour distance of each mesh face (k = 1). If the distance exceeds \textcolor{orange}{0.1 m}, we remove the face. 
% The reviewer is correct that LIV-SLAM is composed of a bag of SOTA techniques but executed in a particular order to fit our needs. We will clarify this point and shorten the narrative.

\section{Details of depth calculation}
\label{app_depth}
%为了更精准地对depth进行计算，我们将每个高斯看作概率密度函数，通过计算该概率密度下深度的期望值的方法来得到每个高斯primitive的深度di。根据论文EWA，我们将ray space下的3D gaussian作为概率密度，可得每个像素的gaussian期望深度为
 
% 其中gi为gaussian的函数，x=[x0,x1,x2]为ray space下的空间坐标，其中x0,x1
% 为像素坐标，x2的坐标轴朝向与像素光线方向相同，t=x2/l为x在相机坐标系下的深度， . 设ray space下的gaussian的中心为p=[p0,p1,p2],方差矩阵为sigma，我们令y=x-p， 公式X可以化简为

 To accurately calculate depth,  we treat each Gaussian as a probability density function. Following~\cite{ewa}, we get the depth of each Gaussian primitive, \( d_i \), by calculating the expected value of depth under this probability density as:
 \begin{equation}
d_i=\frac{\int_{-\infty}^{+\infty} t g_i(\boldsymbol{x}) d x_2}{\int_{-\infty}^{+\infty} g_i(\boldsymbol{x}) d x_2}=\frac{1}{l} \frac{\int_{-\infty}^{+\infty} x_2 g_i\left(x_2 \mid x_0, x_1\right) d x_2}{\int_{-\infty}^{+\infty} g_i\left(x_2 \mid x_0, x_1\right) d x_2},
\label{depth1}
\end{equation}
where $g_i$ is the Gaussian function, $\boldsymbol{x}=[x_0,x_1,x_2]^T$ is the coordinate of the space in the ray space, where $x_0,x_1$ are the pixel coordinates, $x_2$ is oriented in the same direction as the pixel ray. 
As in~\cite{ewa}, $t=x_2/l$ is the depth of $\boldsymbol{x}$ in the camera coordinate system and
 $l=\sqrt{x_0^2+x_1^2+1}$. 
We denote the center of the Gaussian in ray space as \( \boldsymbol{p} = [p_0, p_1, p_2] \) and the covariance matrix as $\boldsymbol{\boldsymbol{\Sigma}}$, by defining \( \boldsymbol{y} = \boldsymbol{x} - \boldsymbol{p} \), the Eqn.~\ref{depth1} can be simplified as:
\begin{equation}
\begin{split}
    d_i &= \frac{1}{l} \frac{\int_{-\infty}^{+\infty}\left(y_2+p_2\right) g_i(\boldsymbol{y}+\boldsymbol{p}) d x_2}{\int_{-\infty}^{+\infty} g_i(\boldsymbol{y}+\boldsymbol{p}) d x_2} \\
&= \frac{1}{l}\left(p_2+\frac{\int_{-\infty}^{+\infty} y_2 g_i(\boldsymbol{y}+\boldsymbol{p}) d y_2}{\int_{-\infty}^{+\infty} g_i(\boldsymbol{y}+\boldsymbol{p}) d y_2}\right).
\end{split}
\label{eq1}
\end{equation}
Then, $g(\boldsymbol{y}+\boldsymbol{p})$ can be expanded as:
%$e^{-A y_2^2+B y_2+C}=e^C \cdot e^{-A y_2^2+B y_2}$,
\begin{equation}
% \begin{split}
    g(\boldsymbol{y}+\boldsymbol{p})=e^{-A y_2^2+B y_2+C} 
    =e^C \cdot e^{-A y_2^2+B y_2},
% \end{split}
\end{equation}
where 
\begin{equation}
    \begin{split}
        A&=\frac{1}{2}\left(\boldsymbol{\Sigma}^{-1}\right)_{2,2}\\
        B&=-\left(\boldsymbol{\Sigma}^{-1}\right)_{2,0} y_0-\left(\boldsymbol{\Sigma}^{-1}\right)_{2,1} y_1\\
        C&=-\frac{1}{2}\left(\boldsymbol{\Sigma}^{-1}\right)_{0,0} y_0^2-\frac{1}{2}\left(\boldsymbol{\Sigma}^{-1}\right)_{1,1} y_1^2-\left(\boldsymbol{\Sigma}^{-1}\right)_{1,0} y_0 y_1\\.
    \end{split}
\end{equation}
% $A=\frac{1}{2}\left(\boldsymbol{\Sigma}^{-1}\right)_{2,2}, \quad B=-\left(\boldsymbol{\Sigma}^{-1}\right)_{2,0} y_0-\left(\boldsymbol{\Sigma}^{-1}\right)_{2,1} y_1$,
% $
% C=-\frac{1}{2}\left(\boldsymbol{\Sigma}^{-1}\right)_{0,0} y_0^2-\frac{1}{2}\left(\boldsymbol{\Sigma}^{-1}\right)_{1,1} y_1^2-\left(\boldsymbol{\Sigma}^{-1}\right)_{1,0} y_0 y_1
% $.
By integrating an arbitrary Gaussian function, we obtain: 
% $\int_{-\infty}^{+\infty} e^{-A u^2+B u} d u=\sqrt{\frac{\pi}{A}} e^{\frac{B^2}{4 A}}$ and $\int_{-\infty}^{+\infty} u e^{-A u^2+B u} d u=\frac{B}{2 A} \sqrt{\frac{\pi}{A}} e^{\frac{B^2}{4 A}}$.
\begin{equation}
    \int_{-\infty}^{+\infty} e^{-A u^2+B u} d u=\sqrt{\frac{\pi}{A}} e^{\frac{B^2}{4 A}}, 
    \label{eq2}
\end{equation}
\begin{equation}
\int_{-\infty}^{+\infty} u e^{-A u^2+B u} d u=\frac{B}{2 A} \sqrt{\frac{\pi}{A}} e^{\frac{B^2}{4 A}}.
\label{eq3}
\end{equation}
Next, substituting Eqn. \ref{eq2} and Eqn. \ref{eq3} into Eqn. \ref{eq1} enables us to obtain:
\begin{equation}
d_i=\frac{1}{l}\left(p_2+\frac{\int_{-\infty}^{+\infty} y_2 e^{-A y_2^2+B y_2} d y_2}{\int_{-\infty}^{+\infty} e^{-A y_2^2+B y_2} d y_2}\right)=\frac{1}{l}\left(p_2+\frac{B}{2 A}\right). 
\label{eq4}
\end{equation}
Finally, through simplification, Eqn.~\ref{eq4} can be transformed into Eqn.~\ref{eq6}.

Before depth supervision, we normalize both the captured depth and predicted depth to $0-1$ as follows:
\begin{equation}
R(D)= \begin{cases}\frac{1}{2 \beta} D & (D<\beta) \\ 1-\frac{\beta}{2 D} & (D \geq \beta)\end{cases},
\end{equation}
where $\beta=10$, as the depth values within 10 meters are accurate. In this way, the same depth error results in a greater loss when closer to the camera than at further distances. It effectively avoids excessive depth loss for distant Gaussians due to inherently large depth values. 
This normalization also facilitates the storage of depth maps.

% \newpage

% \section{Comparison of Training Times for Different Input Point Cloud}
% \label{appendix:trainingtime}

% \begin{table}[H]
% \small
% \setlength{\abovecaptionskip}{0pt}
% \setlength{\belowcaptionskip}{0pt}
% \caption{Comparison of point cloud densities and corresponding training times across two datasets.\textcolor{orange}{We choose one block from each dataset respectively.} The training times are based on tests performed with an NVIDIA A6000 GPU.}

% \setlength{\tabcolsep}{2.5mm}{
%     \begin{tabular}{lcccc}
%         \toprule
%         \multirow{3}{*}{\begin{tabular}[c]{@{}c@{}}Point Cloud\\ Density\end{tabular}} & \multicolumn{2}{c}{Arts Center} & \multicolumn{2}{c}{Shopping Mall 3} \\
%     \cline{2-5}
%        & \multirow{2}{*}{\begin{tabular}[c]{@{}c@{}}Numbers of \\ Gaussians\end{tabular}} &  \multirow{2}{*}{\begin{tabular}[c]{@{}c@{}}Training \\Time\end{tabular}} & \multirow{2}{*}{\begin{tabular}[c]{@{}c@{}}Numbers of \\ Gaussians\end{tabular}} & \multirow{2}{*}{\begin{tabular}[c]{@{}c@{}}Training \\Time\end{tabular}}\\
%           \\
%         \midrule
%         2 cm (Mesh)   & 17698464 & 10:55:07 & 36042237 & 31:01:11 \\
%         6 cm (Mesh)   & 2300174  & 02:47:02 & 4559012  & 06:35:14 \\
%         8 cm (Mesh)   & 1306965  & 02:15:16 & 2566219  & 04:53:08 \\
%         10 cm (Mesh)  & 836921   & 02:02:50 & 1627634  & 04:06:29 \\
%         4 cm (LiDAR)  & 10024717 & 03:03:03 & 21162613 & 04:39:05 \\
%         \midrule
%         4 cm (Mesh)   & 4997081  & 02:23:27 & 10015908 & 10:58:39 \\
%         \bottomrule
%     \end{tabular}}
% \vspace{-2mm}
% \end{table}

\end{document}